\documentclass[10pt,twocolumn,letterpaper]{article}

\usepackage{cvpr}
\usepackage{times}
\usepackage{epsfig}
\usepackage{graphicx}
\usepackage{amsmath}
\usepackage{amssymb}
\usepackage[normalem]{ulem}

\usepackage[pagebackref=true,breaklinks=true,letterpaper=true,colorlinks,bookmarks=false]{hyperref}

\cvprfinalcopy 

\usepackage{caption}
\usepackage{subcaption}
\usepackage{paralist}
\usepackage{array}
\usepackage{placeins}
\usepackage{multirow}
\usepackage{epstopdf}
\usepackage{paralist}
\usepackage[pagebackref=true,breaklinks=true,letterpaper=true,colorlinks,bookmarks=false]{hyperref}

\usepackage[hyphenbreaks]{breakurl}

\usepackage[ruled]{algorithm2e}
\makeatletter
\def\BState{\State\hskip-\ALG@thistlm}
\makeatother
\ifcvprfinal\fi
\usepackage{color}

\ifcvprfinal\fi
\begin{document}

\title{Detect, Replace, Refine: Deep Structured Prediction For Pixel Wise Labeling}

\author{Spyros Gidaris\\
University Paris-Est, LIGM\\
Ecole des Ponts ParisTech\\
{\tt\small spyros.gidaris@imagine.enpc.fr}
\and
Nikos Komodakis\\
University Paris-Est, LIGM\\
Ecole des Ponts ParisTech\\
{\tt\small nikos.komodakis@enpc.fr}\\
}

\maketitle

\begin{abstract}
Pixel wise image labeling is an interesting and challenging problem with great significance in the computer vision community.
In order for a dense labeling algorithm to be able to achieve accurate and precise results, 
it has to consider the dependencies that exist in the joint space of both the input and the output variables.
An implicit approach for modeling those dependencies is by training 
a deep neural network that, given as input
an initial estimate of the output labels and the input image,
it will be able to predict a new refined estimate for the labels.
In this context, our work is concerned with what is the optimal architecture for performing the label improvement task. 
We argue that the prior approaches of 
either directly predicting new label estimates
or predicting residual corrections w.r.t. the initial labels 
with feed-forward deep network architectures are sub-optimal. 
Instead, we propose a generic architecture that decomposes the label improvement task to three steps:
1) \emph{detecting} the initial label estimates that are incorrect, 
2) \emph{replacing} the incorrect labels with new ones, and finally
3) \emph{refining} the renewed labels by predicting residual corrections w.r.t. them.
Furthermore, we explore and compare various other alternative architectures that consist of the aforementioned \emph{Detection}, \emph{Replace}, and \emph{Refine} components.
We extensively evaluate the examined architectures in the challenging task of dense disparity estimation (stereo matching) and we report both quantitative and qualitative results on three different datasets.
Finally, our dense disparity estimation network that implements the proposed generic architecture, 
achieves state-of-the-art results in the KITTI 2015 test surpassing  prior approaches by a significant margin\footnote{We plan to release the source code that trains and tests all architectures examined in this work. For the implementation we have used the Torch framework~\cite{collobert2011torch7}.}
.
\end{abstract}

\section{Introduction}

Dense image labeling is a problem of paramount importance in the computer vision community as it encompasses many 
low or high level vision tasks including  stereo matching~\cite{zbontar2015computing}, optical flow~\cite{horn1981determining}, surface normals estimation~\cite{eigen2015predicting}, and semantic segmentation~\cite{long2015fully}, to mention a few characteristic examples.  
In all these cases the goal is to assign a discrete or continuous value for each pixel in the image. Due to its importance, there is a vast amount of work on this problem. Recent methods can be roughly divided  into   three main classes of approaches.

The first class  focuses on developing
 independent patch classifiers/regressors
~\cite{shotton2009textonboost, shotton2008semantic, shotton2013real, long2015fully, fischer2015flownet, mayer2015large, noh2015learning} that would directly predict the pixel label given as input an image patch centered on it
or, in cases like stereo matching and optical flow, would be used for comparing  patches between  different images in order to pick  pairs of best matching pixels ~\cite{luo2016efficient, zagoruyko2015learning, zbontar2015computing, vzbontar2016stereo}.
Deep convolutional neural networks (DCNNs)~\cite{lecun1998gradient} have demonstrated excellent performance in the aforementioned tasks thanks to their ability to learn complex image representations by harnessing vast amount of training data~\cite{krizhevsky2012imagenet, simonyan2014very, he2015deep}. 
However, despite their great representational power, just applying DCNNs on image patches,
does not capture the structure of output labels, which is an important aspect of dense image labeling tasks. 
For instance, independent feed-forward DCNN patch predictors do not take into consideration the correlations that exist between nearby pixel labels. 
In addition, 
feed-forward DCNNs have the extra disadvantages  that
they usually involve multiple consecutive down-sampling operations (i.e. max-pooling or strided convolutions) and that the top most convolutional layers do not capture factors such as image edges or other fine image structures. 
Both of the above properties  may prevent such methods from achieving precise and accurate results in dense image labeling tasks. 

Another class of methods tries  to model the joint dependencies of both the input and output variables by  use of  probabilistic graphical models such as 
Conditional Random Fields (CRFs)~\cite{lafferty2001conditional}. 
In CRFs, the dense image labeling task is performed through maximum a posteriori (MAP) inference in a graphical model that incorporates 
prior knowledge about the nature of the task in hand with pairwise edge potential between the graph nodes of the label variables. 
For example, in the case of semantic segmentation, those pairwise potentials enforce label consistency among similar or spatially adjacent pixels.
Thanks to their ability to jointly model the input-output variables, 
CRFs have been extensively used in pixel-wise image labelling tasks~\cite{koltun2011efficient,russell2009associative}. 
Recently, a number of methods has attempted to combine them with the representational power of DCNNs by getting 
the former (CRFs) to refine and disambiguate the predictions of the later one~\cite{schwing2015fully,chen2014semantic,zheng2015conditional,chen2015learning}. 
Particularly, in semantic segmentation, DeepLab~\cite{chen2014semantic} uses a fully connected CRF to post-process the pixel-wise predictions of a convolutional neural network 
while 
in CRF-RNN~\cite{zheng2015conditional}, they unify the training of both the DCNN and the CRF by formulating the approximate mean-field inference of fully connected CRFs as Recurrent Neural Networks (RNN). 
However, a major drawback of most CRF based approaches is that the pairwise potentials have to be carefully hand designed in order to incorporate simple human assumptions about the structure of the output labels $Y$ and at the same time to allow for tractable inference. 

A third class of methods
relies on a more data-driven approach for learning the joint space of both the input and the output variables.
More specifically, in this {case} a deep neural network gets as input
an initial estimate of the output labels and (optionally) the input image and it is trained to predict a new refined estimate for the labels, thus being implicitly enforced to learn the joint space of both the input and the output variables. The network can learn either to predict new estimates for all  pixel labels ~(transform-based approaches) \cite{yu2015multi,havaei2016brain,li2015iterative},
or alternatively,  to predict residual corrections w.r.t. the initial label estimates (residual-based approaches)~\cite{carreira2015human}. 
We will hereafter refer to these methods as \emph{deep joint input-output models}. These are, loosely speaking, related to the CRF models in the sense that the deep neural network is enforced to learn the joint dependencies of both the input image and output labels, but with the advantage of being less constrained about the complexity of the input-output dependencies that it can capture. 

\begin{figure}[t]
\centering
\renewcommand{\figurename}{Figure}
\renewcommand{\captionlabelfont}{\bf}
\renewcommand{\captionfont}{\small} 
\centering
\includegraphics[width=0.48\textwidth]{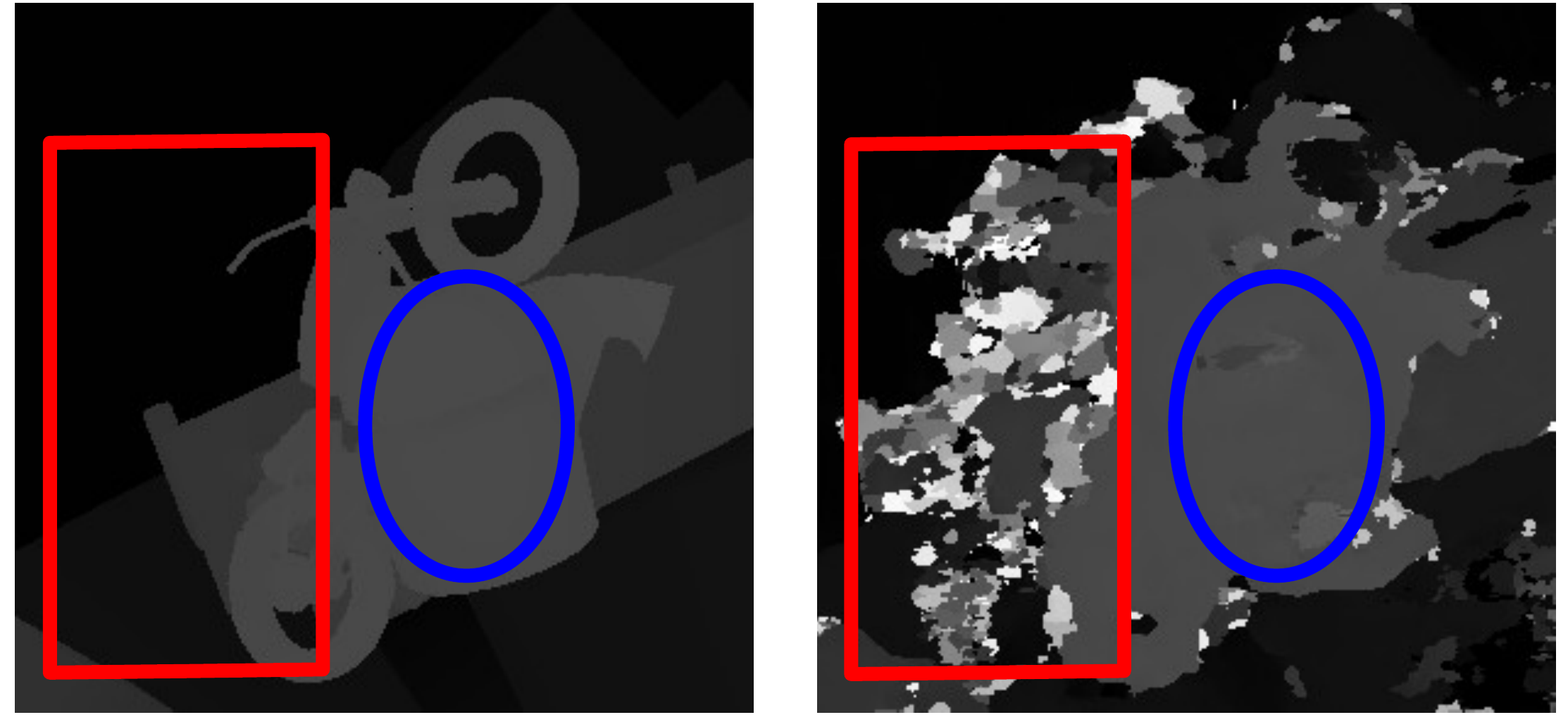}
\vspace{-8pt}
\caption{
In this figure we visualize two different type of erroneously labeled image regions.
On the left hand are the ground truth labels and on the right hand are some initial label estimates.
With the red rectangle we indicate a dense concentration of "hard" mistakes in the initial labels that it is very difficult to be corrected by a residual refinement component. 
Instead, the most suitable action for such a region is to replace them by predicting entirely new labels for them.
In contrast, the blue eclipse indicates an image region with "soft" label mistakes. Those image regions are easier to be handled by a residual refinement components.}
\vspace{-3pt}
\label{fig:errors}        
\end{figure}

Our work {belongs to this} last category of dense image labeling {approaches, thus} it is not constrained on the complexity of the input-output dependencies that it can capture. 
However, {here we argue that  prior approaches in this category use a sub-optimal strategy.}
For instance, the transform-based approaches (that always learn to predict new label estimates) often have to learn something more difficult than  necessary since they must often simply learn to operate as  identity transforms in case of  correct initial labels, yielding  the same label in their output. 
On the other hand, for the residual based approaches it is easier to learn to predict zero residuals in the case of correct initial labels, but it is more difficult for them to refine ``hard'' mistakes that deviate a lot from  the initial labels (see figure~\ref{fig:errors}).
Due to the above reasons, 
in our work we propose  a deep joint input-output model that decomposes the label estimation/refinement process as a sequence of the following easier to execute operations:
1)   \emph{detection} of errors in the input labels,
2)  \emph{replacement} of the erroneous labels with new ones, and finally
3)  an  overall \emph{refinement} of all output labels in the form of residual corrections. Each of the described operations  in our framework is executed by a different component implemented with a deep neural network.
Even more, 
those components are embedded in a unified architecture that is fully differentiable thus allowing for an end-to-end learning of the dense image labeling task
by only applying the objective function on the final output.
As a result of this, we are also able    to  explore a variety of novel  deep network architectures by considering  different ways of combining the above components, including the possibility of performing the above operations  iteratively, as it is done in~\cite{li2015iterative}, thus enabling our model to correct even large, in area,  regions of incorrect labels.
It is also worth noting that 
the error detection component in the proposed architecture,
by being forced to detect the erroneous pixel labels (given both the input and the initial estimates of the output labels), it implicitly learns the joint structure of the input-output space, which is  an important requirement for  a  successful application of any type of structured prediction model.

To summarize, our contributions are as follows:

\begin{itemize}
\item We propose a  deep structured prediction framework for the dense image labeling task, which we call \emph{Detect, Replace, Refine}, that relies on three main building blocks: 1) recognizing errors in the input label maps, 2) replacing the erroneous labels, and 3) performing a final refinement of the output label map.
We  show that all of the aforementioned steps can be embedded in a unified deep neural network architecture that is end-to-end trainable.
\item  In the context of the above framework, we    also explore a variety of   other network  architectures for deep joint input-output models that result from  utilizing different combinations of the above building blocks.\item We implemented and evaluated our framework  on the disparity prediction task (stereo matching) and we provide both qualitative and quantitative evidence about the  advantages of the proposed approach.

\item
We show that our disparity estimation model that implements the proposed \emph{Detect, Replace, Refine} architecture
achieves  state of the art results in the KITTI 2015 test set 
outperforming all prior published work by a significant margin. 
\end{itemize}

The remainder of the paper is structured as follows: 
We first describe our structured dense label prediction framework in \S\ref{sec:Methodology} and its implementation w.r.t. the dense disparity estimation task (stereo matching) in \S\ref{sec:DisparityEstimaton}.
Then, we provide experimental results in \S\ref{sec:ExperimentalResults} and  we finally conclude the paper  in \S\ref{sec:Conclusions}.

\section{Methodology} \label{sec:Methodology} 

Let $X=\{x_i\}_{i=1}^{H \times W}$ be the input image\footnote{Here, for simplicity, we consider images defined on a 2D domain, but our framework can be readily applied to images defined on any domain.} of size $H \times W$, where $x_i$ are the image pixels, and $Y=\{y_i\}_{i=1}^{H \times W}$ be some initial label estimates for this image, where $y_i$ is the label for the i-th pixel.
Our dense image labeling methodology belongs on the broader category of approaches that
consist of a deep joint input-output model model $F(.)$ that given as input the image $X$ and the initial labels $Y$, 
it learns to predict new, more accurate labels $Y' = F(X,Y)$. 
Note that in this setting the initial labels $Y$ could come from another model $F_0(.)$ that depends only on the image $X$.\
Also, in the general case, 
the pixel labels $Y$ can be of either discrete or continuous nature. 
In this work, however, we focus on the continuous case where greater variety of architectures can be explored.

The crucial question is what is the most effective way of implementing the deep joint input-output model $F(.)$.
The two most common approaches in the literature 
involve a feed-forward deep convolutional neural network, $F_{DCNN}(.)$, that
either directly predicts new labels $Y' = F_{DCNN}(X,Y)$ or it predicts the residual correction w.r.t. the input labels: $Y' = Y + F_{DCNN}(X,Y)$. 
We argue that both of them are sub-optimal solutions for implementing the $F(.)$ model. 
Instead, in our work we opt for a decomposition of the task of model $F(.)$ (\ie predicting new, more accurate labels $Y'$) in three different sub-tasks that are executed in sequence.

In the remainder of this section, 
we first describe the proposed architecture in~\S\ref{sec:arch},
then we discuss the intuition behind it and its advantages in~\S\ref{sec:discuss}, 
and finally we describe other alternative architectures that we explored in ~\S\ref{sec:exparch}.

\subsection{Detect, Replace, Refine architecture} \label{sec:arch}

\begin{figure*}
\renewcommand{\figurename}{Figure}
\renewcommand{\captionlabelfont}{\bf}
\renewcommand{\captionfont}{\small} 
\begin{center}
\includegraphics[width=\textwidth]{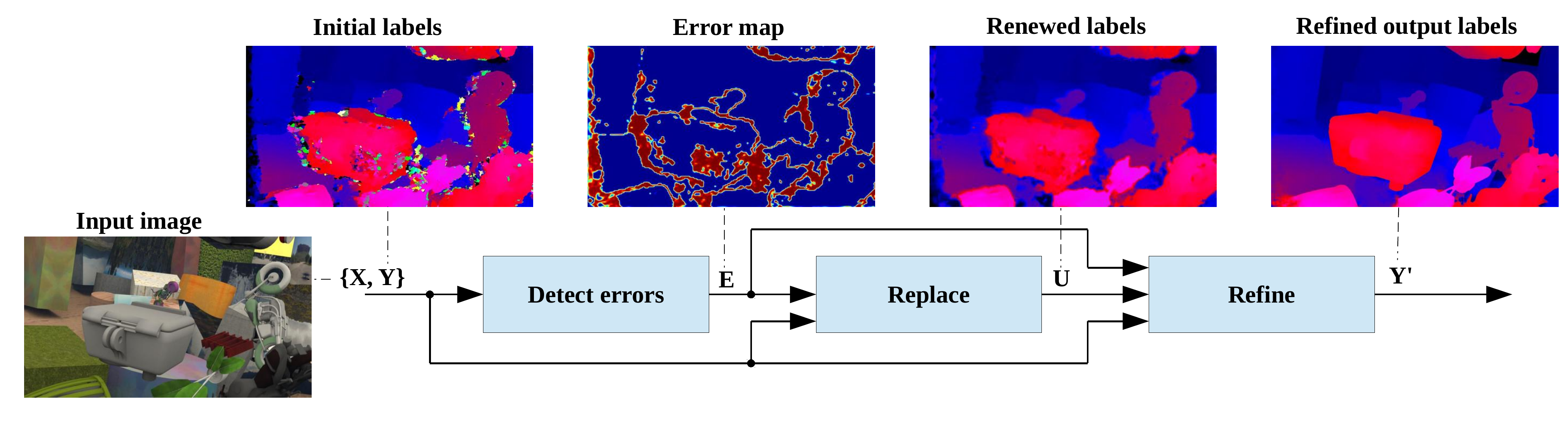}
\end{center}
\vspace{-13pt}
\caption{\small{
In this figure we demonstrate the generic architecture that we propose for the dense image labeling task.
In this architecture the task of the deep joint input-output model is decomposed into three different sub-tasks that are: 1) detection of the erroneous initial labels (based on an estimated error map $E$)      , 2) replacement of the erroneous labels with new ones (leading to a renewed label map $U$), and then 3) refinement $Y'$ of the renewed label map. 
The illustrated example is coming from the dense disparity labeling task (stereo matching).
 }}
\label{fig:architecture}
\vspace{-3pt}
\end{figure*}

The generic dense image labeling architecture that we propose decomposes task of the deep joint input-output model in three sub-tasks each of them handled by a different learn-able network component (see Figure~\ref{fig:architecture}).
Those network components are:
the error detection component $F_e(.)$,
the label replacement component $F_u(.)$, and
the label refinement component $F_r(.)$.
The sub-tasks that they perform, are:

\begin{description}
\item[Detect:]
The first sub-task in our generic pipeline  
is to detect the erroneously labeled pixels of $Y$ by discovering which pixel labels are inconsistent with the remaining labels of $Y$ and the input image $X$. 
This sub-task is performed by the error detection component $F_e(.)$ that basically needs to yield
a probability map $E = F_e(X,Y)$ of the same size as the input labels $Y$ that will have high probabilities for the "hard" mistakes in $Y$. These mistakes should ideally  be forgotten and replaced with entirely new label values in the processing step that follows (see Figures \ref{fig:EMAPs}a, \ref{fig:EMAPs}b, and \ref{fig:EMAPs}c). 
As we will see below, the topology of our generic architecture allows the error detection component $F_e(.)$ to learn its assigned task (\ie detecting the incorrect pixel labels) without explicitly being trained for this, e.g., through the use of an auxiliary loss.
The error detection function $F_e(.)$ can be implemented with any deep (or shallow) neural network with the only constraint being that its output map $E$ must take values in the range $[0, 1]$.

\item[Replace:] 
In the second sub-task, a new label field $U$ is produced by the convex combination of the initial label field $Y$ and the output of the label replacement component $F_u(.)$:
$U = E \odot F_u(X, Y, E) + (1-E) \odot Y$ (see Figures \ref{fig:EMAPs}e and \ref{fig:EMAPs}f).
We observe that the error probabilities generated by the error detection component $F_e(.)$
now act as gates that control which pixel labels of $Y$ will be 
forgotten and replaced by the outputs of $F_u(.)$,
which will  be all  pixel labels that are assigned high probability of being incorrect.  
In this context, the task of the Replace component $F_u(.)$ is to replace the 
erroneous
pixel labels
with new ones that will be in accordance both w.r.t. the input image $X$ and w.r.t. the non-erroneous labels of $Y$.
Note that for this task the Replace component $F_u(.)$ gets as input also the error probability map $E$. The reason for doing this is to help the Replace component to focus its attention only on those image regions that their labels need to be replaced.
The  component $F_u(.)$ can be implemented by any neural network that its output has the same size as the input labels $Y$. 

\item[Refine:] 
The purpose of the erroneous label detection and label replacement 
steps so far was to perform a crude ``fix" of the ``hard" mistakes in the label map $Y$.
In contrast, the purpose of the current step is to do a final refinement of the entire output label map $U$, which is produced by the previous steps, in the form of residual corrections: $Y' = U + F_r(X, Y, E, U)$ (see Figures \ref{fig:EMAPs}g and \ref{fig:EMAPs}h).
Intuitively, the purpose of this step is to correct the ``soft" mistakes of the label map $U$ and to better align the output labels $Y'$ with the fine structures in the image $X$.
The Refine component $F_r(.)$ can be implemented by any neural network that its output has the same size as the input labels $U$. 
\end{description}
The above three steps can be applied for more than one iterations which, as we will see later, allows our generic framework to recover a good estimate of the ground truth labels or, in worst case, to yield more plausible results even when the initial labels $Y$ are severely corrupted (see Figure~\ref{fig:DRRx2} in the experiments section~\S\ref{sec:zoom}).

To summarize, 
the workings of our dense labeling generic architecture can be concisely described by the iterative application of the following three equations:
\begin{equation} \label{eq:Fe}
E = F_e(X, Y),
\end{equation}
\begin{equation} \label{eq:Fu}
U = E \odot F_u(X, Y, E) + (1-E) \odot Y,
\end{equation}
\begin{equation} \label{eq:Fr}
Y' = U + F_r(X, Y, E, U).
\end{equation}
We observe that the above generic architecture is fully differentiable as long as the function components $F_e(.)$, $F_u(.)$, and $F_r(.)$ are also differentiable. 
Due to this fact, the overall proposed architecture is end-to-end learnable by directly applying an objective function (\eg Absolute Difference or Mean Square Error loss functions) on the final output label maps $Y'$.
 
\subsection{Discussion} \label{sec:discuss}

\textbf{Role of the Detection component $F_e(.)$ and its synergy with the Replace component $F_u(.)$:}
The error detection component $F_e(.)$ is a key element in our generic architecture and
its purpose is to indicate which are the image regions that their labels are incorrect.
This type of information is exploited in the next step of label replacement in two ways.
Firstly, the Replace component $F_u(.)$ that gets as input the error map $E$, which is generated by $F_e(.)$, is able to know which are the image regions that their labels needs to be replaced and thus it is able to focus its attention only on those image regions. 
At this point note that, 
in equation \ref{eq:Fu}, the error maps $E$, 
apart from being given as input attention maps to the Replace component $F_u(.)$, they also act as gates that control which way the information will flow both during the forward propagation and during the backward propagation. Specifically, during the forward propagation case, in the 
cases that the error map probabilities are either $0$ or $1$, it holds that:
\begin{equation}\label{eq:FeFuFWD}
U = 
\begin{cases}
        Y, &\text{if } F_e(X, Y) = \mathbf{0},\\
        F_u(X, Y, E), &\text{if } F_e(X, Y) = \mathbf{1},
\end{cases}
\end{equation}
which basically means that the Replace component $F_u(.)$ is being utilized mainly for the erroneously labelled image regions. 
Also, during the backward propagation, it is easy to see that the gradients of the replace function w.r.t. the loss $L$ 
(in the cases that the error probabilities are either $0$ or $1$) 
are: 
\begin{equation}\label{eq:FeFuBWD}
\frac{dL}{d F_u(.)} = 
\begin{cases}
        \mathbf{0}, &\text{if } F_e(X, Y) = \mathbf{0},\\
        \frac{dL}{d U}, &\text{if } F_e(X, Y) = \mathbf{1},
\end{cases}
\end{equation}
which 
means that gradients are back-propagated through the Replace component $F_u(.)$ only for the erroneously labelled image regions.
So, in a nutshell, during the learning procedure the Replace component $F_u(.)$ is explicitly trained to predict new values mainly for the erroneously labelled image regions. 
The second advantage of giving the error maps $E$ as input to the Replace component $F_u(.)$, is that this allows the Replace component  to know
which image regions contain ``trusted'' labels that can be used for 
providing information on how to fill the erroneously labelled regions.

\begin{figure*}[t]
\centering
\renewcommand{\figurename}{Figure}
\renewcommand{\captionlabelfont}{\bf}
\renewcommand{\captionfont}{\small} 
\centering
\begin{subfigure}[b]{\textwidth}
\center
        \begin{center}
        \begin{subfigure}[b]{0.24\textwidth}
        \includegraphics[width=\textwidth]{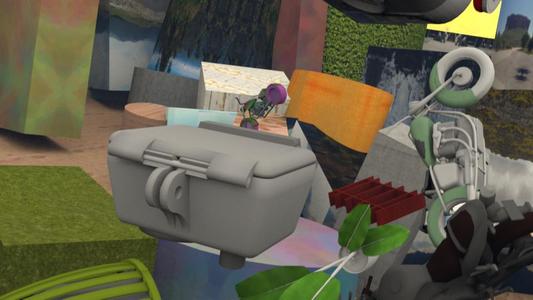}\\
        \vspace{-15pt}                
        \caption{\small{\textbf{Image $X$}}}
        \end{subfigure} 
        \hspace{0.001cm} 
        \begin{subfigure}[b]{0.24\textwidth}
        \includegraphics[width=\textwidth]{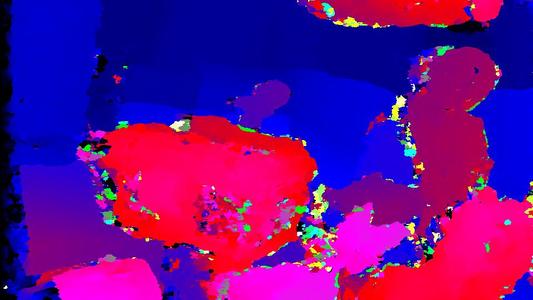}\\
        \vspace{-15pt}      
        \caption{\small{\textbf{Initial labels $Y$}}}
        \end{subfigure} 
        \hspace{0.001cm} 
        \begin{subfigure}[b]{0.24\textwidth}
        \includegraphics[width=\textwidth]{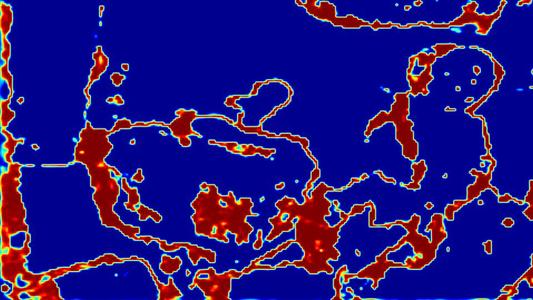}\\
        \vspace{-15pt}                      
        \caption{\small{\textbf{Predicted error map $E$}}}
        \end{subfigure} 
        \hspace{0.001cm} 
        \begin{subfigure}[b]{0.24\textwidth}
        \includegraphics[width=\textwidth]{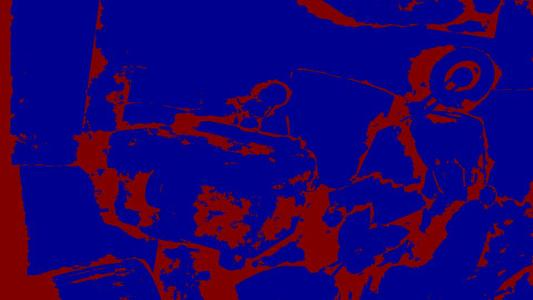}\\
        \vspace{-15pt}     
        \caption{\small{\textbf{Ground truth errors}}}
        \end{subfigure}         
        \end{center}
\end{subfigure}
\begin{subfigure}[b]{\textwidth}
\center
        \begin{center}
        \begin{subfigure}[b]{0.24\textwidth}
        \includegraphics[width=\textwidth]{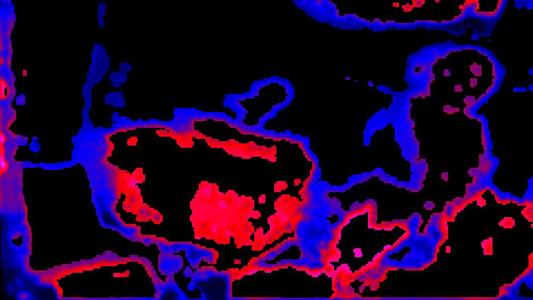}\\
        \vspace{-15pt}                
        \caption{\small{\textbf{$F_u(.)$ predictions}}}
        \end{subfigure} 
        \hspace{0.001cm} 
        \begin{subfigure}[b]{0.24\textwidth}
        \includegraphics[width=\textwidth]{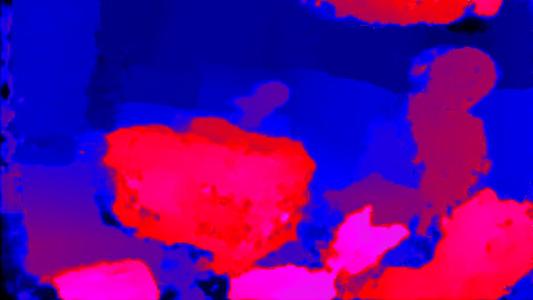}\\
        \vspace{-15pt}      
        \caption{\small{\textbf{Renewed labels $U$}}}
        \end{subfigure} 
        \hspace{0.001cm} 
        \begin{subfigure}[b]{0.24\textwidth}
        \includegraphics[width=\textwidth]{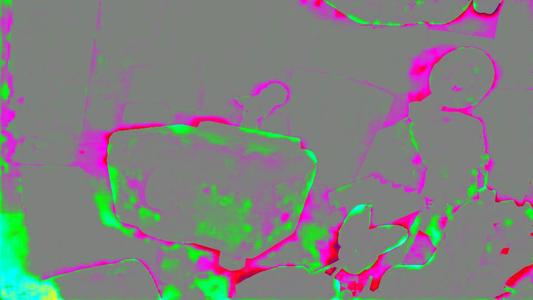}\\
        \vspace{-15pt}                      
        \caption{\small{\textbf{$F_r(.)$ residuals}}}
        \end{subfigure} 
        \hspace{0.001cm} 
        \begin{subfigure}[b]{0.24\textwidth}
        \includegraphics[width=\textwidth]{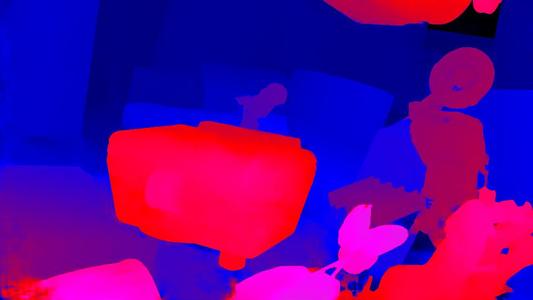}\\
        \vspace{-15pt}     
        \caption{\small{\textbf{Final labels $Y'$}}}
        \end{subfigure}         
        \end{center}
        \vspace{5pt}
\end{subfigure}
\vspace{-15pt}
\caption{
Here we provide an example that illustrates the functions performed by the Detect, Replace, and Refine steps in our proposed  architecture.
The example is coming from the dense disparity labeling task (stereo matching).
Specifically, subfigures \textbf{(a)}, \textbf{(b)}, and \textbf{(c)} depict respectively the input image $X$,
the initial disparity label estimates $Y$, and the error probability map $E$ that the detection component $F_e(.)$ yields for the initial labels $Y$.
Notice the high similarity of  map $E$ with the ground truth error map of the initial labels $Y$ depicted in subfigure \textbf{(d)}, where
the ground truth error map has been computed by thresholding the absolute difference of the initial labels $Y$ from the ground truth labels with a threshold of $3$ pixels (red are the erroneous pixel labels). 
In subfigure \textbf{(e)} we depict the label predictions of the Replace component $F_u(.)$.
For visualization purposes we only depict the $F_u(.)$ pixel predictions that will replace the initial labels that are incorrect (according to the detection component) by drawing the remaining ones (\ie those that their error probability is less than $0.5$) with black color.
In subfigure \textbf{(f)} we depict the renewed labels $U = E \odot F_u(X, Y, E) + (1-E) \odot Y$. 
In subfigure \textbf{(g)} we depict the residual corrections that the Refine component $F_r(.)$ yields for the renewed labels $U$. 
Finally, in the last subfigure \textbf{(h)} we depict the final label estimates $Y' = U + F_r(X,Y,E,U)$ that the Refine step yields.
}
\label{fig:EMAPs}  
\vspace{-3pt}      
\end{figure*}

\textbf{Estimated error probability maps by the Detection component $F_e(.)$:}
Thanks to the topology of our generic architecture,  
by optimizing the reconstruction of the ground truth labels $\hat{Y}$,
the error detection component $F_e(.)$ implicitly learns to act as a joint probability model 
for patches of $X$ and $Y$  centered on each pixel of the input image, assigning a high probability of error for patches that do not appear to belong to the joint input-output space $(X,Y)$.
In Figures~\ref{fig:EMAPs}c and \ref{fig:EMAPs}d we visualize the estimated by the Detection component $F_e(.)$ error maps and the ground truth error maps in the context of the disparity estimation task (more visualizations are provided in Figure~\ref{fig:VisEDC}).
It is interesting to note that the estimated error probability maps are very similar to the ground truth error maps despite the fact that we are not explicitly enforcing this behaviour, e.g., through the use of an auxiliary loss.

\textbf{Error detection component and Highway Networks:}
Note that the way the Detection component $F_e(.)$ and Replace component $F_u(.)$ interact bears some resemblance to the basic building blocks of the Highway Networks~\cite{srivastava2015highway}
that are being utilized for training extremely deep neural network architectures.
Briefly, each highway building block gets as input some hidden feature maps and then predicts transform gates that control which feature values will be carried on the next layer as is and which will be transformed by a non-linear function. 
There are however some important differences. 
For instance, in our case the error gate prediction and the label replacement steps are executed in sequence with the latter one getting as input the output of the former one. 
Instead, in Highway Networks the gate prediction and the non-linear transform of the input feature maps are performed in parallel.
Furthermore, in Highway Networks the components of each building block are implemented by simple affine transforms followed by non-linearities
and the purpose is to have multiple building blocks stacked one on top of the other in order to learn extremely deep image representations.
In contrast, the components of our generic architecture are them selves deep neural networks and the purpose is to learn to reconstruct the input labels $Y$.

\textbf{Two stage refinement approach:}
Another key element in our architecture is that the step of predicting new, more accurate labels $Y'$, given the initial labels $Y$, 
is broken in two stages. 
The first stage is handled by the error detection component $F_e(.)$ and the label replacement component $F_u(.)$. Their job is to correct only the "hard" mistakes of the input labels $Y$. 
They are not meant to correct "soft" mistakes (\ie errors in the label values of small magnitude).
In order to learn to correct those "soft" mistakes, it is more appropriate to use a component that yields residual corrections w.r.t. its input.
This is the purpose of our Refine component $F_r(.)$, in the second stage of our architecture, from which we expect to improve the "details" of the output labels $U$ by better aligning them with the fine structures of the input images.
This separation of roles between the first and the second refinement stages (\ie coarse refinement and then fine-detail refinement)
has the potential advantage, which is exploited in our work, to perform the actions of the first stage in lower resolution thus speeding up the processing and reducing the memory footprint of the network.
Also, the end-to-end training procedure
allows the components in the first stage (\ie $F_e(.)$ and $F_u(.)$)
to make mistakes as long as those are corrected by the second stage. 
This aspect of our architecture has the advantage that each component can more efficiently exploit its available capacity.

\subsection{Explored architectures} \label{sec:exparch}
In order to evaluate the proposed architecture we also devised and tested various others architectures that consist of the same core components as those that we propose. 
In total, the architectures that are explored in our work are:

\textbf{\emph{Detect + Replace + Refine} architecture:} This is the architecture that we proposed in section~\ref{sec:arch}.

\textbf{\emph{Replace} baseline architecture:}
In this case the model directly replaces the old labels with new ones: $Y' = Fu(X,Y)$.

\textbf{\emph{Refine} baseline architecture:}
In this case the model predicts residual corrections w.r.t. the input labels: $Y' = Y + Fr(X,Y)$.

\textbf{\emph{Replace + Refine} architecture:}
Here the model first replaces the entire label map $Y$ with new values $U = Fu(X,Y)$ and 
then residual corrections are predicted w.r.t. the updated values $U$, $Y' = U + Fr(X,Y,U)$.

\textbf{\emph{Detect + Replace} architecture:} 
Here the model first detects errors on the input label maps $E = F_e(X, Y)$ and then replace those erroneous pixel labels $Y' = E \odot F_u(X, Y, E) + (1-E) \odot Y$.

\textbf{\emph{Detect + Refine} architecture:}
In this case, after the detection of the errors $E = F_e(X, Y)$, the erroneous pixel labels are masked out by setting them to the mean label value $l_{mu}$, $U = E \odot l_{mu} + (1-E) \odot Y$.  Then the masked label maps are given as input to a residual refinement model $Y' = U + F_r(X, Y, E, U)$. 
Note that this 
architecture can also be considered as a specific instance of the general  Detect + Replace + Refine  architecture where the Replace component $F_u(.)$ does not have any learnable parameters and constantly returns the mean label value, \ie, $F_u(.) = l_{mu}$.

\textbf{\emph{Parallel} architecture:}
Here, after the detection of the errors, the erroneous labels are replaced by the Replace component $F_u(.)$ while the rest labels are refined by the Refine component $F_r(.)$. More specifically, the operations performed by this architecture are described by the following equations:
\begin{equation} \label{eq:Fe}
E = F_e(X, Y), 
\end{equation}
\begin{equation} \label{eq:Fu}
U_1 = F_u(X,Y,E), U_2 = Y + F_r(X,Y,E),
\end{equation}
\begin{equation} \label{eq:Fr}
Y' = E \odot U_1 + (1-E) \odot U_2.
\end{equation}
Basically, in this architecture the components $F_u(.)$ and $F_r(.)$ are applied in parallel instead of the sequential topology that is chosen in the Detect + Replace + Refine architecture.

\textbf{\emph{Detect + Replace + Refine $\times T$}:} This is basically the Detect + Replace + Refine architecture  but  applied iteratively for $T$ iterations. 
Note that the model implementing this architecture is trained in a multi-iteration manner.

\textbf{\emph{X-Blind Detect + Replace + Refine} architecture:}
This is a "blind" w.r.t. the image $X$ version of the \emph{Detect + Replace + Refine} architecture.
Specifically, the "X-Blind" architecture is exactly the same as the proposed \emph{Detect + Replace + Refine} architecture with the only difference being that it gets as input  only the initial labels $Y$ and not the image $X$ (\ie none of the $F_e(.)$, $F_u(.)$, and $F_r(.)$ components depends on the image $X$).
Hence, the model implemented by  the "X-Blind" architecture  must learn to reconstruct the ground truth labels by only "seeing" a corrupted version of them.

\section{Detect, Replace, Refine for disparity estimation} \label{sec:DisparityEstimaton} 
In order to evaluate the proposed dense image labeling architecture, as well as the other alternative architectures that are explored in our work, we use  the dense disparity estimation (stereo matching) task, according to which, given a left and right image, one needs to assign to 
each pixel of the left image  a continuous label that indicates its horizontal displacement in the right image (disparity).
Such a task forms a very interesting and challenging testbed for the evaluation of dense labeling algorithms since it requires dealing with several challenges such as accurately preserving disparity discontinuities  across object boundaries, dealing with occlusions,  as well as recovering the fine details of disparity maps. At the same time it has many practical applications on various autonomous driving and robot navigation or grasping tasks. 

\subsection{Initial disparities}

\textbf{Generating initial disparity field:}
In all the examined architectures, in order to generate the initial disparity labels $Y$ we used the deep patch matching approach that was proposed by W. Luo \etal~\cite{luo2016efficient} and specifically their architecture with id $37$.
We then train our models to reconstruct the ground truth labels given as input only the left image $X$ and the initial disparity labels $Y$. We would like to stress out that the right image of the stereo pair is not provided to our models. 
This practically means that the trained models cannot rely only on the image evidence for performing the dense disparity labelling task -- since disparity prediction from a single image is an ill-posed problem -- but they have to learn the joint space of both input $X$ and output labels $Y$ in order to perform the task.

\textbf{Image \& disparity field normalization:}
Before we feed an image and its initial disparity field to any of our examined architectures, 
we normalize them to zero mean and unit variance (\ie mean subtraction and division by the standard deviation). The mean and standard deviation values of the RGB colors and disparity labels are computed on the entire training set. 
The disparity target labels are also normalized with the same mean and standard deviation values and during inference the normalization effect is inverted on the disparity fields predicted by the examined architectures.

\subsection{Deep neural network architectures}
Each component of our generic architecture can be implemented by a deep neural network. 
For our disparity estimation experiments we chose the following implementations:

\textbf{Error detection component:} 
It is implemented by 5 convolutional layers of which the last one yields the error probability map $E$. 
All the convolutional layers, apart from the last one, are followed by batch normalization~\cite{ioffe2015batch} plus ReLU~\cite{maas2013rectifier} units. Instead, the last convolutional layer is followed by a sigmoid unit. 
The first two convolutions are followed by max-pooling layers of kernel size 2 that in total reduce the input resolution by a factor of 4.
To compensate, a bi-linear up-sampling layer is placed on top of the last convolution layer in order the output probability map $E$ to have the same resolution as the input image. 
The number of output feature planes of each of the 5 convolutional layers is $32$, $64$, $128$, $256$, and $1$ correspondingly.

\textbf{Replace component:}
It is implemented with a convolutional architecture that first "compress" the resolution of the feature maps to $\frac{1}{64}$ of the input resolution and then "decompress" the resolution to $\frac{1}{4}$ of the input resolution. 
For its implementation we follow the guidelines of A. Newel \etal \cite{newell2016stacked} which
are to use residual blocks~\cite{he2015deep} on each layer and parametrized (by residual blocks) skip connection between the symmetric layers in the "compressing" and the "decompressing" parts of the architecture. 
The "compressing" part of the architecture uses
max-pooling layers with kernel size 2 to down-sample the resolution
while 
the "decompressing" part uses nearest-neighbor up-sampling (by a factor of 2).
We refer for more details to A. Newel \etal \cite{newell2016stacked}.
In our case, during the "compression" part there are in total 6 down-sampling convolutional blocks and during the "decompression" part 4 up-sampling convolutional blocks. 
The number of output feature planes in the first layer is $32$ and each time the resolution is down-sampled the number of feature planes is increased by a factor of $2$. 
For GPU memory efficiency reasons, 
we do not allow the number of output feature planes of any layer to exceed that of $512$.
During the "decompression" part, each time we up-sample the resolution we also decrease by a factor of 2 the number of feature planes. The last convolution layer yields a single feature plane with the new disparity labels (without any non-linearity).
As already explained, during the "decompressing" part the resolution is increased till that of $\frac{1}{4}$ of the input resolution. The reason for early-stopping the "decompression" is that the Replace component is needed to only perform crude "fixes" of the initial labels and thus further "decompression" steps are not necessary.
Before the disparity labels are fed to the next processing steps, bi-linear up-sampling by a factor of 4 (without any learn-able parameter) is being used in order to restore the resolution to that of the input resolution. 

\textbf{Refine component:} 
It follows the same architecture as the replace component with the exception that during the "compressing" part the resolution of the feature maps is reduced till $\frac{1}{16}$ of the input resolution and then during the "decompressing" part the resolution is restored to that of the input resolution.  

\textbf{Alternative architectures:} 
In case the alternative architectures
have missing components, 
then the number of layers and/or the number of feature planes per layer of the remaining components is being increased such that the total capacity (\ie number of learn-able parameters) remains the same. 
For the architectures that include only the Replace or Refine components (\ie \emph{Replace}, \emph{Refine}, \emph{Detect+Replace}, and \emph{Detect+Refine} architectures) the "compression" - "decompression" architecture of this component "compresses" the resolution till $\frac{1}{64}$ of the input resolution and then "decompresses" it to the same resolution as the input image.

\textbf{Weight initialization:} 
In order to initialize the weights of each convolutional layer we use the initialization scheme proposed by K. He \etal~\cite{he2015delving}.

\subsection{Training details}
We used the $L1$ loss as objective function and the networks were optimized 
using the Adam~\cite{kingma2014adam} method with $\beta_1 = 0.9$ and $\beta_2 = 0.99$.
The learning rate $lr$ was set to $10^{-3}$ and was decreased after 20 epochs to $10^{-4}$ and then after $15$ epochs to $10^{-5}$. We then continued optimizing for another $5$ epochs.
Each epoch lasted approximately $2000$ batch iterations where each batch consisted of $24$ training samples. 
Each training sample consists of patches with spatial size $256 \times 256$ and $4$ channels (3 RGB color channels + 1 initial disparity label channel). 
The patches are generated by randomly cropping with uniform distribution an image and its corresponding initial disparity labels. 

\textbf{Augmentation:} 
During training we used horizontal flip augmentation and chromatic transformations such as color, contrast, and brightness transformations.

\section{Experimental results} \label{sec:ExperimentalResults}

In this section we present an exhaustive experimental evaluation of the proposed architecture as well as of the other explored architectures in the task of dense disparity estimation.
Specifically, 
we first describe the evaluation settings  used in our experiments (section \ref{sec:expsetting}),
 then we  report  detailed quantitative results w.r.t. the examined architectures (section \ref{sec:quantresults}), 
and finally we provide qualitative results of the proposed \emph{Detect, Replace, Refine} architecture and all of its components, trying in this way to more clearly illustrate their role (section \ref{sec:qualresults}).

\subsection{Experimental settings} \label{sec:expsetting}

\textbf{Training set:} 
In order to train the explored architectures we used 
the large scale synthetic dataset for disparity estimation that was recently introduced by N. Mayer \etal~\cite{mayer2015large}. We call this dataset the Synthetic dataset. It consists of three different type of synthetic image sequences and includes around $34k$ stereo images. Also, we enriched this training set with $160$ images from the training set of the KITTI 2015 dataset~\cite{Menze2015CVPR,Menze2015ISA}\footnote{The entire training set of KITTI 2015 includes $200$ images. In our case we split those $200$ images in $160$ images that were used for training purposes and $40$ images that were used for validation purposes}.

\textbf{Evaluation sets:}
We evaluated our architectures on three different datasets.
On 2000 images from the test split of the Synthetic dataset,
on 40 validation images coming from KITTI 2015 training dataset, and on 15 images from the training set of the Middlebury dataset~\cite{scharstein2014high}. 
Prior to evaluating the explored architectures in the KITTI 2015 validation set, we fine-tuned the models that implement them only on the $160$ image of the KITTI 2015 training split.
In this case, we start training for $20$ epochs with a learning rate of $10^{-4}$,
we then reduce the learning rate to $10^{-5}$ and continue training for $15$ epochs, and then reduce again the learning rate to $10^{-6}$ and continue training for $5$ more epochs (in total $40$ epochs). 
The epoch size is set to $400$ batch iterations.

\textbf{Evaluation metrics:}
For evaluation we used the end-point-error (EPE), which is the averaged euclidean distance from the ground truth disparity, and the percentage of disparity estimates that their absolute difference from the ground truth disparity is more than $t$ pixels ($>$ $t$ pixel).
Those metrics are reported for the non-occluded pixels (Non-Occ), all the pixels (All), and only the occluded pixels (Occ).

\subsection{Quantitative results} \label{sec:quantresults}

\subsubsection{Disparity estimation performance}
\begin{table}
\centering
\renewcommand{\figurename}{Table}
\renewcommand{\captionlabelfont}{\bf}
\renewcommand{\captionfont}{\small} 
\resizebox{0.5\textwidth}{!}{
{\setlength{\extrarowheight}{2pt}\scriptsize
{\begin{tabular}{l <{\hspace{-0.3em}}||>{\hspace{-0.5em}} r || >{\hspace{-0.5em}} r || >{\hspace{-0.5em}} r || >{\hspace{-0.5em}} r || >{\hspace{-0.5em}} r  }
\hline
\multicolumn{1}{c||}{} & \multicolumn{1}{c||}{$>$ 2 pixel} & \multicolumn{1}{c||}{$>$ 3 pixel} & \multicolumn{1}{c||}{$>$ 4 pixel} & \multicolumn{1}{c||}{$>$ 5 pixel} & \multicolumn{1}{c}{EPE}\\
\hline
Architectures & All & All & All & All & All\\
\hline
Initial labels $Y$ & 24.3175 & 22.9004 & 21.9140 & 21.1680 &  12.0218\\
\hline
\multicolumn{6}{c}{Single-iteration results}\\
\hline
\emph{Replace} (baseline) & 12.8007 & 10.4512 & 8.8966 & 7.7467 & 2.4456\\
\hline
\emph{Refine} (baseline) & 14.5996 & 12.2246 & 10.3046 & 8.7873 & 2.1235\\
\hline
\emph{Replace + Refine} & 11.1152 & 9.1821 & 7.8430 & 6.8550 & 2.2356\\
\hline
\emph{Detect + Replace} & 11.6970 & 9.2419 & 7.6812 & 6.6018 & 2.1504\\
\hline
\emph{Detect + Refine} & 10.5309 & 8.5565 & 7.2154 & 6.2186 & 1.8210\\
\hline
\emph{Parallel} & 11.0146 & 8.9261 & 7.5029 & 6.4742 & 2.0241\\
\hline
\emph{Detect + Replace + Refine} & 9.5981 & 7.9764 & 6.7895 & 5.9074 & 1.8569\\ 
\hline
\multicolumn{6}{c}{Multi-iteration results}\\
\hline
\emph{Detect + Replace + Refine x2} & \textbf{8.8411} & \textbf{7.2187} & \textbf{6.0987} & \textbf{5.2853} & \textbf{1.6899}\\
\hline
\end{tabular}}}}
\vspace{3pt}
\caption{\small{Stereo matching results on the Synthetic dataset.}}
\label{tab:Synthetic}
\vspace{-7pt}
\end{table}
\begin{table*}
\centering
\renewcommand{\figurename}{Table}
\renewcommand{\captionlabelfont}{\bf}
\renewcommand{\captionfont}{\small} 
\resizebox{1.0\textwidth}{!}{
{\setlength{\extrarowheight}{2pt}\scriptsize
{\begin{tabular}{l <{\hspace{-0.3em}}||>{\hspace{-0.5em}} r | >{\hspace{-0.5em}} r | >{\hspace{-0.5em}} r || >{\hspace{-0.5em}} r | >{\hspace{-0.5em}} r | >{\hspace{-0.5em}} r || >{\hspace{-0.5em}} r | >{\hspace{-0.5em}} r | >{\hspace{-0.5em}} r || >{\hspace{-0.5em}} r | >{\hspace{-0.5em}} r | >{\hspace{-0.5em}} r || >{\hspace{-0.5em}} r | >{\hspace{-0.5em}} r | >{\hspace{-0.5em}} r }
\hline
\multicolumn{1}{c||}{} & \multicolumn{3}{c||}{$>$ 2 pixel} & \multicolumn{3}{c||}{$>$ 3 pixel} & \multicolumn{3}{c||}{$>$ 4 pixel} & \multicolumn{3}{c||}{$>$ 5 pixel} & \multicolumn{3}{c}{EPE}\\
\hline
Architectures & Non-Occ & All & Occ & Non-Occ & All & Occ & Non-Occ & All & Occ & Non-Occ & All & Occ & Non-Occ & All & Occ\\
\hline
Initial labels $Y$ & 18.243 & 26.714 & 86.125 & 15.664 & 23.986 & 82.330 & 14.208 & 22.282 & 78.758 & 13.237 & 21.044 & 75.579 & 6.058 & 8.709 & 25.598\\
\hline
\multicolumn{16}{c}{Single-iteration results}\\
\hline
\emph{Replace} (baseline) & 15.767 & 21.089 & 57.197 & 12.323 & 16.793 & 46.303 & 10.312 & 14.020 & 37.922 & 9.032 & 12.147 & 31.770 & 2.731 & 3.221 & 5.818\\
\hline
\emph{Refine} (baseline) & 13.981 & 19.742 & 58.039 & 11.110 & 16.042 & 47.732 & 9.266 & 13.406 & 39.218 & 7.889 & 11.392 & 32.467 & 1.953 & 2.551 & 5.665\\
\hline
\emph{Replace + Refine} & 14.262 & 19.257 & 52.036 & 11.297 & 15.701 & 43.905 & 9.552 & 13.459 & 37.910 & 8.408 & 11.891 & 33.125 & 2.292 & 2.908 & 6.216\\
\hline
\emph{Detect + Replace} & 15.368 & 20.984 & 58.745 & 11.243 & 16.169 & 48.568 & 8.957 & 13.176 & 40.663 & 7.571 & 11.179 & 34.482 & 2.013 & 2.676 & 6.462\\
\hline
\emph{Detect + Refine} & 13.732 & 19.375 & 56.383 & 10.718 & 15.552 & 46.281 & 8.893 & 12.975 & 38.197 & 7.600 & 11.012 & 31.478 & 2.105 & 2.626 & 5.389\\
\hline
\emph{Parallel} & 14.917 & 20.345 & 57.459 & 11.363 & 15.907 & 46.221 & 9.234 & 12.941 & 37.218 & 7.840 & 10.940 & 30.854 & 2.012 & 2.552 & 5.607\\
\hline
\emph{Detect + Replace + Refine} & 12.845 & 17.825 & 50.407 & 10.096 & 14.379 & 41.704 & 8.285 & 11.957 & 34.801 & 7.057 & 10.253 & 29.560 & \textbf{1.774} & 2.368 & 5.457\\
\hline
\multicolumn{16}{c}{Multi-iteration results}\\
\hline
\emph{Detect + Replace + Refine x2} & \textbf{11.529} & \textbf{16.414} & \textbf{47.922} & \textbf{8.757} & \textbf{12.874} & \textbf{37.977} & \textbf{6.997} & \textbf{10.482} & \textbf{30.634} & \textbf{5.911} & \textbf{8.916} & \textbf{25.514} & 1.789 & \textbf{2.321} & \textbf{4.971}\\
\hline
\end{tabular}}}}
\vspace{3pt}
\caption{\small{Stereo matching results on Middlebury.}}
\label{tab:Middleburry}
\vspace{-7pt}
\end{table*}
\begin{table*}
\centering
\renewcommand{\figurename}{Table}
\renewcommand{\captionlabelfont}{\bf}
\renewcommand{\captionfont}{\small} 
\resizebox{1.0\textwidth}{!}{
{\setlength{\extrarowheight}{2pt}\scriptsize
{\begin{tabular}{l <{\hspace{-0.3em}}||>{\hspace{-0.5em}} r | >{\hspace{-0.5em}} r | >{\hspace{-0.5em}} r || >{\hspace{-0.5em}} r | >{\hspace{-0.5em}} r | >{\hspace{-0.5em}} r || >{\hspace{-0.5em}} r | >{\hspace{-0.5em}} r | >{\hspace{-0.5em}} r || >{\hspace{-0.5em}} r | >{\hspace{-0.5em}} r | >{\hspace{-0.5em}} r || >{\hspace{-0.5em}} r | >{\hspace{-0.5em}} r | >{\hspace{-0.5em}} r }
\hline
\multicolumn{1}{c||}{} & \multicolumn{3}{c||}{$>$ 2 pixel} & \multicolumn{3}{c||}{$>$ 3 pixel} & \multicolumn{3}{c||}{$>$ 4 pixel} & \multicolumn{3}{c||}{$>$ 5 pixel} & \multicolumn{3}{c}{EPE}\\
\hline
Architectures & Non-Occ & All & Occ & Non-Occ & All & Occ & Non-Occ & All & Occ & Non-Occ & All & Occ & Non-Occ & All & Occ\\
\hline
Initial labels $Y$ & 8.831 & 10.649 & 98.098 & 6.412 & 8.253 & 96.559 & 5.222 & 7.059 & 94.742 & 4.514 & 6.339 & 93.139 & 1.700 & 2.457 & 31.214\\
\hline
\multicolumn{16}{c}{Single-iteration results}\\
\hline
\emph{Replace} (Baseline) & 4.997 & 5.668 & 37.327 & 3.329 & 3.888 & 27.890 & 2.452 & 2.892 & 19.643 & 1.924 & 2.292 & 15.226 & 0.858 & 0.923 & 3.165\\
\hline
\emph{Refine} (Baseline) & 4.429 & 5.165 & 33.028 & 3.075 & 3.714 & 25.107 & 2.370 & 2.924 & 19.610 & 1.933 & 2.404 & 15.978 & 0.867 & 0.953 & 3.384\\
\hline
\emph{Replace + Refine} & 3.963 & 4.529 & \textbf{27.411} & 2.712 & 3.209 & 21.465 & 2.082 & 2.507 & 16.481 & 1.735 & 2.098 & 13.611 & 0.802 & 0.865 & 2.859\\
\hline
\emph{Detect + Replace} & 5.126 & 5.751 & 35.554 & 3.469 & 4.005 & 27.656 & 2.517 & 2.953 & 20.519 & 1.911 & 2.269 & 15.947 & 0.886 & 0.943 & 3.108\\
\hline
\emph{Detect + Refine} & 4.482 & 5.169 & 34.992 & 3.054 & 3.634 & 26.453 & 2.328 & 2.799 & 19.004 & 1.865 & 2.258 & 14.686 & 0.863 & 0.926 & 2.952\\
\hline
\emph{Parallel} & 5.239 & 5.952 & 38.392 & 3.530 & 4.139 & 29.436 & 2.522 & 3.017 & 21.208 & 1.943 & 2.338 & 15.748 & 0.904 & 0.962 & 3.095\\
\hline
\emph{Detect + Replace + Refine} & 3.919 & 4.610 & 33.947 & 2.708 & 3.294 & 25.697 & 2.082 & 2.570 & 19.123 & 1.699 & 2.112 & 15.140 & 0.790 & 0.858 & 3.056\\
\hline
\multicolumn{16}{c}{Multi-iteration results}\\
\hline
\emph{Detect + Replace + Refine x2} & \textbf{3.685} & \textbf{4.277} & 28.164 & \textbf{2.577} & \textbf{3.075} & \textbf{20.762} & \textbf{2.001} &\textbf{ 2.424} & \textbf{16.086} & \textbf{1.652} & \textbf{2.004} & \textbf{13.056} & \textbf{0.779} & \textbf{0.835} & \textbf{2.723}\\
\hline
\end{tabular}}}}
\vspace{3pt}
\caption{\small{Stereo matching results on KITTI 2015 validation set.}}
\label{tab:Kitti2015}
\vspace{-7pt}
\end{table*}

In Tables~\ref{tab:Synthetic},~\ref{tab:Middleburry}, and~\ref{tab:Kitti2015} we report the stereo matching performance of the examined architectures in the Synthetic, Middlebury, and KITTI 2015 evaluation sets correspondingly. 

\textbf{Single-iteration results:}
We first evaluate all the examined architectures when they are applied for a single iteration.
We observe that all of them are able to improve the initial label estimates $Y$.
However, they do not all of them achieve it with the same success.
For instance, the baseline models \emph{Replace} and \emph{Refine} tend to be less accurate than the rest models. 
Compared to them, the \emph{Detect + Replace} and the \emph{Detect + Refine} architectures perform considerably better in two out of three datasets, the Synthetic and the Middlebury datasets. 
This improvement can only be attributed to the error detection step, which is what it distinguishes them from the baselines, and indicates the importance of having an error detection component in the dense labelling task. 
Overall, the best single-iteration performance is achieved by the \emph{Detect + Replace + Refine} architecture that we propose in this paper and combines both the merits of the error detection component and the two stage refinement strategy.
Compared to it, the \emph{Parallel} architecture has considerably worse performance, which indicates that the sequential order in the proposed architecture is important for achieving accurate results.  

\textbf{Multi-iteration results:}
We also evaluated our best performing architecture, 
which is the \emph{Detect + Replace + Refine} architecture that we propose, 
in the multiple iteration case.
Specifically, the last entry \emph{Detect + Replace + Refine x2} 
in Tables~\ref{tab:Synthetic},~\ref{tab:Middleburry}, and~\ref{tab:Kitti2015} 
indicates the results of the proposed architecture for 2 iterations
and we observe that it further improves the performance w.r.t. the single iteration case.
For more than 2 iterations we did not see any further improvement and for this reason we chose not to include those results.
Note that in order to train this two iterations model, 
we first pre-train the single iteration version and then fine-tune the two iterations version by adding the generated disparity labels from the first iteration in the training set.

\subsubsection{Label prediction accuracy Vs initial labels quality} \label{sec:analysis}
\begin{figure*}[t]
\center
\renewcommand{\figurename}{Figure}
\renewcommand{\captionlabelfont}{\bf}
\renewcommand{\captionfont}{\small} 
\begin{subfigure}[b]{0.45\textwidth}
\begin{center}
\includegraphics[width=\textwidth]{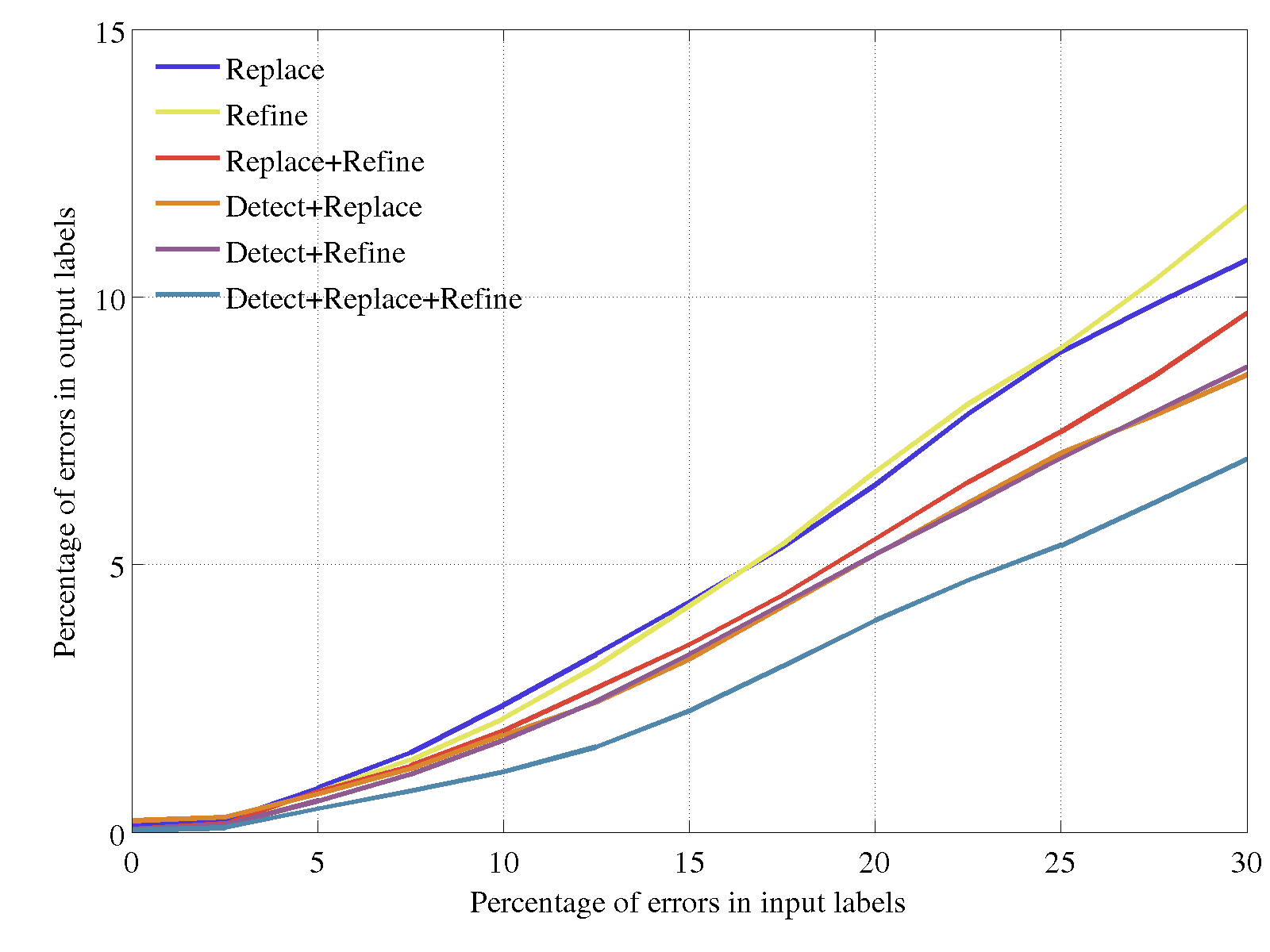} 
\end{center}
\vspace{-20pt}     
\caption{\small{\textbf{Error threshold $\tau = 3$ pixels}}}
\end{subfigure}
\begin{subfigure}[b]{0.45\textwidth}
\begin{center}
\includegraphics[width=\textwidth]{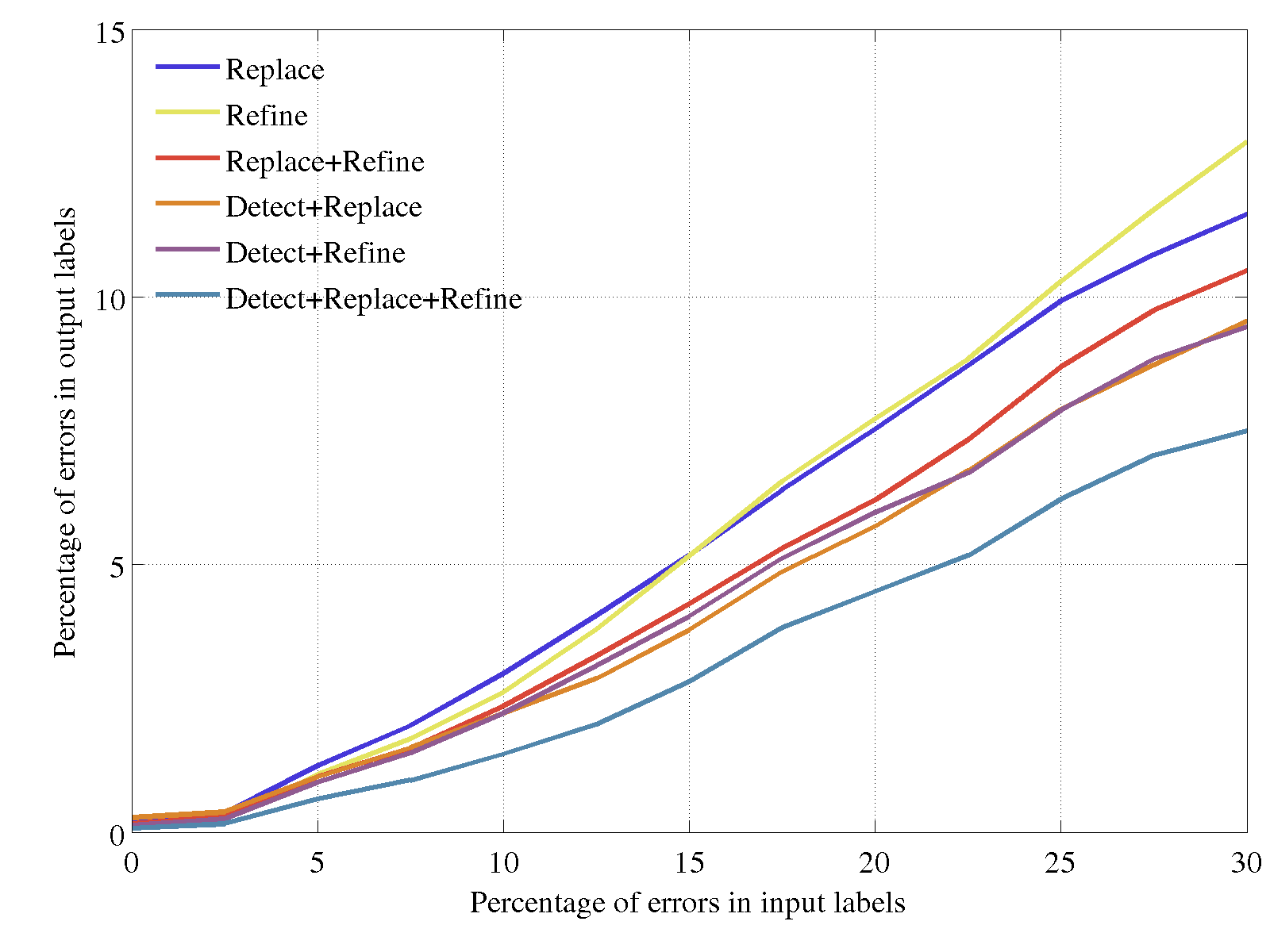}   
\end{center}
\vspace{-20pt}     
\caption{\small{\textbf{Error threshold $\tau = 5$ pixels}}}
\end{subfigure} \\
\begin{subfigure}[b]{0.45\textwidth}
\begin{center}
\includegraphics[width=\textwidth]{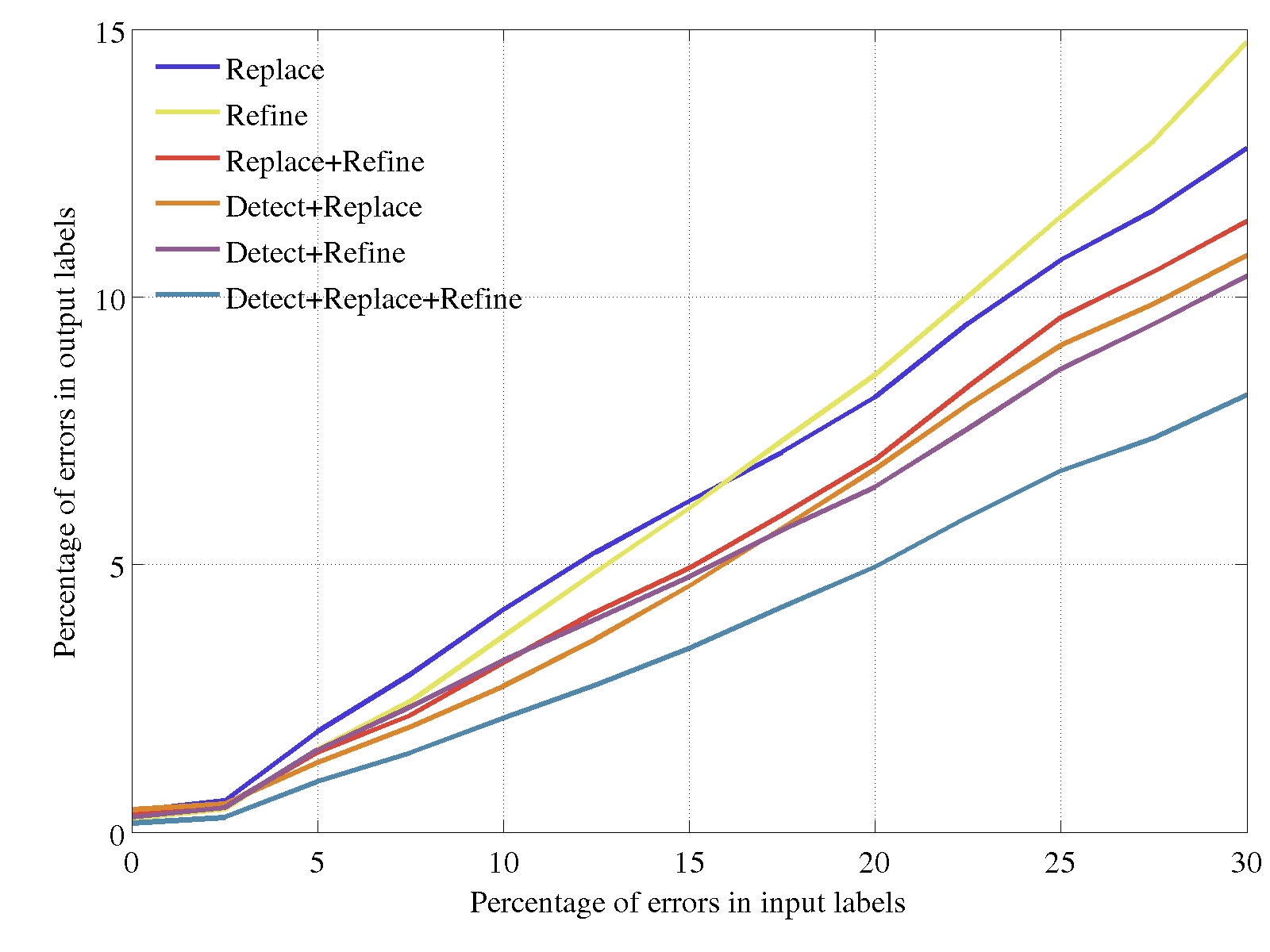} 
\end{center}
\vspace{-20pt}     
\caption{\small{\textbf{Error threshold $\tau = 8$ pixels}}}
\end{subfigure}
\begin{subfigure}[b]{0.45\textwidth}
\begin{center}
\includegraphics[width=\textwidth]{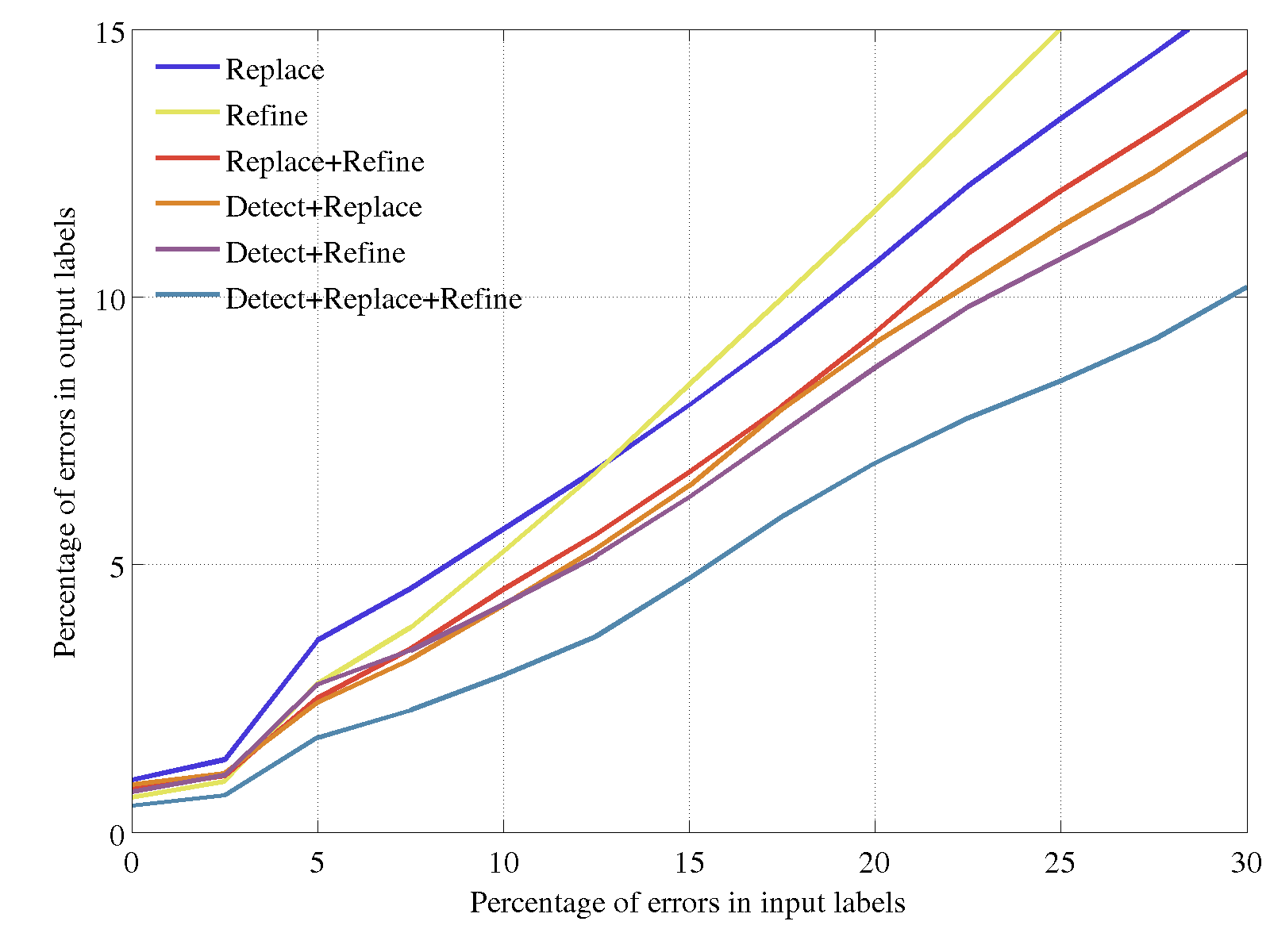}   
\end{center}
\vspace{-20pt}     
\caption{\small{\textbf{Error threshold $\tau = 15$ pixels}}}
\end{subfigure} \\
\vspace{5pt}
\caption{\small{
Percentage of erroneously estimated disparity labels for a pixel $x$
as a function of the percentage of erroneous initial disparity labels in the patch of size $w \times w$ centered on the pixel of interest $x$. 
The patch size $w$ is set to $65$. 
An estimated pixel label $y'$ is considered erroneous if its absolute difference from the ground truth label is more than $\tau_0=3$ pixels. 
For the initial disparity labels in each patch, 
the threshold $\tau$ of considering them incorrect is set to
\textbf{(a)} $3$ pixels, \textbf{(b)} $5$ pixels, \textbf{(c)} $8$ pixels, and 
\textbf{(d)} $15$ pixels. The evaluation is performed on $50$ images of the \emph{Synthetic} test set.}}
\label{tab:analysis}
\vspace{-7pt}
\end{figure*}

In Figure~\ref{tab:analysis}
we evaluate the ability of each architecture to predict the correct disparity label for each pixel $x$ as a function of the "quality" of the initial disparity labels in a $w \times w$ neighborhood of that pixel. To that end, we plot for each architecture
the percentage of erroneously estimated disparity labels as a function of the percentage of erroneous initial disparity labels that exist in the patch of size $w \times w$ centered on the pixel of interest $x$.  
In our case, the size of the neighborhood $w$ is set to $65$.
An estimated pixel label $y'$ for the pixel $x$ is considered erroneous if its absolute difference from the ground truth label is more than $\tau_0=3$ pixels. 
For the initial disparity labels in the patch centered on $x$, 
the threshold $\tau$ of considering them incorrect is set to 
$\tau = 3$ (Fig.~\ref{tab:analysis}.a),
$\tau = 5$ (Fig.~\ref{tab:analysis}.b),
$\tau = 8$ (Fig.~\ref{tab:analysis}.c), or
$\tau = 15$ (Fig.~\ref{tab:analysis}.d). 
We make the following observations (that are more clearly illustrated from sub-figures~\ref{tab:analysis}.c and~\ref{tab:analysis}.d):
\begin{itemize}
\item 
In the case of the \emph{Replace} and \emph{Refine} architectures, 
when the percentage of erroneous initial labels  is low (\eg less than $10\%$) then the \emph{Refine} architecture
(which predicts residual corrections)
is considerably more accurate than the \emph{Replace} architecture (which directly predicts new label values).
However, when the percentage of  erroneous initial labels is high (\eg more than $20\%$) then the \emph{Replace} architecture is more accurate than the \emph{Refine} one.
This observation supports our argument that residual corrections are more suitable for ``soft'' mistakes in the initial labels while predicting an entirely new label value is a better choice for the ``hard'' mistakes.
\item 
By introducing the error detection component,
both the \emph{Refine} and the \emph{Replace} architectures manage to significantly improve their predictions.
In the \emph{Detect+Refine} case, the improvement is due to the fact that the error detection component sets the ``hard'' mistakes to the mean label values (see the description of the \emph{Detect+Refine} architecture in the main paper) thus allowing the \emph{Refine} component to ignore the values of the ``hard'' mistakes of the initial labels  and instead make residual predictions w.r.t. the mean label values (these  mean values  are fixed and known in advance and thus it is easier for the network to learn to make residual predictions w.r.t. them).
In the case of the \emph{Detect+Replace} architecture,
the error detection component ``dictates" the \emph{Replace} component to predict new label values for the incorrect initial labels while allowing the propagation of the correct ones in the output.
\item Finally, the best \emph{"label prediction accuracy Vs initial labels quality"} behavior is achieved by the proposed \emph{Detect + Replace + Refine} architecture, which efficiently combines the error detection component with the two-stage label improvement approach.
Interestingly, the improvement margins w.r.t. the rest architectures is increased as the quality of the initial labels is decreased. 
\end{itemize}

\subsubsection{KITTI 2015 test set results}
\begin{table*}
\centering
\renewcommand{\figurename}{Table}
\renewcommand{\captionlabelfont}{\bf}
\renewcommand{\captionfont}{\small} 
\resizebox{0.9\textwidth}{!}{
{\setlength{\extrarowheight}{2pt}\scriptsize
{\begin{tabular}{l <{\hspace{-0.3em}}|>{\hspace{-0.5em}} r | >{\hspace{-0.5em}} r | >{\hspace{-0.5em}} r | >{\hspace{-0.5em}} r | >{\hspace{-0.5em}} r | >{\hspace{-0.5em}} r | >{\hspace{-0.5em}} r | >{\hspace{-0.5em}} r | >{\hspace{-0.5em}} r | >{\hspace{-0.5em}} r | >{\hspace{-0.5em}} r | >{\hspace{-0.5em}} r | >{\hspace{-0.5em}} r}
\hline
\multicolumn{1}{c|}{} & \multicolumn{3}{c|}{All / All} & \multicolumn{3}{c|}{All / Est} & \multicolumn{3}{c|}{Noc / All} & \multicolumn{3}{c}{Noc / Est} & Runtime \\
\hline
Architectures & D1-bg & D1-fg & D1-all & D1-bg & D1-fg & D1-all & D1-bg & D1-fg & D1-all & D1-bg & D1-fg & D1-all & (secs) \\
\hline
\emph{Ours} & \textbf{2.58} & 6.04 & \textbf{3.16} & \textbf{2.58} & 6.04 & \textbf{3.16} & 2.34 & 4.87 & \textbf{2.76} & 2.34 & 4.87 & \textbf{2.76} & 0.4 \\
\hline
DispNetC~\cite{mayer2015large} & 4.32 & \textbf{4.41} & 4.34 & 4.32 & \textbf{4.41} & 4.34 & 4.11 & \textbf{3.72} & 4.05 & 4.11 & \textbf{3.72} & 4.05 & 0.06\\
\hline
PBCB~\cite{Seki2016BMVC} & \textbf{2.58}        & 8.74 & 3.61 & \textbf{2.58} & 8.74 & 3.6 & \textbf{2.27} &          7.71 & 3.17 & \textbf{2.27} & 7.71 & 3.17 & 68\\
\hline
Displets v2~\cite{guney2015displets} & 3.00 & 5.56 & 3.43 & 3.00 & 5.56 & 3.43 & 2.73 & 4.95 & 3.09 & 2.73 & 4.95 & 3.09 & 265\\
\hline 
MC-CNN~\cite{vzbontar2016stereo} & 2.89 & 8.88 & 3.89 & 2.89 & 8.88 & 3.88 & 2.48 & 7.64 & 3.33 & 2.48 & 7.64 & 3.33 & 67\\
\hline
SPS-St~\cite{yamaguchi2014efficient} & 3.84 & 12.67 & 5.31 & 3.84 & 12.67 & 5.31 & 3.50 & 11.61 & 4.84 & 3.50 & 11.61 &  4.84 & 2\\
\hline
MBM~\cite{einecke2015multi} & 4.69 & 13.05 & 6.08 & 4.69 & 13.05 & 6.08 & 4.33 & 12.12 & 5.61 & 4.33 & 12.12 & 5.61 & 0.13\\
\hline
\end{tabular}}}}
\vspace{3pt}
\caption{\small{Stereo matching results on KITTI 2015 test set.}}
\label{tab:Kitti2015test}
\vspace{-7pt}
\end{table*}

We submitted our best solution, 
which is the proposed \emph{Detect + Replace + Refine} architecture applied for two iterations, 
on the KITTI 2015 test set evaluation server and we achieved state-of-the-art results in the main evaluation metric, D1-all, surpassing all prior work by a significant margin.
The results of our submission, as well as of other competing methods, are reported in Table~\ref{tab:Kitti2015test}\footnote{The link to our KITTI 2015 submission that contains more thorough test set results -- both qualitative and quantitative -- is:\\ 
{\url{http://www.cvlibs.net/datasets/kitti/eval_scene_flow_detail.php?benchmark=stereo&result=365eacbf1effa761ed07aaa674a9b61c60fe9300}}}.
Note that our improvement w.r.t. the best prior approach corresponds 
to a more than $10\%$ relative reduction of the error rate. 
Our total execution time is 0.4 secs, of which around 0.37 secs is used by the patch matching algorithm for generating the initial disparity labels and the rest 0.03 by our \emph{Detect + Replace + Refine x2} architecture (measured in a Titan X GPU).
For this submission, 
after having train the \emph{Detect + Replace + Refine x2} model on the training split (160 images),
we further fine-tuned it on both the training and the validation splits (in which we divided the 200 images of KITTI 2015 training dataset).
\subsubsection{"X-Blind" Detect + Replace + Refine architecture}
\begin{figure*}[t]
\centering
\renewcommand{\figurename}{Figure}
\renewcommand{\captionlabelfont}{\bf}
\renewcommand{\captionfont}{\small} 
\begin{subfigure}[b]{\textwidth}
\center
        \begin{center}
        \begin{subfigure}[b]{0.22\textwidth}
        \includegraphics[width=\textwidth]{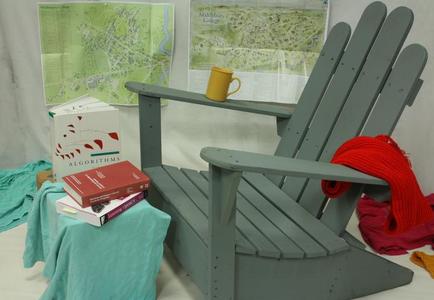}\\
        \vspace{-10pt}
        \includegraphics[width=\textwidth]{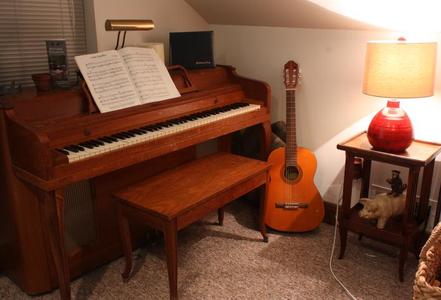}\\
        \vspace{-10pt}
        \includegraphics[width=\textwidth]{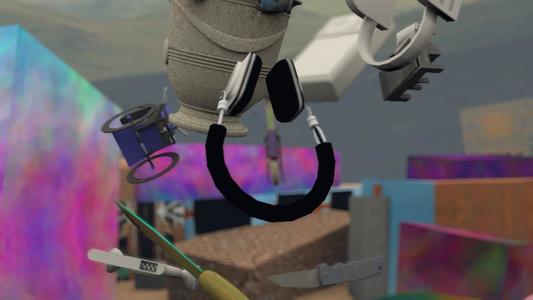}
        \vspace{-15pt}                
        \caption{\small{\textbf{Image $X$}}}
        \end{subfigure} 
        \hspace{0.001cm} 
        \begin{subfigure}[b]{0.22\textwidth}
        \includegraphics[width=\textwidth]{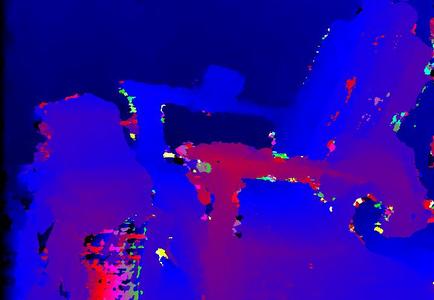}\\
        \vspace{-10pt}
        \includegraphics[width=\textwidth]{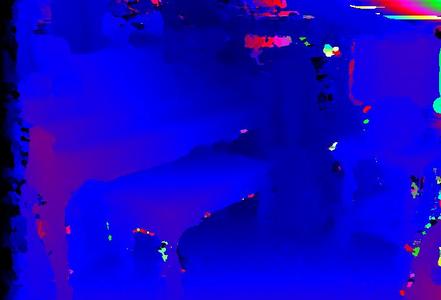}\\
        \vspace{-10pt}
        \includegraphics[width=\textwidth]{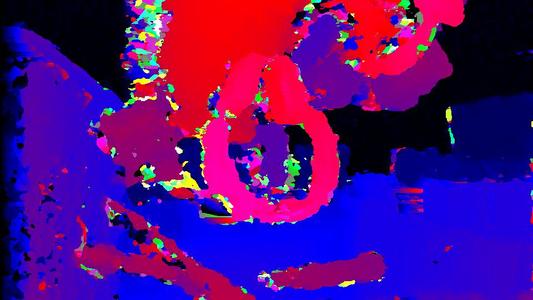}
        \vspace{-15pt}      
        \caption{\small{\textbf{Initial labels $Y$}}}
        \end{subfigure} 
        \hspace{0.001cm} 
        \begin{subfigure}[b]{0.22\textwidth}
        \includegraphics[width=\textwidth]{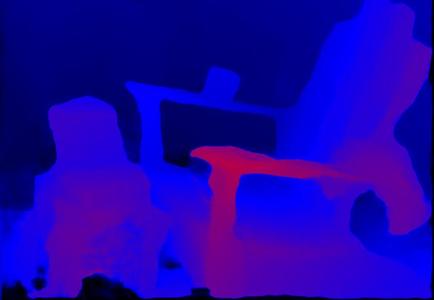}\\
        \vspace{-10pt}
        \includegraphics[width=\textwidth]{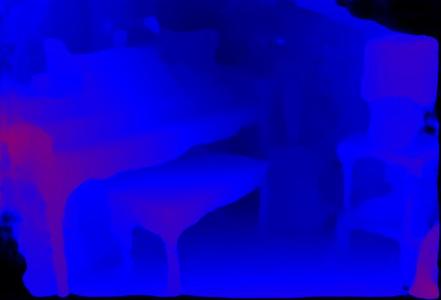}\\
        \vspace{-10pt}
        \includegraphics[width=\textwidth]{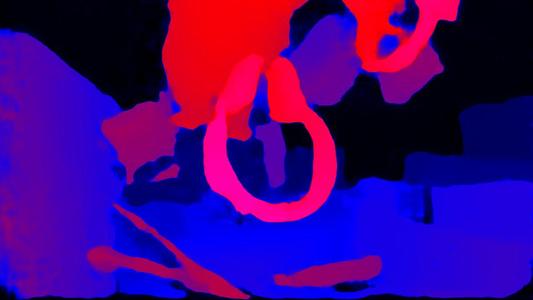}
        \vspace{-15pt}     
        \caption{\small{\textbf{Final labels $Y'$}}}
        \end{subfigure}
        \hspace{0.001cm}
        \begin{subfigure}[b]{0.22\textwidth}
        \includegraphics[width=\textwidth]{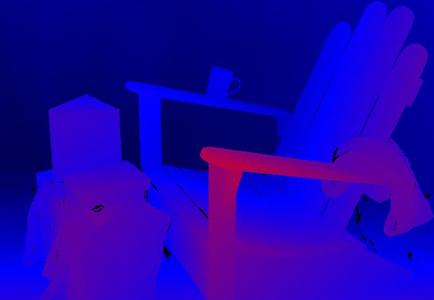}\\
        \vspace{-10pt}
        \includegraphics[width=\textwidth]{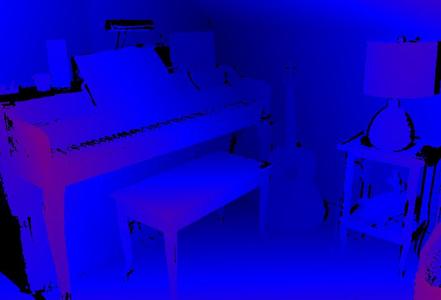}\\
        \vspace{-10pt}
        \includegraphics[width=\textwidth]{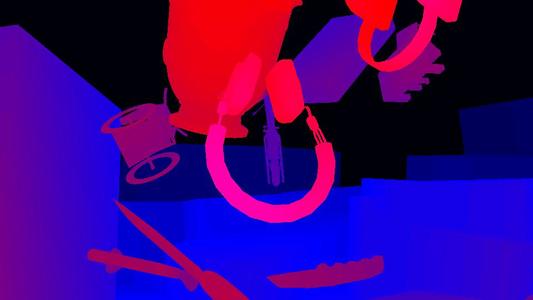}
        \vspace{-15pt}     
        \caption{\small{\textbf{Ground truth labels}}}
        \end{subfigure}
        \hspace{0.001cm}                                
        \end{center}
        \vspace{5pt}
\end{subfigure}
\vspace{-15pt}
\caption{Here we illustrate some examples of the disparity predictions that the "X-Blind" architecture performs. The illustrated examples are from the Synthetic and the Middlebury datasets.}
\label{fig:BlindAll}  
\vspace{-10pt}    
\end{figure*}
\begin{table*}
\centering
\renewcommand{\figurename}{Table}
\renewcommand{\captionlabelfont}{\bf}
\renewcommand{\captionfont}{\small} 
\resizebox{1.0\textwidth}{!}{
{\setlength{\extrarowheight}{2pt}\scriptsize
{\begin{tabular}{l <{\hspace{-0.3em}}||>{\hspace{-0.5em}} r | >{\hspace{-0.5em}} r | >{\hspace{-0.5em}} r || >{\hspace{-0.5em}} r | >{\hspace{-0.5em}} r | >{\hspace{-0.5em}} r || >{\hspace{-0.5em}} r | >{\hspace{-0.5em}} r | >{\hspace{-0.5em}} r || >{\hspace{-0.5em}} r | >{\hspace{-0.5em}} r | >{\hspace{-0.5em}} r || >{\hspace{-0.5em}} r | >{\hspace{-0.5em}} r | >{\hspace{-0.5em}} r }
\hline
\multicolumn{1}{c||}{} & \multicolumn{3}{c||}{$>$ 2 pixel} & \multicolumn{3}{c||}{$>$ 3 pixel} & \multicolumn{3}{c||}{$>$ 4 pixel} & \multicolumn{3}{c||}{$>$ 5 pixel} & \multicolumn{3}{c}{EPE}\\
\hline
Architectures & Non-Occ & All & Occ & Non-Occ & All & Occ & Non-Occ & All & Occ & Non-Occ & All & Occ & Non-Occ & All & Occ\\
\hline
\multicolumn{16}{c}{Synthetic dataset}\\
\hline
Initial labels $Y$ &  & 24.3175  &  &  & 22.9004 &  &  & 21.9140 &  &  & 21.1680 &  &  & 12.0218 & \\
\hline
\emph{Detect + Replace + Refine} &  & 9.5981 &  &  & 7.9764 &  &  & 6.7895 &  &  & 5.9074 &  &  & 1.8569 & \\
\hline
\emph{"X-Blind"} &  & 16.0014 &  &  & 14.0196 &  &  & 12.5170 &  &  & 11.3758 &  &  & 3.8810 & \\
\hline
\multicolumn{16}{c}{Middlebury dataset}\\
\hline
Initial labels $Y$ & 18.243 & 26.714 & 86.125 & 15.664 & 23.986 & 82.330 & 14.208 & 22.282 & 78.758 & 13.237 & 21.044 & 75.579 & 6.058 & 8.709 & 25.598\\
\hline
\emph{Detect + Replace + Refine} & 12.845 & 17.825 & 50.407 & 10.096 & 14.379 & 41.704 & 8.285 & 11.957 & 34.801 & 7.057 & 10.253 & 29.560 & 1.774 & 2.368 & 5.457\\
\hline
\emph{"X-Blind"} & 16.845 & 22.037 & 57.324 & 14.038 & 18.562 & 48.356 & 12.212 & 16.217 & 41.941 & 10.914 & 14.509 & 37.022 & 2.878 & 3.656 & 7.945\\
\hline
\multicolumn{16}{c}{KITTI 2015 dataset}\\
\hline
Initial labels $Y$ & 8.831 & 10.649 & 98.098 & 6.412 & 8.253 & 96.559 & 5.222 & 7.059 & 94.742 & 4.514 & 6.339 & 93.139 & 1.700 & 2.457 & 31.214\\
\hline
\emph{Detect + Replace + Refine} & 3.919 & 4.610 & 33.947 & 2.708 & 3.294 & 25.697 & 2.082 & 2.570 & 19.123 & 1.699 & 2.112 & 15.140 & 0.790 & 0.858 & 3.056\\
\hline
\emph{"X-Blind"} & 5.040 & 5.602 & 32.575 & 3.671 & 4.135 & 24.566 & 2.722 & 3.099 & 18.069 & 2.191 & 2.505 & 14.359 & 0.910 & 0.966 & 2.997\\
\hline
\end{tabular}}}}
\vspace{3pt}
\caption{\small{Stereo matching results for the "X-Blind" architecture. 
We also include the corresponding results of the proposed \emph{Detect + Replace + Refine} architecture to facilitate their comparison.}}
\label{tab:blind}
\vspace{-7pt}
\end{table*}

Here we evaluate the "X-Blind" architecture that, as already explained, it is exactly the same as the proposed \emph{Detect + Replace + Refine} architecture with the only difference being that as input gets only the initial labels $Y$ and not the image $X$.
The purpose of evaluating such an architecture is not to examine a competitive variant of the main \emph{Detect + Replace + Refine} architecture, but rather to explore the capabilities of the latter one in such a scenario.
In Table~\ref{tab:blind} we provide the stereo matching results of the "X-Blind" architecture.
We observe that it might not be able to compete the original \emph{Detect + Replace + Refine} architecture but it still can significantly improve the initial disparity label estimates.
In Figure~\ref{fig:BlindAll} we illustrate some disparity prediction examples generated by the "X-Blind" architecture. 
We observe that the "X-Blind" architecture manages to considerably improve the quality of the initial disparity label estimates, however,
since it does not have the image $X$ to guide it, it is not able to accurately reconstruct the disparity field on the borders of the objects.

\subsection{Qualitative results} \label{sec:qualresults}
This section includes qualitative examples that help illustrating the role of the various components of our proposed architecture.

\subsubsection{Error Detection step} \label{sec:ed_vis}
\begin{figure*}[t]
\centering
\renewcommand{\figurename}{Figure}
\renewcommand{\captionlabelfont}{\bf}
\renewcommand{\captionfont}{\small} 
\centering \textbf{Middlebury dataset}
\begin{subfigure}[b]{\textwidth}
\center
        \begin{center}
        \begin{subfigure}[b]{0.22\textwidth}
        \includegraphics[width=\textwidth]{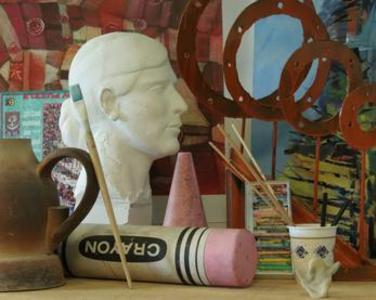}\\
        \vspace{-10pt}
        \includegraphics[width=\textwidth]{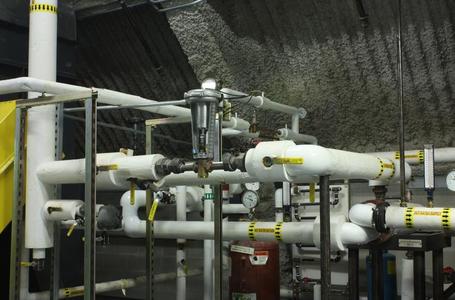}\\
        \vspace{-10pt}
        \includegraphics[width=\textwidth]{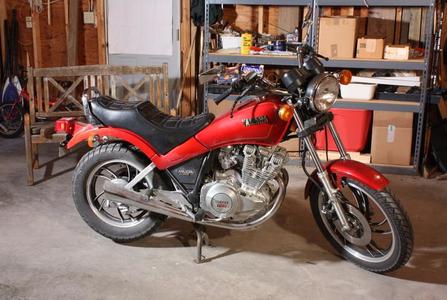}
        \vspace{-15pt}                
        \end{subfigure} 
        \hspace{0.001cm} 
        \begin{subfigure}[b]{0.22\textwidth}
        \includegraphics[width=\textwidth]{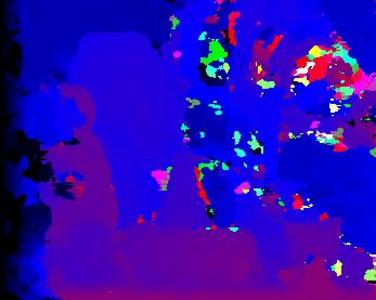}\\
        \vspace{-10pt}        
        \includegraphics[width=\textwidth]{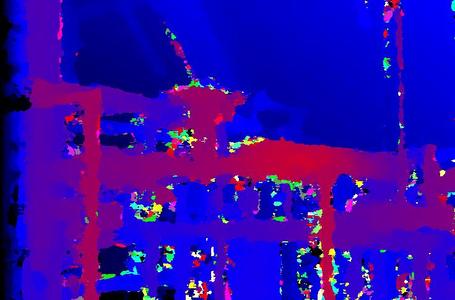}\\
        \vspace{-10pt}
        \includegraphics[width=\textwidth]{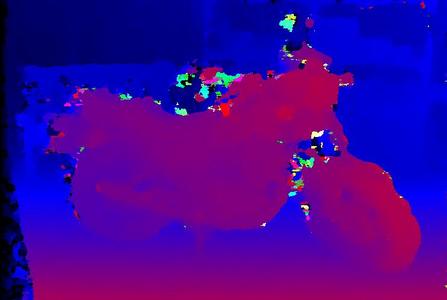}
        \vspace{-15pt}        
        \end{subfigure} 
        \hspace{0.001cm} 
        \begin{subfigure}[b]{0.22\textwidth}
        \includegraphics[width=\textwidth]{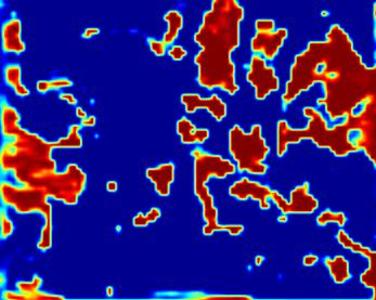}\\
        \vspace{-10pt}        
        \includegraphics[width=\textwidth]{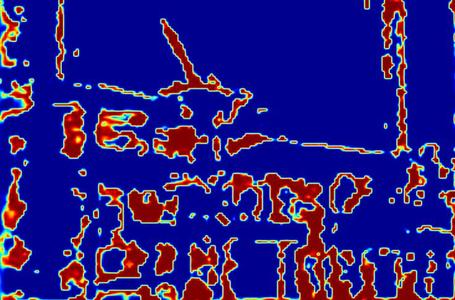}\\
        \vspace{-10pt}
        \includegraphics[width=\textwidth]{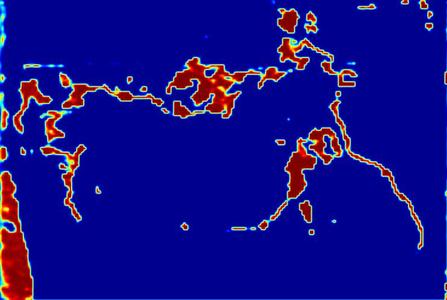}  
        \vspace{-15pt}                      
        \end{subfigure} 
        \hspace{0.001cm} 
        \begin{subfigure}[b]{0.22\textwidth}
        \includegraphics[width=\textwidth]{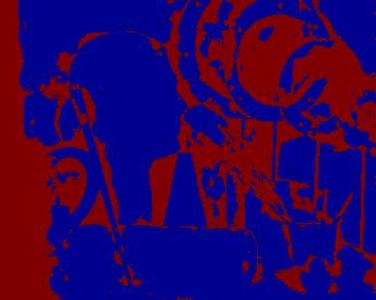}\\
        \vspace{-10pt}        
        \includegraphics[width=\textwidth]{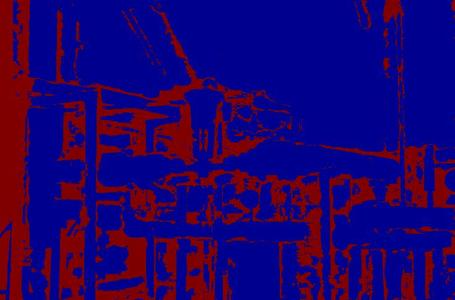} \\  
        \vspace{-10pt}        
        \includegraphics[width=\textwidth]{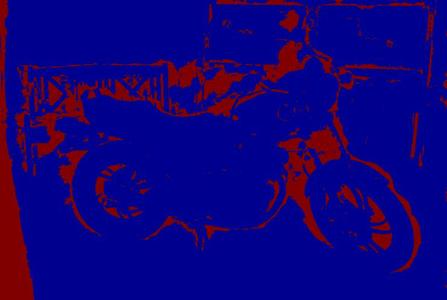}  
        \vspace{-15pt}        
        \end{subfigure}         
        \end{center}
        \vspace{5pt}
\end{subfigure}
\centering \textbf{Synthetic Dataset}
\begin{subfigure}[b]{\textwidth}
\center
        \begin{center}
        \begin{subfigure}[b]{0.22\textwidth}
        \includegraphics[width=\textwidth]{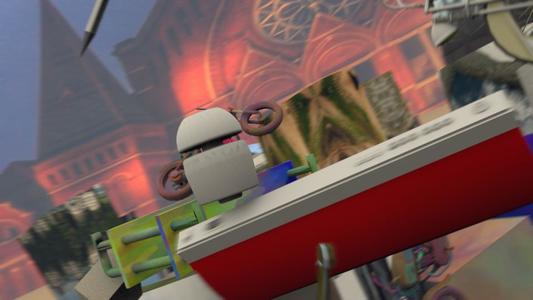}\\
        \vspace{-10pt}
        \includegraphics[width=\textwidth]{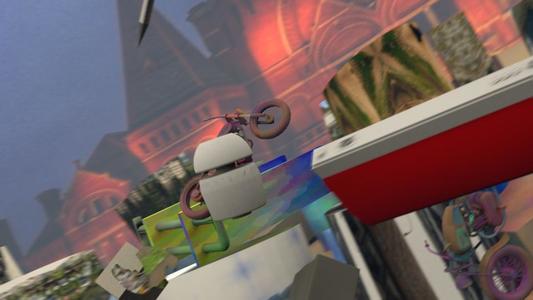}\\
        \vspace{-10pt}
        \includegraphics[width=\textwidth]{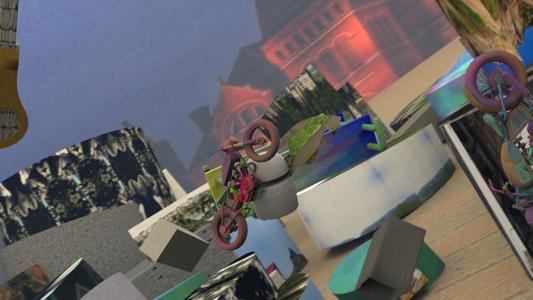}
        \vspace{-15pt}                
        \end{subfigure} 
        \hspace{0.001cm} 
        \begin{subfigure}[b]{0.22\textwidth}
        \includegraphics[width=\textwidth]{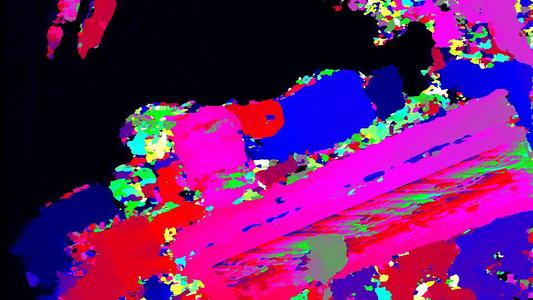}\\
        \vspace{-10pt}
        \includegraphics[width=\textwidth]{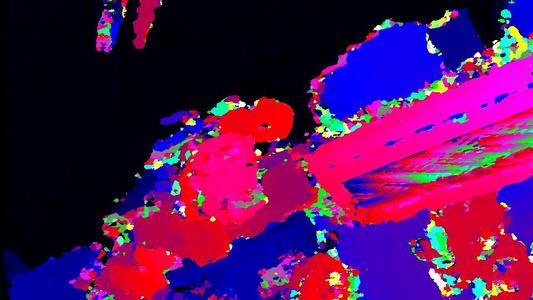}\\
        \vspace{-10pt}
        \includegraphics[width=\textwidth]{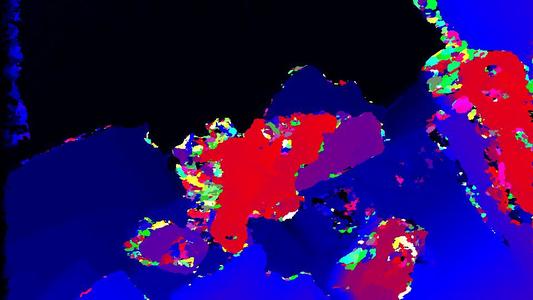}
        \vspace{-15pt}      
        \end{subfigure} 
        \hspace{0.001cm} 
        \begin{subfigure}[b]{0.22\textwidth}
        \includegraphics[width=\textwidth]{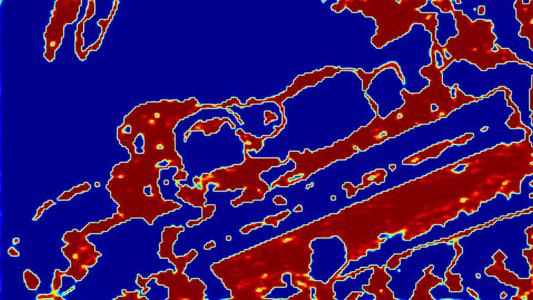}\\
        \vspace{-10pt}
        \includegraphics[width=\textwidth]{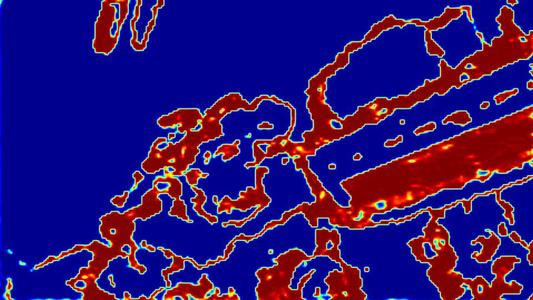}\\
        \vspace{-10pt}
        \includegraphics[width=\textwidth]{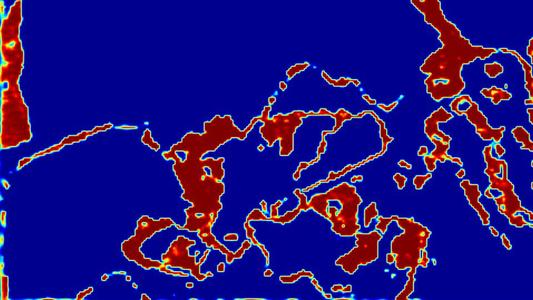}
        \vspace{-15pt}                      
        \end{subfigure} 
        \hspace{0.001cm} 
        \begin{subfigure}[b]{0.22\textwidth}
        \includegraphics[width=\textwidth]{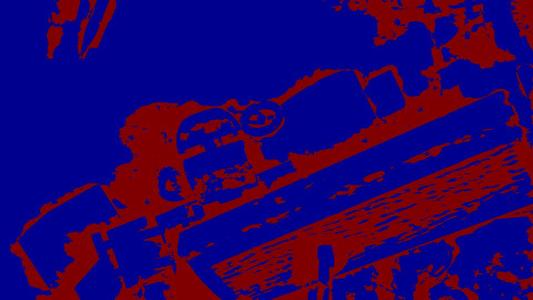}\\
        \vspace{-10pt}
        \includegraphics[width=\textwidth]{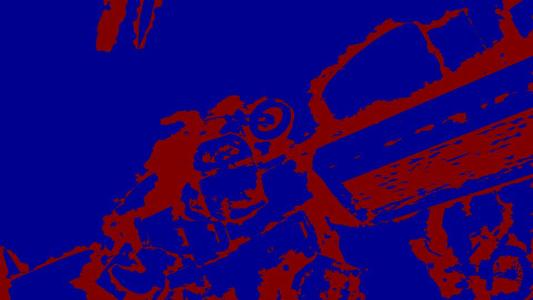}\\
        \vspace{-10pt}
        \includegraphics[width=\textwidth]{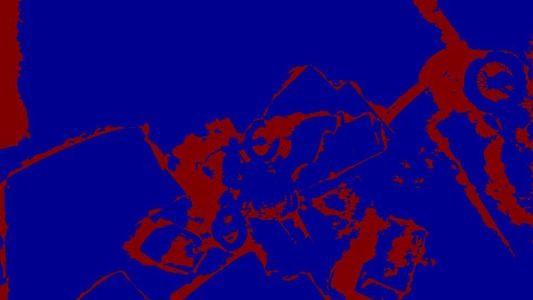}
        \vspace{-15pt}     
        \end{subfigure}         
        \end{center}
        \vspace{5pt}
\end{subfigure}
\centering \textbf{KITTI 2015 Dataset}
\begin{subfigure}[b]{\textwidth}
\center
        \begin{center}
        \begin{subfigure}[b]{0.22\textwidth}
        \includegraphics[width=\textwidth]{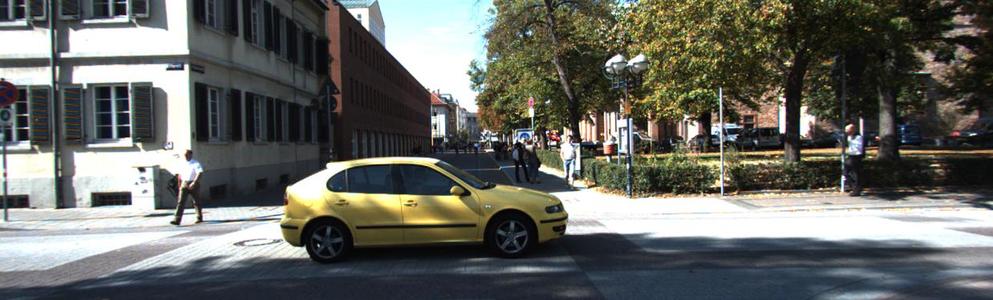}\\
        \vspace{-10pt}
        \includegraphics[width=\textwidth]{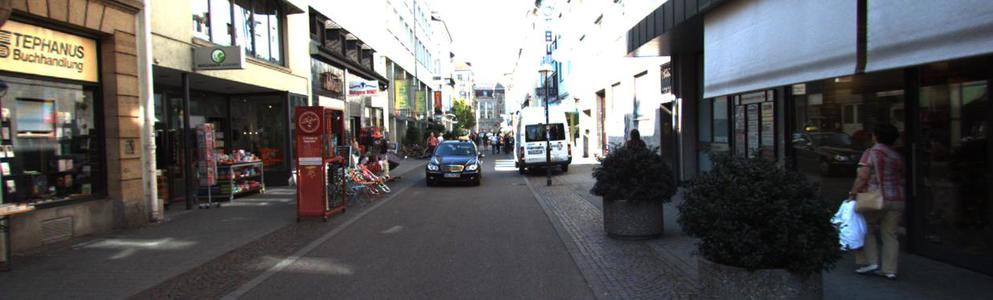}\\
        \vspace{-10pt}
        \includegraphics[width=\textwidth]{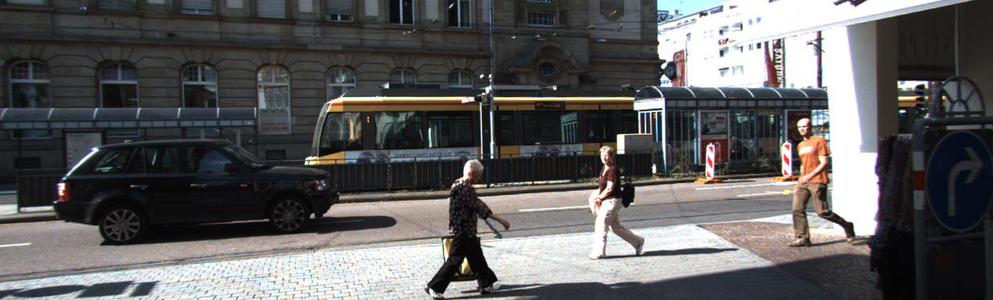}
        \vspace{-15pt}                
        \caption{\small{\textbf{Image $X$}}}
        \end{subfigure} 
        \hspace{0.001cm} 
        \begin{subfigure}[b]{0.22\textwidth}
        \includegraphics[width=\textwidth]{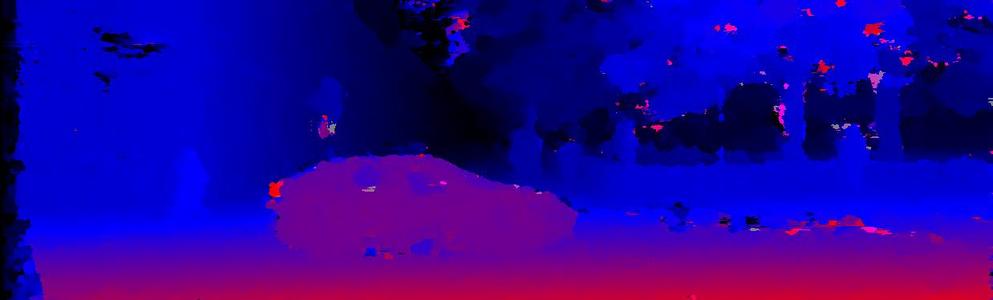}\\
        \vspace{-10pt}
        \includegraphics[width=\textwidth]{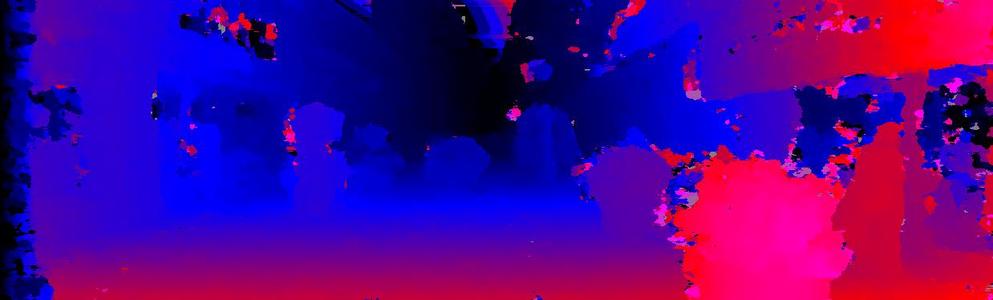}\\
        \vspace{-10pt}
        \includegraphics[width=\textwidth]{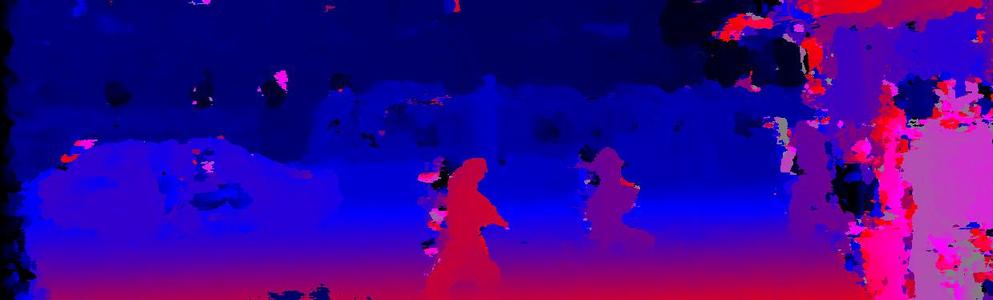}
        \vspace{-15pt}      
        \caption{\small{\textbf{Initial labels $Y$}}}
        \end{subfigure} 
        \hspace{0.001cm} 
        \begin{subfigure}[b]{0.22\textwidth}
        \includegraphics[width=\textwidth]{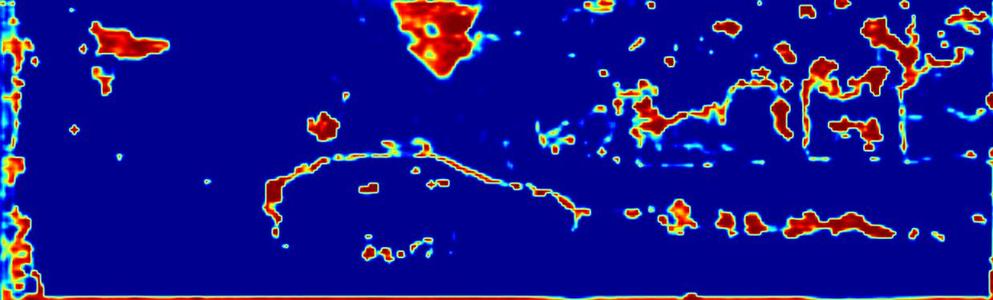}\\
        \vspace{-10pt}
        \includegraphics[width=\textwidth]{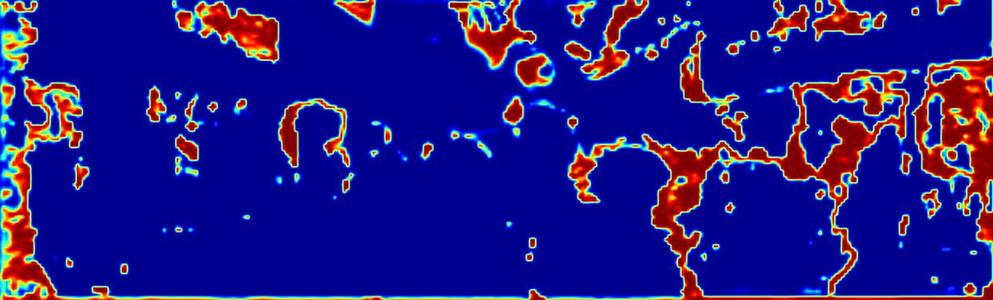}\\
        \vspace{-10pt}
        \includegraphics[width=\textwidth]{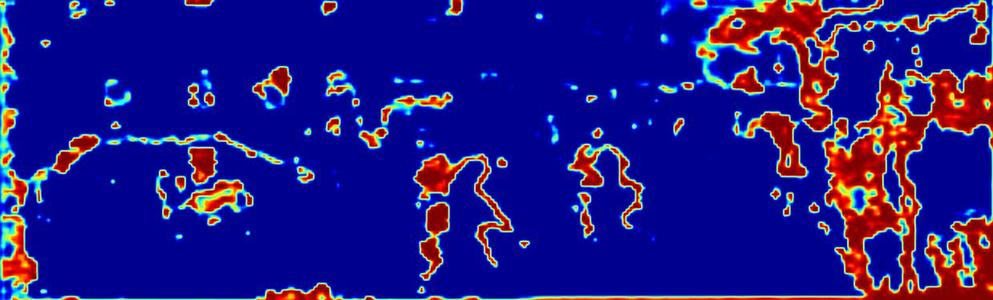}
        \vspace{-15pt}                      
        \caption{\small{\textbf{Predicted error map $E$}}}
        \end{subfigure} 
        \hspace{0.001cm} 
        \begin{subfigure}[b]{0.22\textwidth}
        \includegraphics[width=\textwidth]{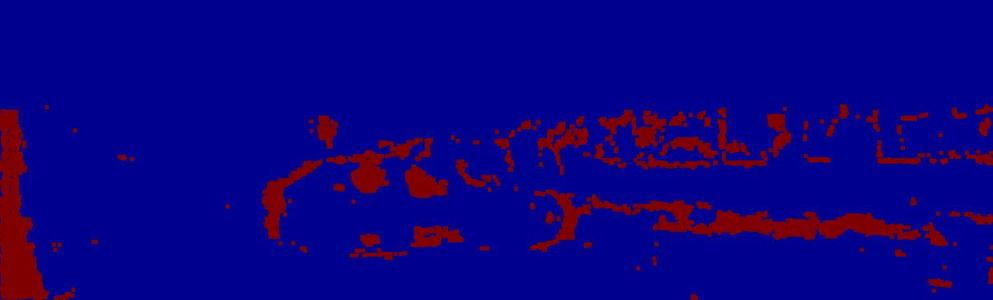}\\
        \vspace{-10pt}
        \includegraphics[width=\textwidth]{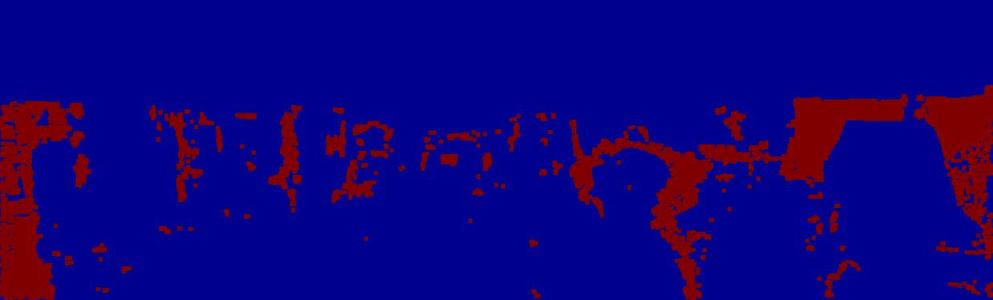}\\
        \vspace{-10pt}
        \includegraphics[width=\textwidth]{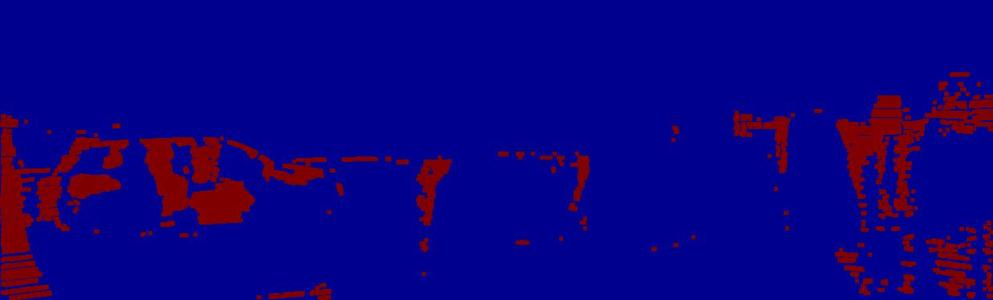}
        \vspace{-15pt}     
        \caption{\small{\textbf{Ground truth errors}}}
        \end{subfigure}         
        \end{center}
        \vspace{5pt}
\end{subfigure}
\vspace{-15pt}
\caption{
Illustration of the error probability maps $E$ that the error detection component $F_e(X,Y)$ yields.
The ground truth error maps are computed by thresholding the absolute difference of the initial labels $Y$ from the ground truth labels with a threshold of $3$ pixels (red are the erroneous pixel labels). 
Note that in the case of the KITTI 2015 dataset, the available ground truth labels are sparse and do not cover the entire image (\eg usually there is no annotation for the sky),
which is why some obviously erroneous initial label estimates are not coloured as incorrect (with red color) in the ground truth error maps.
}
\label{fig:VisEDC}        
\end{figure*}

In Figure~\ref{fig:VisEDC} we provide additional  examples of  error probability maps $E$ (that the error detection component $F_e(X,Y)$ generated w.r.t. the initial labels $Y$) and compare them with the ground truth error maps of the initial labels.
The ground truth error maps are computed by thresholding the absolute difference of the initial labels $Y$ from the ground truth labels with a threshold of $3$ pixels (red are the erroneous pixel labels in the figure). 
Note that this is the logic that is usually followed in the disparity task for considering a pixel label erroneous. 
We observe that, despite the fact the error detection component $F_e(.)$ is not explicitly trained to produce such ground truth error maps,
its predictions still highly correlate with them. 
This implies that   the error detection component $F_e(.)$ seems to have learnt to recognize  the areas that look  abnormal/atypical with respect to the joint input-output space $\{X,Y\}$  (\ie, it has learnt the ``structure'' of that space). 

\subsubsection{Replace step} \label{sec:rep_vis}

\begin{figure*}[t]
\centering
\renewcommand{\figurename}{Figure}
\renewcommand{\captionlabelfont}{\bf}
\renewcommand{\captionfont}{\small} 
\centering \textbf{Middlebury dataset}
\begin{subfigure}[b]{\textwidth}
\center
        \begin{center}
        \begin{subfigure}[b]{0.19\textwidth}
        \includegraphics[width=\textwidth]{figures/Middlebury/ArtL/X.jpg}\\
        \vspace{-10pt}
        \includegraphics[width=\textwidth]{figures/Middlebury/Pipes/X.jpg}\\
        \vspace{-10pt}
        \includegraphics[width=\textwidth]{figures/Middlebury/Motorcycle/X.jpg}
        \vspace{-15pt}                
        \end{subfigure} 
        \hspace{0.001cm} 
        \begin{subfigure}[b]{0.19\textwidth}
        \includegraphics[width=\textwidth]{figures/Middlebury/ArtL/Y.jpg}\\
        \vspace{-10pt}        
        \includegraphics[width=\textwidth]{figures/Middlebury/Pipes/Y.jpg}\\
        \vspace{-10pt}
        \includegraphics[width=\textwidth]{figures/Middlebury/Motorcycle/Y.jpg}
        \vspace{-15pt}        
        \end{subfigure} 
        \hspace{0.001cm} 
        \begin{subfigure}[b]{0.19\textwidth}
        \includegraphics[width=\textwidth]{figures/Middlebury/ArtL/E.jpg}\\
        \vspace{-10pt}        
        \includegraphics[width=\textwidth]{figures/Middlebury/Pipes/E.jpg}\\
        \vspace{-10pt}
        \includegraphics[width=\textwidth]{figures/Middlebury/Motorcycle/E.jpg}  
        \vspace{-15pt}                      
        \end{subfigure} 
        \hspace{0.001cm} 
        \begin{subfigure}[b]{0.19\textwidth}
        \includegraphics[width=\textwidth]{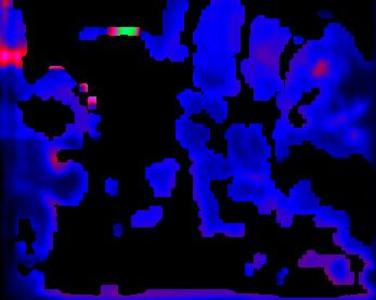}\\
        \vspace{-10pt}        
        \includegraphics[width=\textwidth]{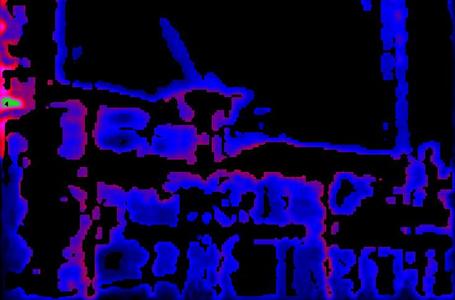} \\  
        \vspace{-10pt}        
        \includegraphics[width=\textwidth]{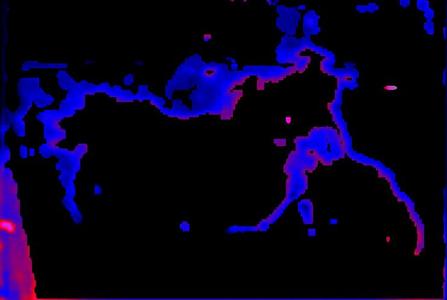}  
        \vspace{-15pt}        
        \end{subfigure} 
        \hspace{0.001cm} 
        \begin{subfigure}[b]{0.19\textwidth}
        \includegraphics[width=\textwidth]{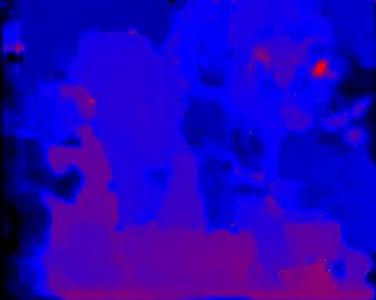}\\
        \vspace{-10pt}        
        \includegraphics[width=\textwidth]{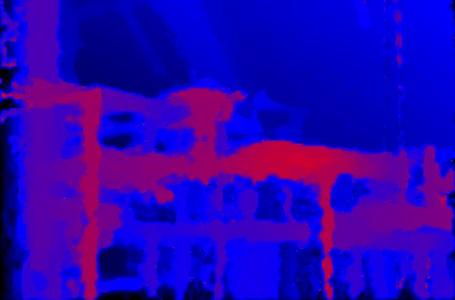} \\  
        \vspace{-10pt}        
        \includegraphics[width=\textwidth]{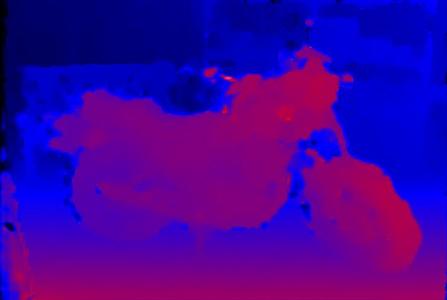}  
        \vspace{-15pt}        
        \end{subfigure}                
        \end{center}
        \vspace{5pt}
\end{subfigure}
\centering \textbf{Synthetic Dataset}
\begin{subfigure}[b]{\textwidth}
\center
        \begin{center}
        \begin{subfigure}[b]{0.19\textwidth}
        \includegraphics[width=\textwidth]{figures/SuntheticExamples/exmp000001/X.jpg}\\
        \vspace{-10pt}
        \includegraphics[width=\textwidth]{figures/SuntheticExamples/exmp000006/X.jpg}\\
        \vspace{-10pt}
        \includegraphics[width=\textwidth]{figures/SuntheticExamples/exmp000007/X.jpg}
        \vspace{-15pt}                
        \end{subfigure} 
        \hspace{0.001cm} 
        \begin{subfigure}[b]{0.19\textwidth}
        \includegraphics[width=\textwidth]{figures/SuntheticExamples/exmp000001/Y.jpg}\\
        \vspace{-10pt}
        \includegraphics[width=\textwidth]{figures/SuntheticExamples/exmp000006/Y.jpg}\\
        \vspace{-10pt}
        \includegraphics[width=\textwidth]{figures/SuntheticExamples/exmp000007/Y.jpg}
        \vspace{-15pt}      
        \end{subfigure} 
        \hspace{0.001cm} 
        \begin{subfigure}[b]{0.19\textwidth}
        \includegraphics[width=\textwidth]{figures/SuntheticExamples/exmp000001/E.jpg}\\
        \vspace{-10pt}
        \includegraphics[width=\textwidth]{figures/SuntheticExamples/exmp000006/E.jpg}\\
        \vspace{-10pt}
        \includegraphics[width=\textwidth]{figures/SuntheticExamples/exmp000007/E.jpg}
        \vspace{-15pt}                      
        \end{subfigure} 
        \hspace{0.001cm} 
        \begin{subfigure}[b]{0.19\textwidth}
        \includegraphics[width=\textwidth]{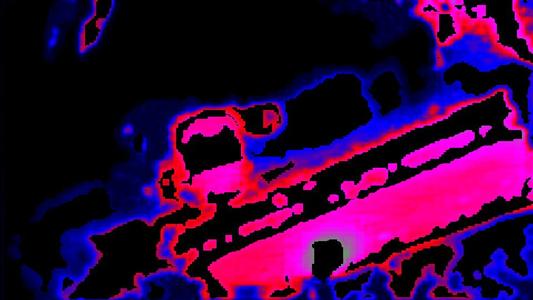}\\
        \vspace{-10pt}
        \includegraphics[width=\textwidth]{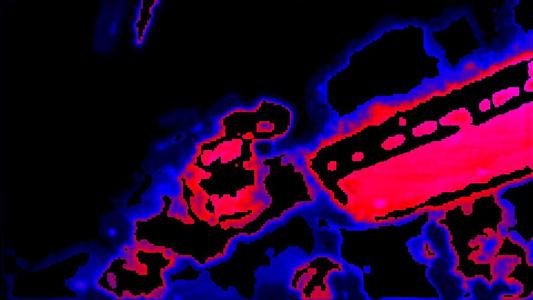}\\
        \vspace{-10pt}
        \includegraphics[width=\textwidth]{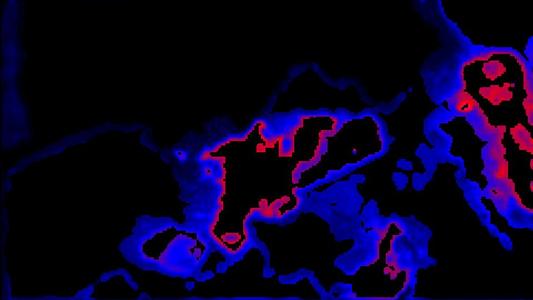}
        \vspace{-15pt}     
        \end{subfigure} 
        \hspace{0.001cm} 
        \begin{subfigure}[b]{0.19\textwidth}
        \includegraphics[width=\textwidth]{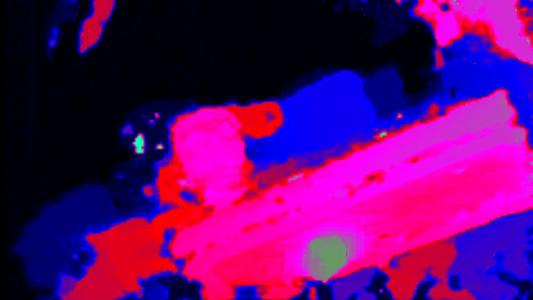}\\
        \vspace{-10pt}
        \includegraphics[width=\textwidth]{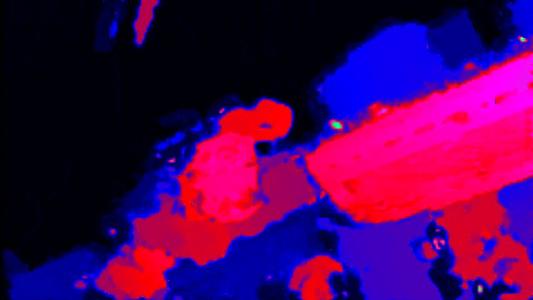}\\
        \vspace{-10pt}
        \includegraphics[width=\textwidth]{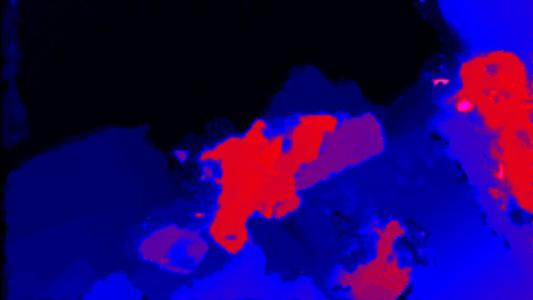}
        \vspace{-15pt}     
        \end{subfigure}                
        \end{center}
        \vspace{5pt}
\end{subfigure}
\centering \textbf{KITTI 2015 Dataset}
\begin{subfigure}[b]{\textwidth}
\center
        \begin{center}
        \begin{subfigure}[b]{0.19\textwidth}
        \includegraphics[width=\textwidth]{figures/KittiExamples/exmp000167_10/X.jpg}\\
        \vspace{-10pt}
        \includegraphics[width=\textwidth]{figures/KittiExamples/exmp000168_10/X.jpg}\\
        \vspace{-10pt}
        \includegraphics[width=\textwidth]{figures/KittiExamples/exmp000169_10/X.jpg}
        \vspace{-15pt}                
        \caption{\small{\textbf{Image $X$}}}
        \end{subfigure} 
        \hspace{0.001cm} 
        \begin{subfigure}[b]{0.19\textwidth}
        \includegraphics[width=\textwidth]{figures/KittiExamples/exmp000167_10/Y.jpg}\\
        \vspace{-10pt}
        \includegraphics[width=\textwidth]{figures/KittiExamples/exmp000168_10/Y.jpg}\\
        \vspace{-10pt}
        \includegraphics[width=\textwidth]{figures/KittiExamples/exmp000169_10/Y.jpg}
        \vspace{-15pt}      
        \caption{\small{\textbf{Initial labels $Y$}}}
        \end{subfigure} 
        \hspace{0.001cm} 
        \begin{subfigure}[b]{0.19\textwidth}
        \includegraphics[width=\textwidth]{figures/KittiExamples/exmp000167_10/E.jpg}\\
        \vspace{-10pt}
        \includegraphics[width=\textwidth]{figures/KittiExamples/exmp000168_10/E.jpg}\\
        \vspace{-10pt}
        \includegraphics[width=\textwidth]{figures/KittiExamples/exmp000169_10/E.jpg}
        \vspace{-15pt}                      
        \caption{\small{\textbf{Error map $E$}}}
        \end{subfigure} 
        \hspace{0.001cm} 
        \begin{subfigure}[b]{0.19\textwidth}
        \includegraphics[width=\textwidth]{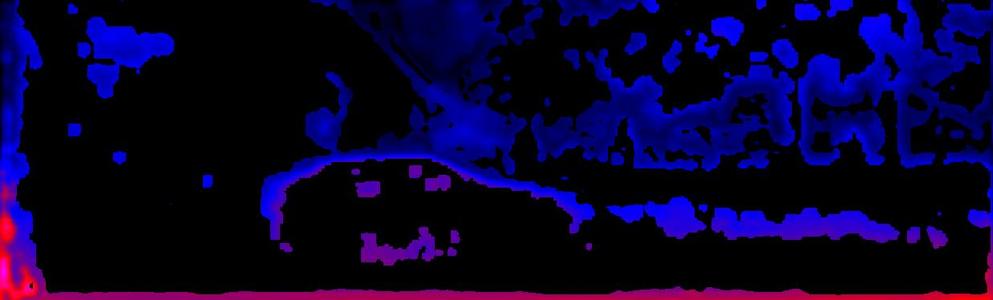}\\
        \vspace{-10pt}
        \includegraphics[width=\textwidth]{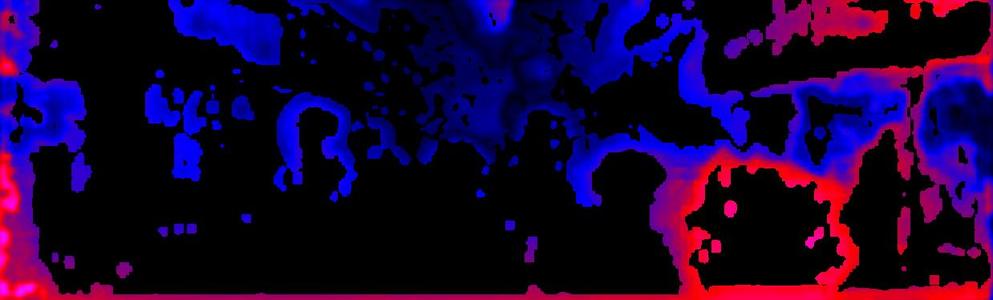}\\
        \vspace{-10pt}
        \includegraphics[width=\textwidth]{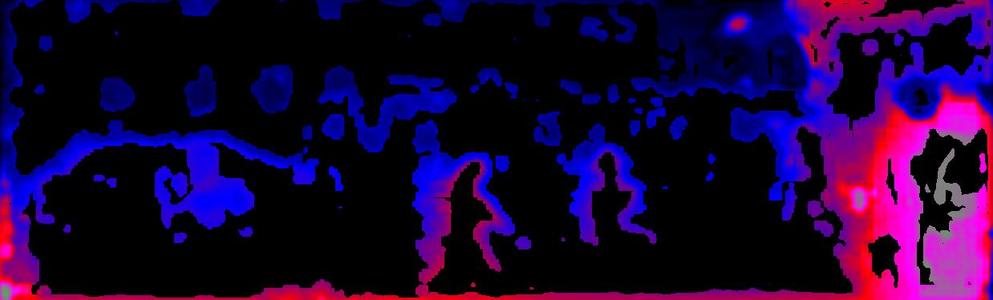}
        \vspace{-15pt}     
        \caption{\small{\textbf{$F_u(.)$ predictions}}}
        \end{subfigure}
        \hspace{0.001cm} 
        \begin{subfigure}[b]{0.19\textwidth}
        \includegraphics[width=\textwidth]{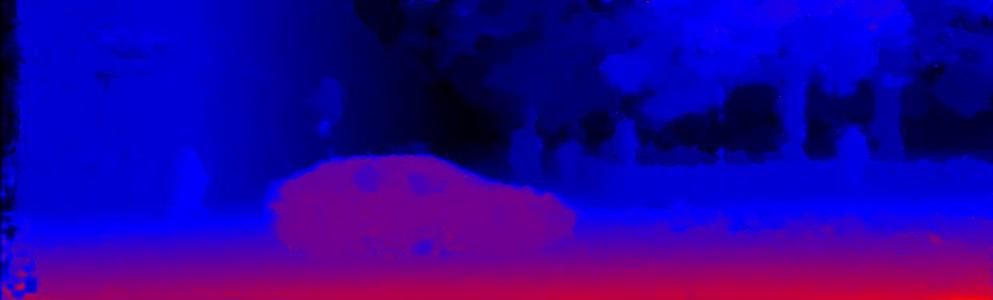}\\
        \vspace{-10pt}
        \includegraphics[width=\textwidth]{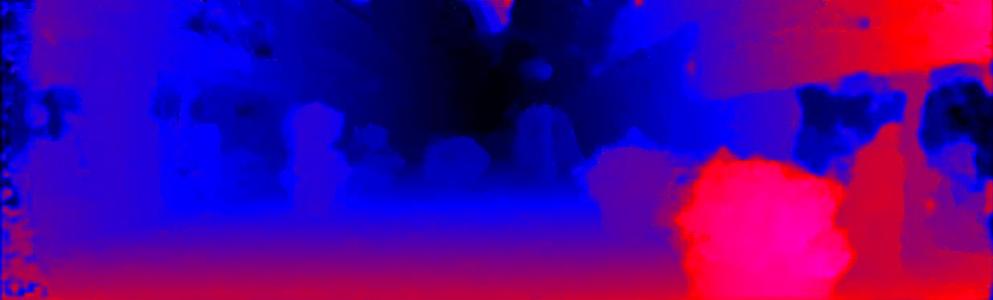}\\
        \vspace{-10pt}
        \includegraphics[width=\textwidth]{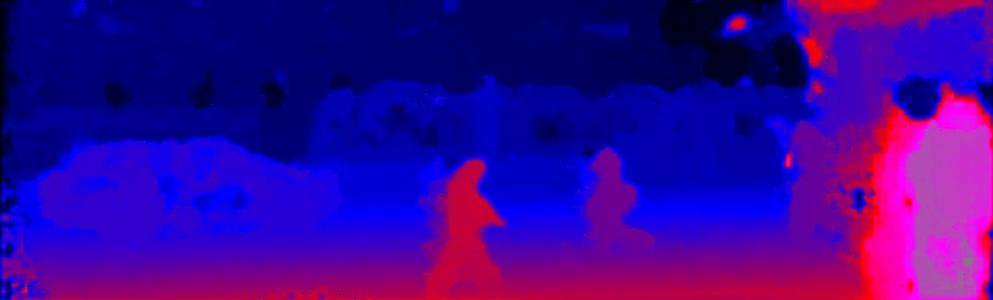}
        \vspace{-15pt}     
        \caption{\small{\textbf{Renewed labels $U$}}}
        \end{subfigure}                 
        \end{center}
        \vspace{5pt}
\end{subfigure}
\vspace{-10pt}
\caption{Here we provide more examples that illustrate the function performed by the Replace step in our proposed  architecture. Specifically, sub-figures \textbf{(a)}, \textbf{(b)}, and \textbf{(c)} depict the input image $X$,
the initial disparity label estimates $Y$, and the error probability map $E$ that the detection component $F_e(.)$ yields for the initial labels $Y$.
In sub-figure \textbf{(d)} we depict the label predictions of the replace component $F_u(.)$.
For visualization purposes we only depict the $F_u(.)$ pixel predictions that will replace the initial labels that are incorrect (according to the detection component) by drawing the remaining ones (\ie those that their error probability is less than $0.5$) with black color.
Finally, in the last sub-figure \textbf{(e)} we depict the renewed labels $U = E \odot F_u(X, Y, E) + (1-E) \odot Y$. We can readily observe that most of the ``hard" mistakes of the initial labels $Y$ have now been crudely ``fixed" by the Replace component. }
\label{fig:VisRepC}        
\end{figure*}

In Figure~\ref{fig:VisRepC} we provide several examples that  more clearly  illustrate the function performed by    the Replace step in our proposed  architecture.
Specifically, in sub-figures \ref{fig:VisRepC}a, \ref{fig:VisRepC}b, and \ref{fig:VisRepC}c we depict the input image $X$,
the initial disparity label estimates $Y$, and the error probability map $E$ that the detection component $F_e(.)$ yields for the initial labels $Y$.
In sub-figure \ref{fig:VisRepC}d we depict the label predictions of the replace component $F_u(.)$.
For visualization purposes we only depict the $F_u(.)$ pixel predictions that will replace the initial labels that are incorrect (according to the detection component) by drawing the remaining ones (\ie those that their error probability is less than $0.5$) with black color.
Finally, in the last sub-figure \ref{fig:VisRepC}e we depict the renewed labels $U = E \odot F_u(X, Y, E) + (1-E) \odot Y$. We can readily observe that most of the ``hard" mistakes of the initial labels $Y$ have now been crudely ``fixed" by the Replace component. 

\subsubsection{Refine step} \label{sec:rep_vis}

\begin{figure*}[t]
\centering
\renewcommand{\figurename}{Figure}
\renewcommand{\captionlabelfont}{\bf}
\renewcommand{\captionfont}{\small} 
\centering \textbf{Middlebury dataset}
\begin{subfigure}[b]{\textwidth}
\center
        \begin{center}
        \begin{subfigure}[b]{0.19\textwidth}
        \includegraphics[width=\textwidth]{figures/Middlebury/ArtL/X.jpg}\\
        \vspace{-10pt}
        \includegraphics[width=\textwidth]{figures/Middlebury/Pipes/X.jpg}\\
        \vspace{-10pt}
        \includegraphics[width=\textwidth]{figures/Middlebury/Motorcycle/X.jpg}
        \vspace{-15pt}                
        \end{subfigure} 
        \hspace{0.001cm} 
        \begin{subfigure}[b]{0.19\textwidth}
        \includegraphics[width=\textwidth]{figures/Middlebury/ArtL/Y.jpg}\\
        \vspace{-10pt}        
        \includegraphics[width=\textwidth]{figures/Middlebury/Pipes/Y.jpg}\\
        \vspace{-10pt}
        \includegraphics[width=\textwidth]{figures/Middlebury/Motorcycle/Y.jpg}
        \vspace{-15pt}        
        \end{subfigure} 
        \hspace{0.001cm} 
        \begin{subfigure}[b]{0.19\textwidth}
        \includegraphics[width=\textwidth]{figures/Middlebury/ArtL/U.jpg}\\
        \vspace{-10pt}        
        \includegraphics[width=\textwidth]{figures/Middlebury/Pipes/U.jpg}\\
        \vspace{-10pt}
        \includegraphics[width=\textwidth]{figures/Middlebury/Motorcycle/U.jpg}  
        \vspace{-15pt}                      
        \end{subfigure} 
        \hspace{0.001cm} 
        \begin{subfigure}[b]{0.19\textwidth}
        \includegraphics[width=\textwidth]{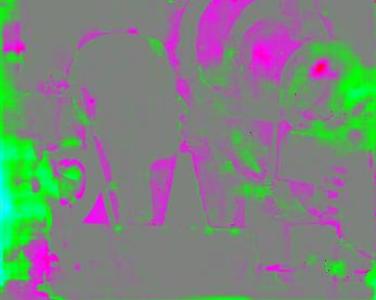}\\
        \vspace{-10pt}        
        \includegraphics[width=\textwidth]{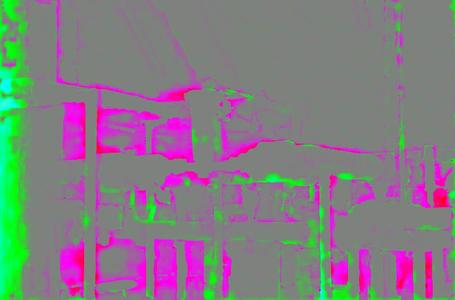} \\  
        \vspace{-10pt}        
        \includegraphics[width=\textwidth]{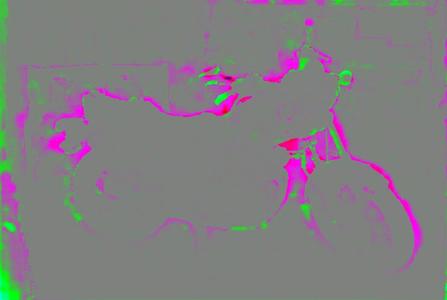}  
        \vspace{-15pt}        
        \end{subfigure} 
        \hspace{0.001cm} 
        \begin{subfigure}[b]{0.19\textwidth}
        \includegraphics[width=\textwidth]{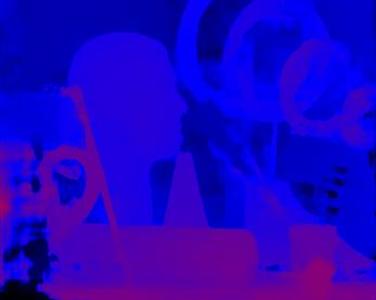}\\
        \vspace{-10pt}        
        \includegraphics[width=\textwidth]{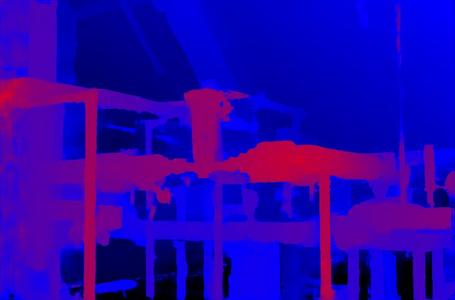} \\  
        \vspace{-10pt}        
        \includegraphics[width=\textwidth]{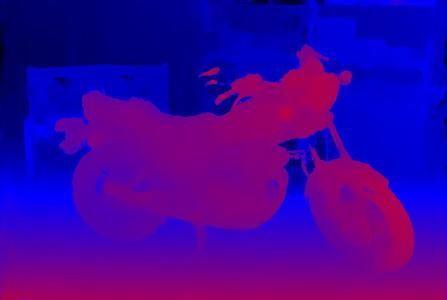}  
        \vspace{-15pt}        
        \end{subfigure}                
        \end{center}
        \vspace{5pt}
\end{subfigure}
\centering \textbf{Synthetic Dataset}
\begin{subfigure}[b]{\textwidth}
\center
        \begin{center}
        \begin{subfigure}[b]{0.19\textwidth}
        \includegraphics[width=\textwidth]{figures/SuntheticExamples/exmp000001/X.jpg}\\
        \vspace{-10pt}
        \includegraphics[width=\textwidth]{figures/SuntheticExamples/exmp000006/X.jpg}\\
        \vspace{-10pt}
        \includegraphics[width=\textwidth]{figures/SuntheticExamples/exmp000007/X.jpg}
        \vspace{-15pt}                
        \end{subfigure} 
        \hspace{0.001cm} 
        \begin{subfigure}[b]{0.19\textwidth}
        \includegraphics[width=\textwidth]{figures/SuntheticExamples/exmp000001/Y.jpg}\\
        \vspace{-10pt}
        \includegraphics[width=\textwidth]{figures/SuntheticExamples/exmp000006/Y.jpg}\\
        \vspace{-10pt}
        \includegraphics[width=\textwidth]{figures/SuntheticExamples/exmp000007/Y.jpg}
        \vspace{-15pt}      
        \end{subfigure} 
        \hspace{0.001cm} 
        \begin{subfigure}[b]{0.19\textwidth}
        \includegraphics[width=\textwidth]{figures/SuntheticExamples/exmp000001/U.jpg}\\
        \vspace{-10pt}
        \includegraphics[width=\textwidth]{figures/SuntheticExamples/exmp000006/U.jpg}\\
        \vspace{-10pt}
        \includegraphics[width=\textwidth]{figures/SuntheticExamples/exmp000007/U.jpg}
        \vspace{-15pt}                      
        \end{subfigure} 
        \hspace{0.001cm} 
        \begin{subfigure}[b]{0.19\textwidth}
        \includegraphics[width=\textwidth]{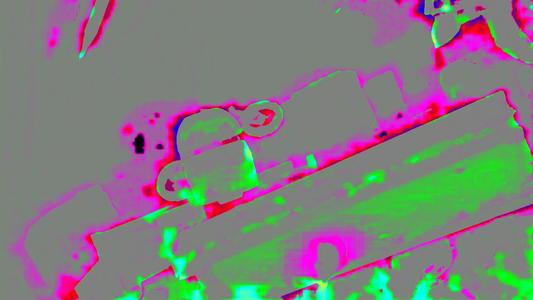}\\
        \vspace{-10pt}
        \includegraphics[width=\textwidth]{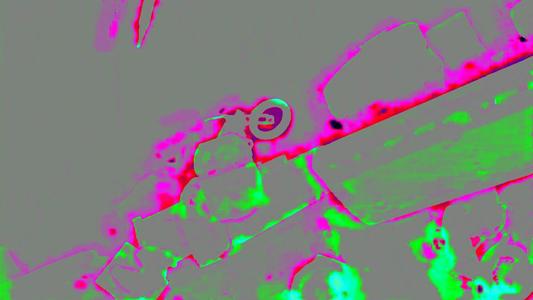}\\
        \vspace{-10pt}
        \includegraphics[width=\textwidth]{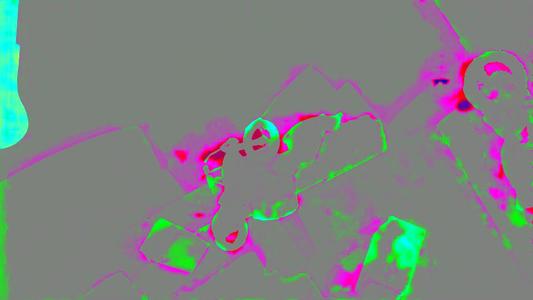}
        \vspace{-15pt}     
        \end{subfigure} 
        \hspace{0.001cm} 
        \begin{subfigure}[b]{0.19\textwidth}
        \includegraphics[width=\textwidth]{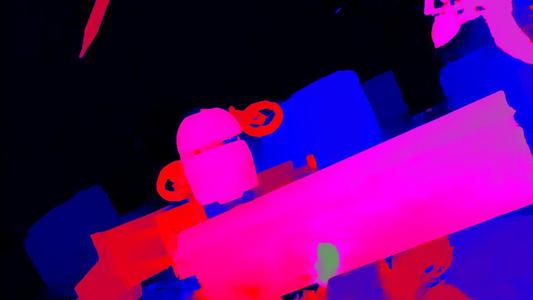}\\
        \vspace{-10pt}
        \includegraphics[width=\textwidth]{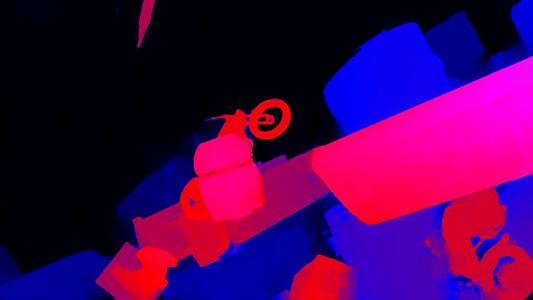}\\
        \vspace{-10pt}
        \includegraphics[width=\textwidth]{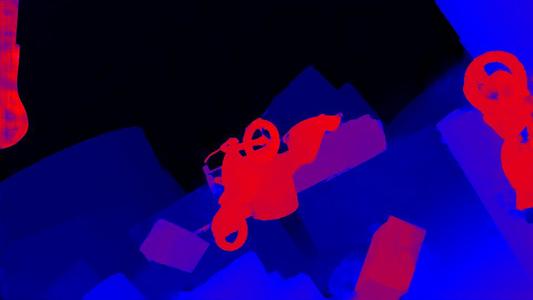}
        \vspace{-15pt}     
        \end{subfigure}                
        \end{center}
        \vspace{5pt}
\end{subfigure}
\centering \textbf{KITTI 2015 Dataset}
\begin{subfigure}[b]{\textwidth}
\center
        \begin{center}
        \begin{subfigure}[b]{0.19\textwidth}
        \includegraphics[width=\textwidth]{figures/KittiExamples/exmp000167_10/X.jpg}\\
        \vspace{-10pt}
        \includegraphics[width=\textwidth]{figures/KittiExamples/exmp000168_10/X.jpg}\\
        \vspace{-10pt}
        \includegraphics[width=\textwidth]{figures/KittiExamples/exmp000169_10/X.jpg}
        \vspace{-15pt}                
        \caption{\small{\textbf{Image $X$}}}
        \end{subfigure} 
        \hspace{0.001cm} 
        \begin{subfigure}[b]{0.19\textwidth}
        \includegraphics[width=\textwidth]{figures/KittiExamples/exmp000167_10/Y.jpg}\\
        \vspace{-10pt}
        \includegraphics[width=\textwidth]{figures/KittiExamples/exmp000168_10/Y.jpg}\\
        \vspace{-10pt}
        \includegraphics[width=\textwidth]{figures/KittiExamples/exmp000169_10/Y.jpg}
        \vspace{-15pt}      
        \caption{\small{\textbf{Initial labels $Y$}}}
        \end{subfigure} 
        \hspace{0.001cm} 
        \begin{subfigure}[b]{0.19\textwidth}
        \includegraphics[width=\textwidth]{figures/KittiExamples/exmp000167_10/U.jpg}\\
        \vspace{-10pt}
        \includegraphics[width=\textwidth]{figures/KittiExamples/exmp000168_10/U.jpg}\\
        \vspace{-10pt}
        \includegraphics[width=\textwidth]{figures/KittiExamples/exmp000169_10/U.jpg}
        \vspace{-15pt}                      
        \caption{\small{\textbf{Renewed labels $U$}}}
        \end{subfigure} 
        \hspace{0.001cm} 
        \begin{subfigure}[b]{0.19\textwidth}
        \includegraphics[width=\textwidth]{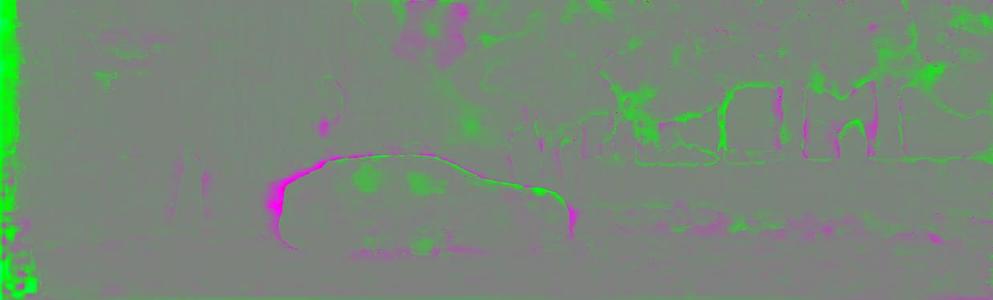}\\
        \vspace{-10pt}
        \includegraphics[width=\textwidth]{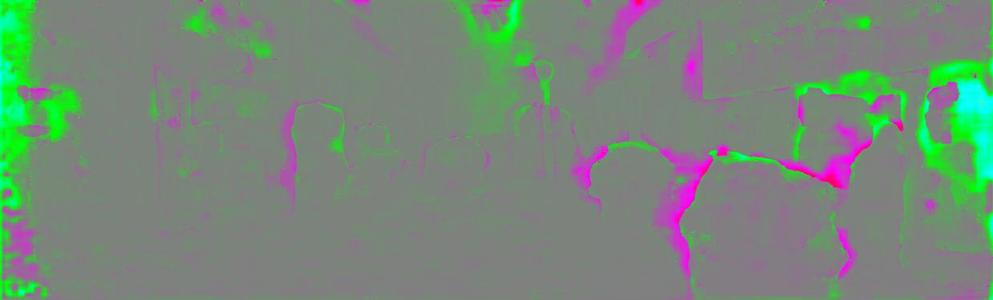}\\
        \vspace{-10pt}
        \includegraphics[width=\textwidth]{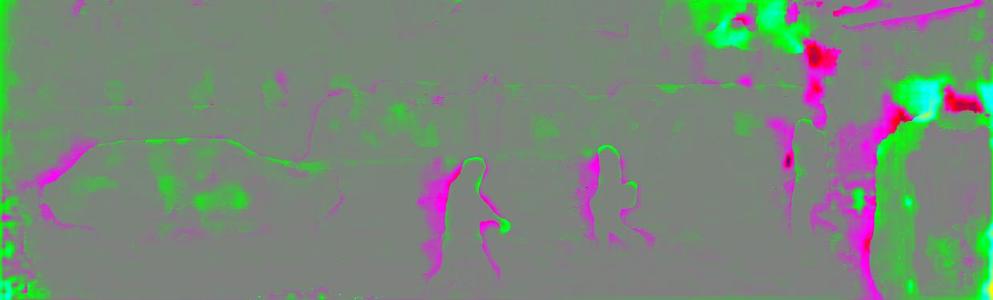}
        \vspace{-15pt}     
        \caption{\small{\textbf{$F_r(.)$ residuals}}}
        \end{subfigure}
        \hspace{0.001cm} 
        \begin{subfigure}[b]{0.19\textwidth}
        \includegraphics[width=\textwidth]{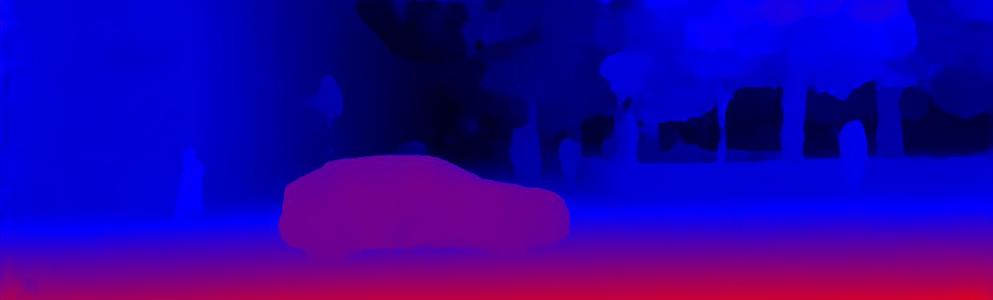}\\
        \vspace{-10pt}
        \includegraphics[width=\textwidth]{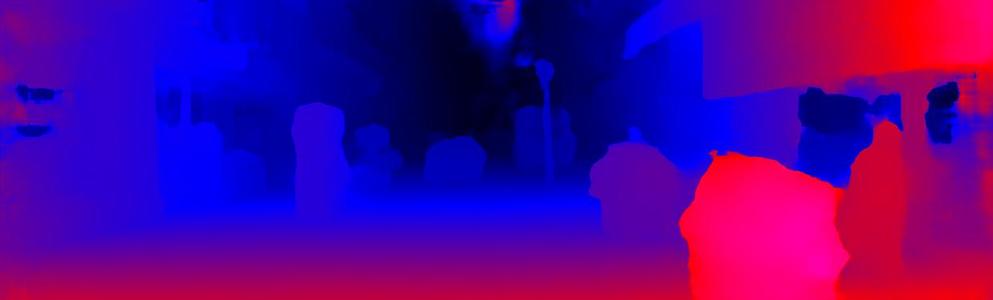}\\
        \vspace{-10pt}
        \includegraphics[width=\textwidth]{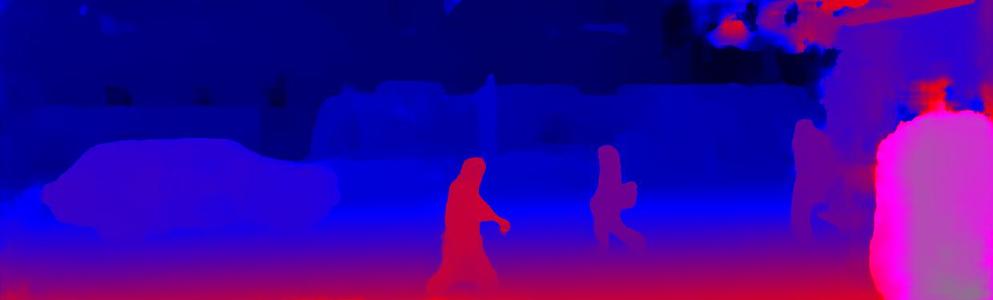}
        \vspace{-15pt}     
        \caption{\small{\textbf{Final labels $Y'$}}}
        \end{subfigure}                 
        \end{center}
        \vspace{5pt}
\end{subfigure}
\vspace{-10pt}
\caption{
Here we provide more examples that illustrate the function performed by the Refine step in our proposed  architecture.
Specifically, in sub-figures \textbf{(a)}, \textbf{(b)}, and \textbf{(c)} we depict the input image $X$,
the initial disparity label estimates $Y$, and the renewed labels $U$ that the Replace step yields.
In sub-figure \textbf{(d)} we depict the residual corrections that the Refine component $F_r(.)$ yields for the renewed labels $U$. 
Finally, in the last sub-figure \textbf{(e)} we depict the final label estimates $Y' = U + F_r(X,Y,E,U)$ that the Refine step yields.}
\label{fig:VisRefC}        
\end{figure*}

In Figure~\ref{fig:VisRefC} we provide several examples that  more clearly  illustrate the function performed by the Refine step in our proposed  architecture.
Specifically, in sub-figures \ref{fig:VisRefC}a, \ref{fig:VisRefC}b, and \ref{fig:VisRefC}c we depict the input image $X$,
the initial disparity label estimates $Y$, and the renewed labels $U$ that the Replace step yields.
In sub-figure \ref{fig:VisRefC}d we depict the residual corrections that the Refine component $F_r(.)$ yields for the renewed labels $U$. 
Finally, in last sub-figure \ref{fig:VisRefC}e we depict the final label estimates $Y' = U + F_r(X,Y,E,U)$ that the Refine step yields.
We observe that most of residual corrections that the Refine component $F_r(.)$ yields are concentrated on the borders of the objects.
Furthermore, by adding those residuals on the renewed labels $U$, the Refine step manages to refine the renewed labels $U$ and align the estimated labels $Y'$ with the fine image structures in $X$.

\subsubsection{Detect, Replace, Refine pipeline} \label{sec:drr_vis}

\begin{figure*}[t]
\centering
\renewcommand{\figurename}{Figure}
\renewcommand{\captionlabelfont}{\bf}
\renewcommand{\captionfont}{\small} 
\centering \textbf{Middlebury dataset}
\begin{subfigure}[b]{\textwidth}
\center
        \begin{center}
        \begin{subfigure}[b]{0.15\textwidth}
        \includegraphics[width=\textwidth]{figures/Middlebury/ArtL/X.jpg}\\
        \vspace{-10pt}
        \includegraphics[width=\textwidth]{figures/Middlebury/Pipes/X.jpg}\\
        \vspace{-10pt}
        \includegraphics[width=\textwidth]{figures/Middlebury/Motorcycle/X.jpg}
        \vspace{-15pt}                
        \end{subfigure} 
        \hspace{0.001cm} 
        \begin{subfigure}[b]{0.15\textwidth}
        \includegraphics[width=\textwidth]{figures/Middlebury/ArtL/Y.jpg}\\
        \vspace{-10pt}        
        \includegraphics[width=\textwidth]{figures/Middlebury/Pipes/Y.jpg}\\
        \vspace{-10pt}
        \includegraphics[width=\textwidth]{figures/Middlebury/Motorcycle/Y.jpg}
        \vspace{-15pt}        
        \end{subfigure} 
        \hspace{0.001cm} 
        \begin{subfigure}[b]{0.15\textwidth}
        \includegraphics[width=\textwidth]{figures/Middlebury/ArtL/E.jpg}\\
        \vspace{-10pt}        
        \includegraphics[width=\textwidth]{figures/Middlebury/Pipes/E.jpg}\\
        \vspace{-10pt}
        \includegraphics[width=\textwidth]{figures/Middlebury/Motorcycle/E.jpg}  
        \vspace{-15pt}                      
        \end{subfigure} 
        \hspace{0.001cm} 
        \begin{subfigure}[b]{0.15\textwidth}
        \includegraphics[width=\textwidth]{figures/Middlebury/ArtL/U.jpg}\\
        \vspace{-10pt}        
        \includegraphics[width=\textwidth]{figures/Middlebury/Pipes/U.jpg} \\  
        \vspace{-10pt}        
        \includegraphics[width=\textwidth]{figures/Middlebury/Motorcycle/U.jpg}  
        \vspace{-15pt}        
        \end{subfigure}
        \hspace{0.001cm} 
        \begin{subfigure}[b]{0.15\textwidth}
        \includegraphics[width=\textwidth]{figures/Middlebury/ArtL/YY.jpg}\\
        \vspace{-10pt}        
        \includegraphics[width=\textwidth]{figures/Middlebury/Pipes/YY.jpg} \\  
        \vspace{-10pt}        
        \includegraphics[width=\textwidth]{figures/Middlebury/Motorcycle/YY.jpg}  
        \vspace{-15pt}        
        \end{subfigure}
        \hspace{0.001cm} 
        \begin{subfigure}[b]{0.15\textwidth}
        \includegraphics[width=\textwidth]{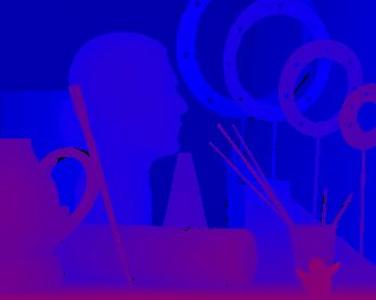}\\
        \vspace{-10pt}        
        \includegraphics[width=\textwidth]{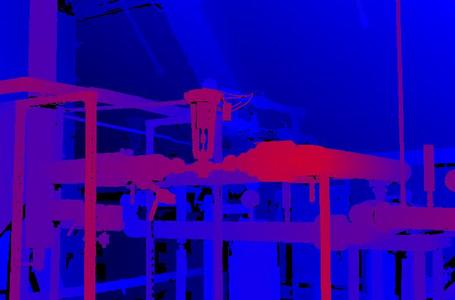} \\  
        \vspace{-10pt}        
        \includegraphics[width=\textwidth]{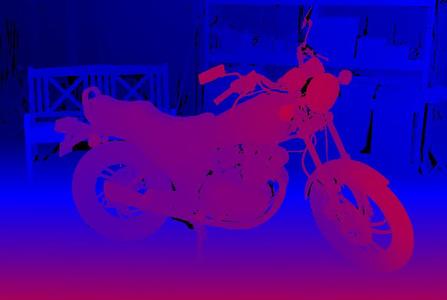}  
        \vspace{-15pt}        
        \end{subfigure}                               
        \end{center}
        \vspace{5pt}
\end{subfigure}
\centering \textbf{Synthetic Dataset}
\begin{subfigure}[b]{\textwidth}
\center
        \begin{center}
        \begin{subfigure}[b]{0.15\textwidth}
        \includegraphics[width=\textwidth]{figures/SuntheticExamples/exmp000001/X.jpg}\\
        \vspace{-10pt}
        \includegraphics[width=\textwidth]{figures/SuntheticExamples/exmp000006/X.jpg}\\
        \vspace{-10pt}
        \includegraphics[width=\textwidth]{figures/SuntheticExamples/exmp000007/X.jpg}
        \vspace{-15pt}                
        \end{subfigure} 
        \hspace{0.001cm} 
        \begin{subfigure}[b]{0.15\textwidth}
        \includegraphics[width=\textwidth]{figures/SuntheticExamples/exmp000001/Y.jpg}\\
        \vspace{-10pt}
        \includegraphics[width=\textwidth]{figures/SuntheticExamples/exmp000006/Y.jpg}\\
        \vspace{-10pt}
        \includegraphics[width=\textwidth]{figures/SuntheticExamples/exmp000007/Y.jpg}
        \vspace{-15pt}      
        \end{subfigure} 
        \hspace{0.001cm} 
        \begin{subfigure}[b]{0.15\textwidth}
        \includegraphics[width=\textwidth]{figures/SuntheticExamples/exmp000001/E.jpg}\\
        \vspace{-10pt}
        \includegraphics[width=\textwidth]{figures/SuntheticExamples/exmp000006/E.jpg}\\
        \vspace{-10pt}
        \includegraphics[width=\textwidth]{figures/SuntheticExamples/exmp000007/E.jpg}
        \vspace{-15pt}                      
        \end{subfigure} 
        \hspace{0.001cm} 
        \begin{subfigure}[b]{0.15\textwidth}
        \includegraphics[width=\textwidth]{figures/SuntheticExamples/exmp000001/U.jpg}\\
        \vspace{-10pt}
        \includegraphics[width=\textwidth]{figures/SuntheticExamples/exmp000006/U.jpg}\\
        \vspace{-10pt}
        \includegraphics[width=\textwidth]{figures/SuntheticExamples/exmp000007/U.jpg}
        \vspace{-15pt}     
        \end{subfigure} 
        \hspace{0.001cm} 
        \begin{subfigure}[b]{0.15\textwidth}
        \includegraphics[width=\textwidth]{figures/SuntheticExamples/exmp000001/YY.jpg}\\
        \vspace{-10pt}
        \includegraphics[width=\textwidth]{figures/SuntheticExamples/exmp000006/YY.jpg}\\
        \vspace{-10pt}
        \includegraphics[width=\textwidth]{figures/SuntheticExamples/exmp000007/YY.jpg}
        \vspace{-15pt}     
        \end{subfigure}
        \hspace{0.001cm} 
        \begin{subfigure}[b]{0.15\textwidth}
        \includegraphics[width=\textwidth]{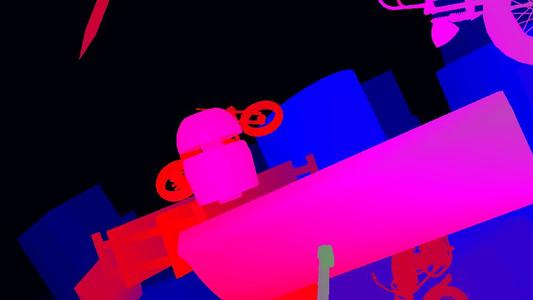}\\
        \vspace{-10pt}
        \includegraphics[width=\textwidth]{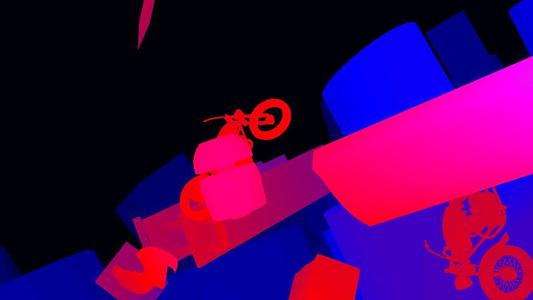}\\
        \vspace{-10pt}
        \includegraphics[width=\textwidth]{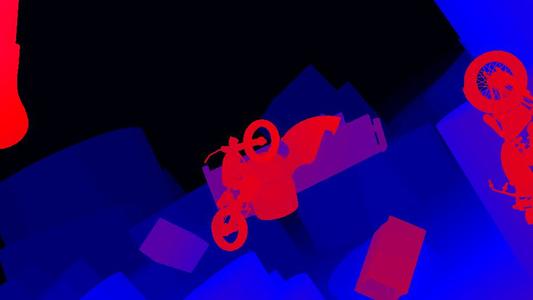}
        \vspace{-15pt}     
        \end{subfigure}                                  
        \end{center}
        \vspace{5pt}
\end{subfigure}
\centering \textbf{KITTI 2015 Dataset}
\begin{subfigure}[b]{\textwidth}
\center
        \begin{center}
        \begin{subfigure}[b]{0.15\textwidth}
        \includegraphics[width=\textwidth]{figures/KittiExamples/exmp000167_10/X.jpg}\\
        \vspace{-10pt}
        \includegraphics[width=\textwidth]{figures/KittiExamples/exmp000168_10/X.jpg}\\
        \vspace{-10pt}
        \includegraphics[width=\textwidth]{figures/KittiExamples/exmp000169_10/X.jpg}
        \vspace{-15pt}                
        \caption{\small{\textbf{Image $X$}}}
        \end{subfigure} 
        \hspace{0.001cm} 
        \begin{subfigure}[b]{0.15\textwidth}
        \includegraphics[width=\textwidth]{figures/KittiExamples/exmp000167_10/Y.jpg}\\
        \vspace{-10pt}
        \includegraphics[width=\textwidth]{figures/KittiExamples/exmp000168_10/Y.jpg}\\
        \vspace{-10pt}
        \includegraphics[width=\textwidth]{figures/KittiExamples/exmp000169_10/Y.jpg}
        \vspace{-15pt}      
        \caption{\small{\textbf{Initial labels $Y$}}}
        \end{subfigure} 
        \hspace{0.001cm} 
        \begin{subfigure}[b]{0.15\textwidth}
        \includegraphics[width=\textwidth]{figures/KittiExamples/exmp000167_10/E.jpg}\\
        \vspace{-10pt}
        \includegraphics[width=\textwidth]{figures/KittiExamples/exmp000168_10/E.jpg}\\
        \vspace{-10pt}
        \includegraphics[width=\textwidth]{figures/KittiExamples/exmp000169_10/E.jpg}
        \vspace{-15pt}                      
        \caption{\small{\textbf{Error map $E$}}}
        \end{subfigure} 
        \hspace{0.001cm} 
        \begin{subfigure}[b]{0.15\textwidth}
        \includegraphics[width=\textwidth]{figures/KittiExamples/exmp000167_10/U.jpg}\\
        \vspace{-10pt}
        \includegraphics[width=\textwidth]{figures/KittiExamples/exmp000168_10/U.jpg}\\
        \vspace{-10pt}
        \includegraphics[width=\textwidth]{figures/KittiExamples/exmp000169_10/U.jpg}
        \vspace{-15pt}     
        \caption{\small{\textbf{Labels $U$}}}
        \end{subfigure} 
        \begin{subfigure}[b]{0.15\textwidth}
        \includegraphics[width=\textwidth]{figures/KittiExamples/exmp000167_10/YY.jpg}\\
        \vspace{-10pt}
        \includegraphics[width=\textwidth]{figures/KittiExamples/exmp000168_10/YY.jpg}\\
        \vspace{-10pt}
        \includegraphics[width=\textwidth]{figures/KittiExamples/exmp000169_10/YY.jpg}
        \vspace{-15pt}     
        \caption{\small{\textbf{Final labels $Y'$}}}
        \end{subfigure}
        \hspace{0.001cm}
        \begin{subfigure}[b]{0.15\textwidth}
        \includegraphics[width=\textwidth]{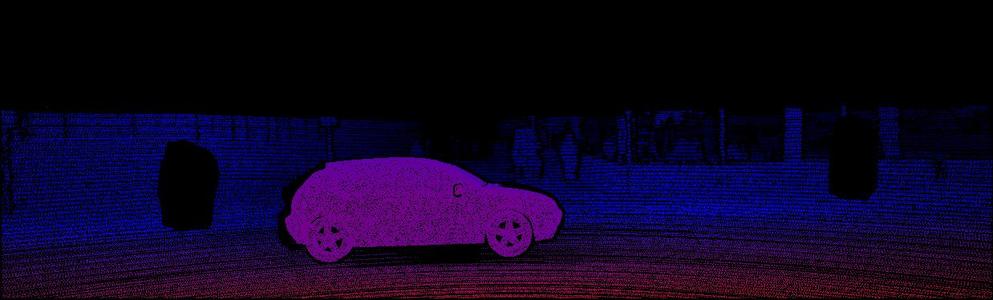}\\
        \vspace{-10pt}
        \includegraphics[width=\textwidth]{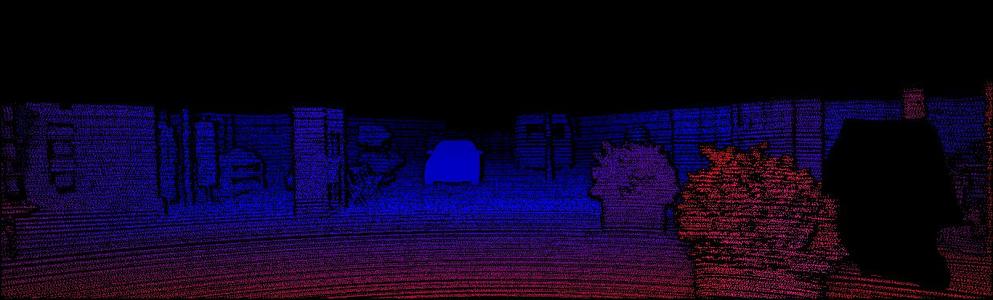}\\
        \vspace{-10pt}
        \includegraphics[width=\textwidth]{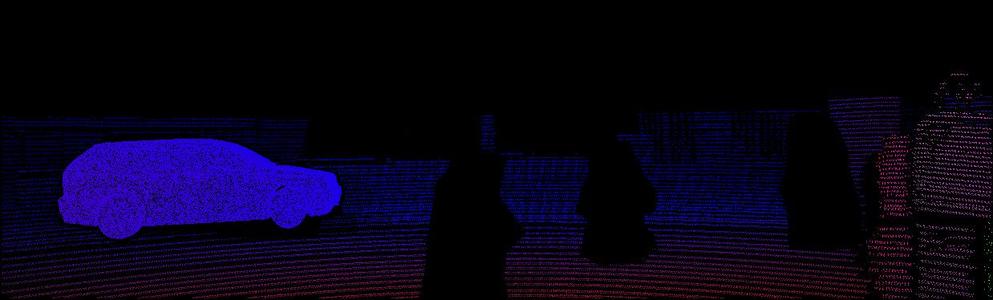}
        \vspace{-15pt}     
        \caption{\small{\textbf{Ground truth}}}
        \end{subfigure}
        \hspace{0.001cm}                                
        \end{center}
        \vspace{5pt}
\end{subfigure}
\vspace{-10pt}
\caption{Illustration of the intermediate steps of the \emph{Detect + Replace + Refine} work-flow.
We observe that the final Refine component $F_r(.)$, by predicting residual corrections, manages to refine the renewed labels $U$ and align the output labels $Y'$ with the fine image structures in image $X$. Note that in the case of the KITTI 2015 dataset, the available ground truth labels are sparse and do not cover the entire image.}
\label{fig:VisAll}        
\end{figure*}

In Figure~\ref{fig:VisAll} we illustrate the entire work-flow of the \emph{Detect + Replace + Refine} architecture that we propose and we compare its predictions $Y'$ with the ground truth disparity labels.

\subsubsection{Multi-iteration architecture} \label{sec:zoom}
\begin{figure*}[t]
\centering
\renewcommand{\figurename}{Figure}
\renewcommand{\captionlabelfont}{\bf}
\renewcommand{\captionfont}{\small} 
\centering \textbf{Middlebury Dataset}
\begin{subfigure}[b]{\textwidth}
\center
        \begin{center}
        \begin{subfigure}[b]{0.19\textwidth}
        \includegraphics[width=\textwidth]{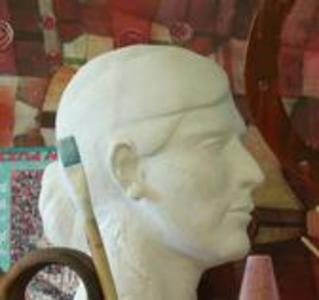}\\
        \vspace{-10pt}
        \includegraphics[width=\textwidth]{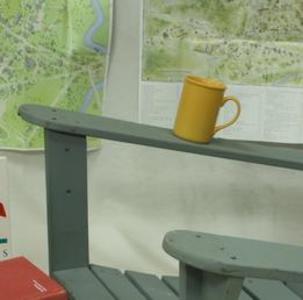}\\
        \vspace{-10pt}
        \includegraphics[width=\textwidth]{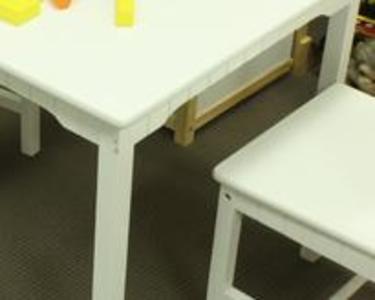}
        \vspace{-15pt}                
        \end{subfigure} 
        \hspace{0.001cm} 
        \begin{subfigure}[b]{0.19\textwidth}
        \includegraphics[width=\textwidth]{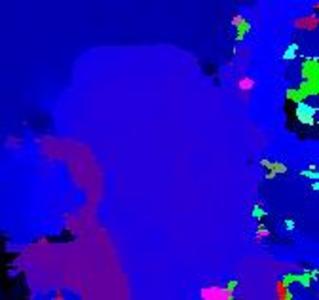}\\
        \vspace{-10pt}
        \includegraphics[width=\textwidth]{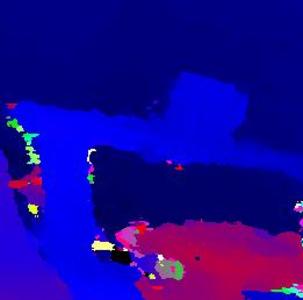}\\
        \vspace{-10pt}
        \includegraphics[width=\textwidth]{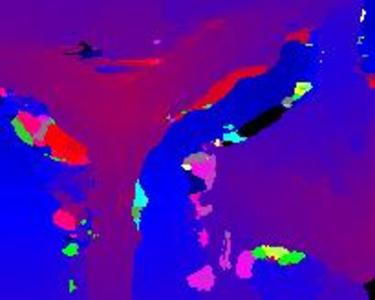}
        \vspace{-15pt}      
        \end{subfigure} 
        \hspace{0.001cm} 
        \begin{subfigure}[b]{0.19\textwidth}
        \includegraphics[width=\textwidth]{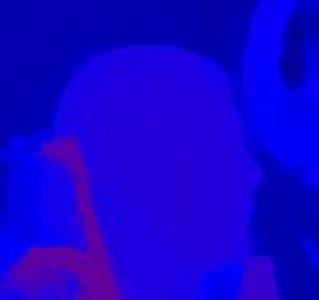}\\
        \vspace{-10pt}
        \includegraphics[width=\textwidth]{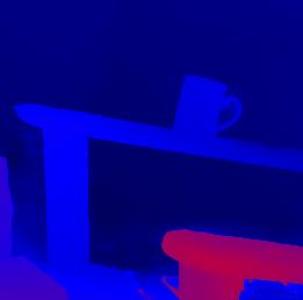}\\
        \vspace{-10pt}
        \includegraphics[width=\textwidth]{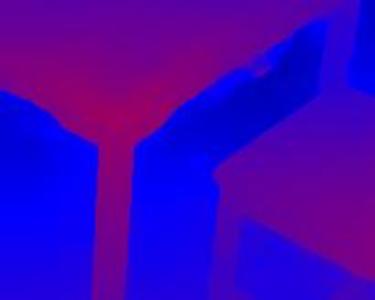}
        \vspace{-15pt}                      
        \end{subfigure} 
        \hspace{0.001cm} 
        \begin{subfigure}[b]{0.19\textwidth}
        \includegraphics[width=\textwidth]{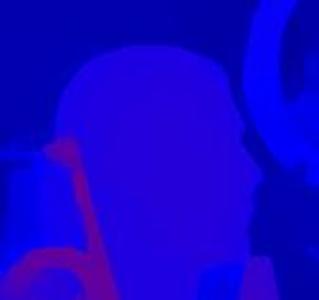}\\
        \vspace{-10pt}
        \includegraphics[width=\textwidth]{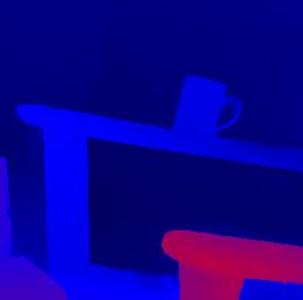}\\
        \vspace{-10pt}
        \includegraphics[width=\textwidth]{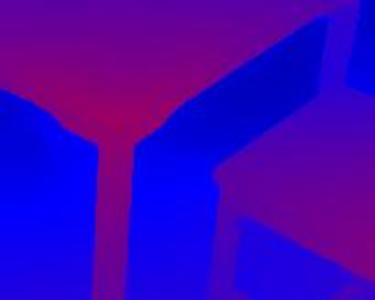}
        \vspace{-15pt}     
        \end{subfigure}
        \hspace{0.001cm} 
        \begin{subfigure}[b]{0.19\textwidth}
        \includegraphics[width=\textwidth]{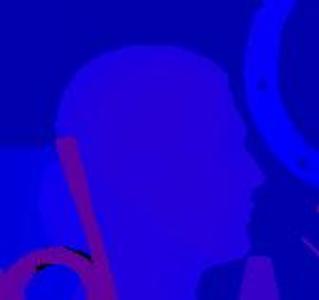}\\
        \vspace{-10pt}
        \includegraphics[width=\textwidth]{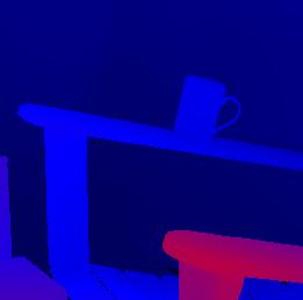}\\
        \vspace{-10pt}
        \includegraphics[width=\textwidth]{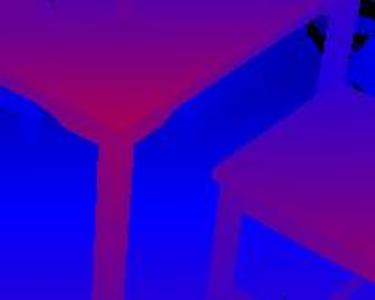}
        \vspace{-15pt}     
        \end{subfigure}                 
        \end{center}
        \vspace{5pt}
\end{subfigure}
\centering \textbf{Synthetic Dataset}
\begin{subfigure}[b]{\textwidth}
\center
        \begin{center}
        \begin{subfigure}[b]{0.19\textwidth}
        \includegraphics[width=\textwidth]{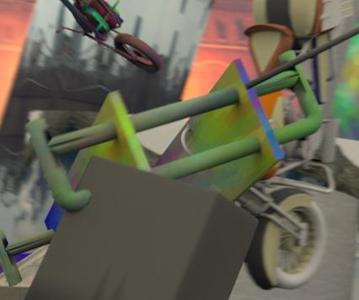}\\
        \vspace{-10pt}
        \includegraphics[width=\textwidth]{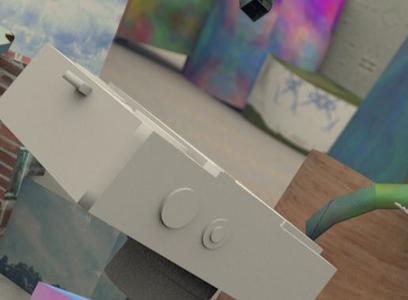}\\
        \vspace{-10pt}
        \includegraphics[width=\textwidth]{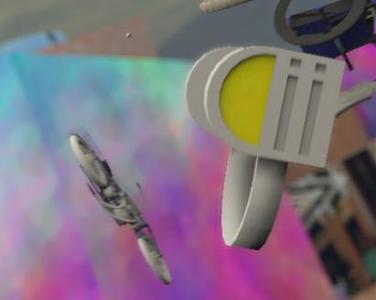}
        \vspace{-15pt}                
        \caption{\small{\textbf{Image $X$}}}
        \end{subfigure} 
        \hspace{0.001cm} 
        \begin{subfigure}[b]{0.19\textwidth}
        \includegraphics[width=\textwidth]{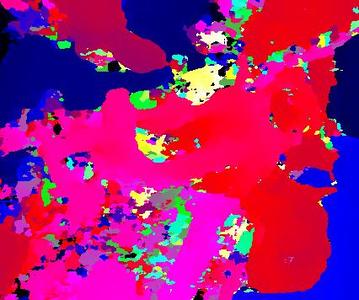}\\
        \vspace{-10pt}
        \includegraphics[width=\textwidth]{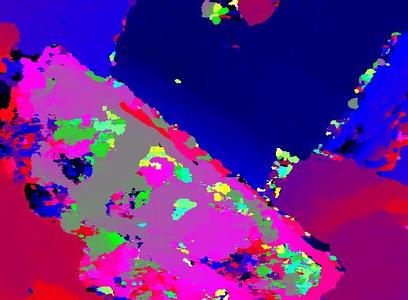}\\
        \vspace{-10pt}
        \includegraphics[width=\textwidth]{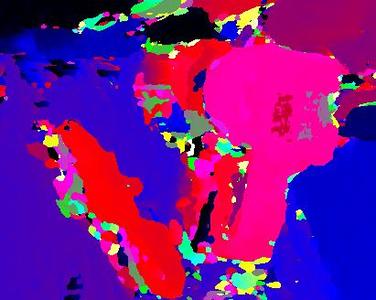}
        \vspace{-15pt}      
        \caption{\small{\textbf{Initial labels $Y$}}}
        \end{subfigure} 
        \hspace{0.001cm} 
        \begin{subfigure}[b]{0.19\textwidth}
        \includegraphics[width=\textwidth]{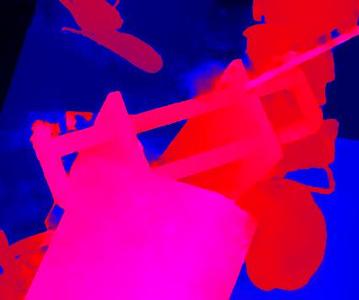}\\
        \vspace{-10pt}
        \includegraphics[width=\textwidth]{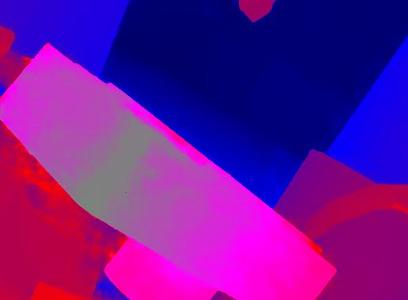}\\
        \vspace{-10pt}
        \includegraphics[width=\textwidth]{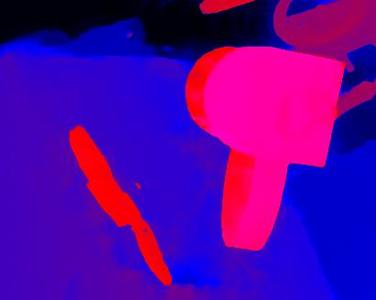}
        \vspace{-15pt}                      
        \caption{\small{\textbf{1st iteration labels}}}
        \end{subfigure} 
        \hspace{0.001cm} 
        \begin{subfigure}[b]{0.19\textwidth}
        \includegraphics[width=\textwidth]{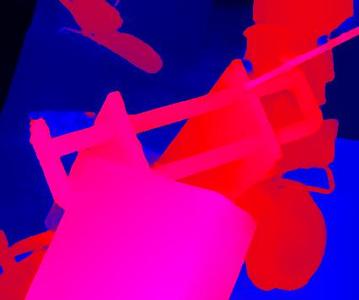}\\
        \vspace{-10pt}
        \includegraphics[width=\textwidth]{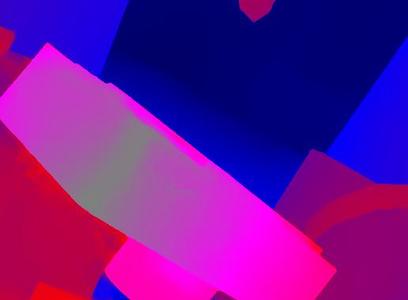}\\
        \vspace{-10pt}
        \includegraphics[width=\textwidth]{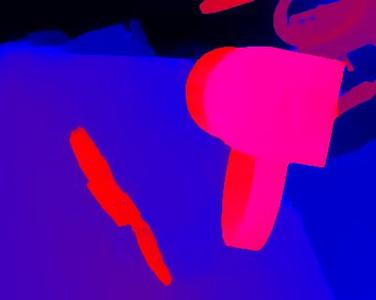}
        \vspace{-15pt}     
        \caption{\small{\textbf{2nd iteration labels}}}
        \end{subfigure}
        \hspace{0.001cm} 
        \begin{subfigure}[b]{0.19\textwidth}
        \includegraphics[width=\textwidth]{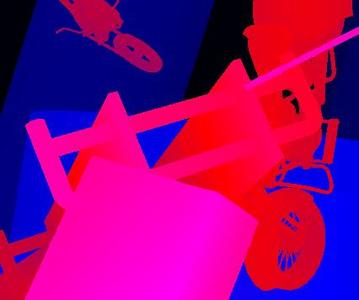}\\
        \vspace{-10pt}
        \includegraphics[width=\textwidth]{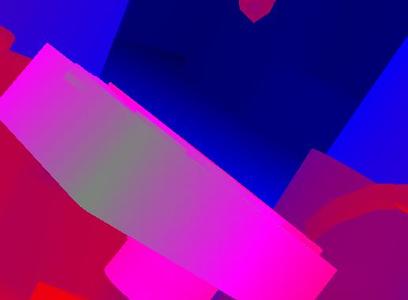}\\
        \vspace{-10pt}
        \includegraphics[width=\textwidth]{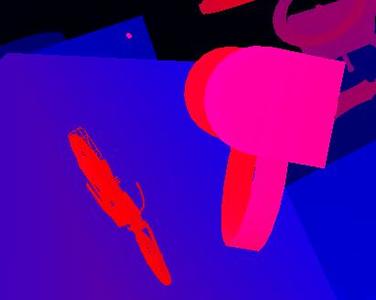}
        \vspace{-15pt}     
        \caption{\small{\textbf{Ground truth labels}}}
        \end{subfigure}                 
        \end{center}
        \vspace{5pt}
\end{subfigure}
\caption{Illustration of the estimated labels on each iteration of the \emph{Detect, Replace, Refine x2} multi-iteration architecture.
The visualised examples are from zoomed-in patches from the Middlebury and the Synthetic datasets.}
\label{fig:DRRx2}        
\end{figure*}

In Figure~\ref{fig:DRRx2}, we illustrate the estimated disparity labels after each iteration of our multi-iteration architecture \emph{Detect + Replace + Refine x2} that in our experiments achieved the most accurate results.
We observe that the 2$^{\mathrm{nd}}$ iteration further improves the fine details of the estimated disparity labels delivering a higher fidelity disparity field.
Furthermore, applying the model for a 2$^{\mathrm{nd}}$ iteration results in a disparity field that looks more ``natural", \ie, visually  plausible.

\subsubsection{KITTI 2015 qualititive results} \label{sec:zoom}

\begin{figure*}[t]
\centering
\renewcommand{\figurename}{Figure}
\renewcommand{\captionlabelfont}{\bf}
\renewcommand{\captionfont}{\small} 
\centering
\includegraphics[width=0.24\textwidth]{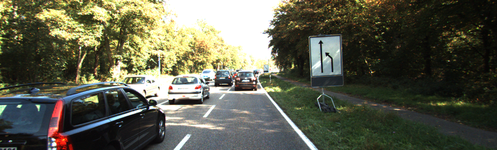}
\includegraphics[width=0.24\textwidth]{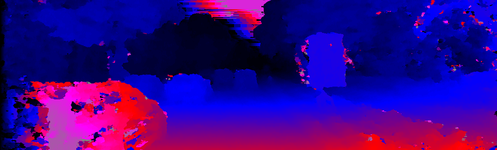}
\includegraphics[width=0.24\textwidth]{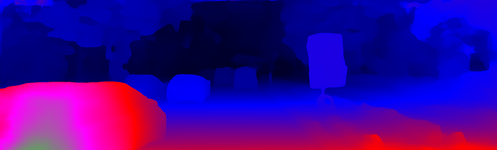}
\includegraphics[width=0.24\textwidth]{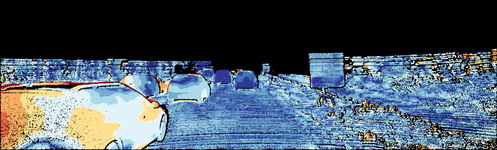}\\
\vspace{1pt}
\includegraphics[width=0.24\textwidth]{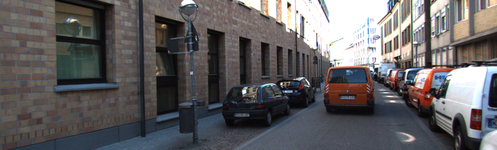}
\includegraphics[width=0.24\textwidth]{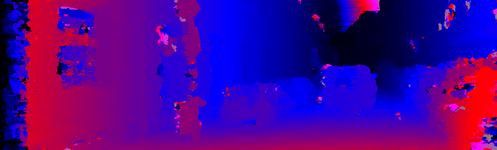}
\includegraphics[width=0.24\textwidth]{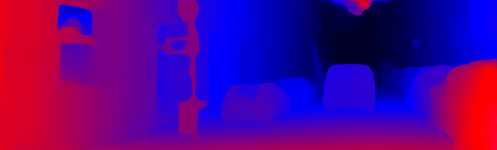}
\includegraphics[width=0.24\textwidth]{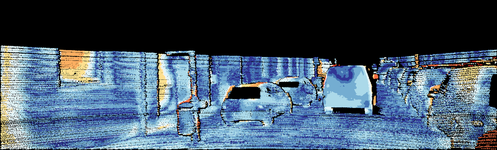}\\
\vspace{1pt}
\includegraphics[width=0.24\textwidth]{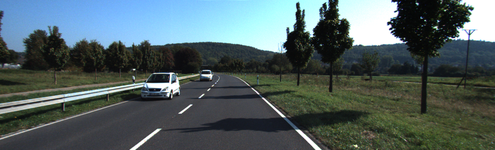}
\includegraphics[width=0.24\textwidth]{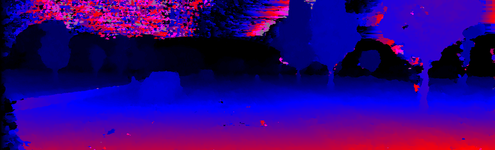}
\includegraphics[width=0.24\textwidth]{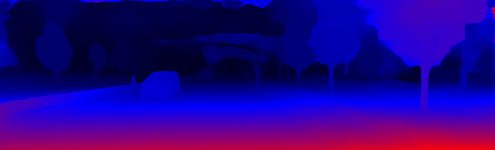}
\includegraphics[width=0.24\textwidth]{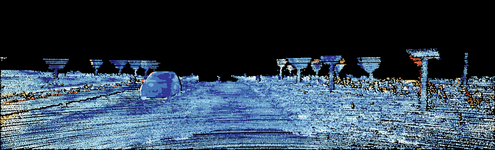}\\
\vspace{1pt}
\includegraphics[width=0.24\textwidth]{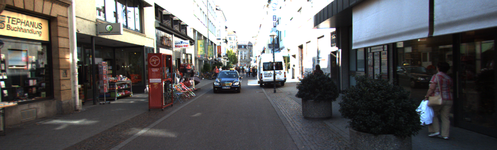}
\includegraphics[width=0.24\textwidth]{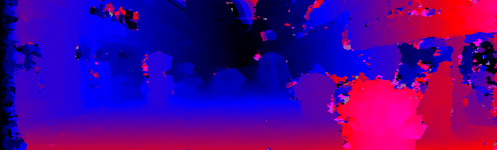}
\includegraphics[width=0.24\textwidth]{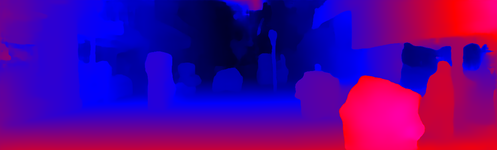}
\includegraphics[width=0.24\textwidth]{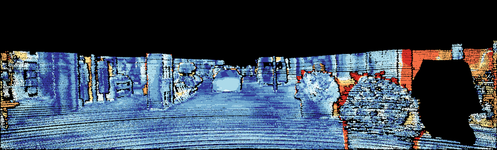}\\
\vspace{1pt}
\includegraphics[width=0.24\textwidth]{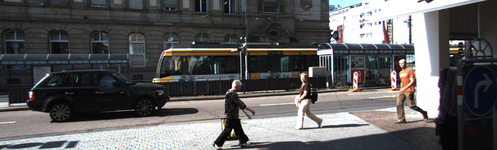}
\includegraphics[width=0.24\textwidth]{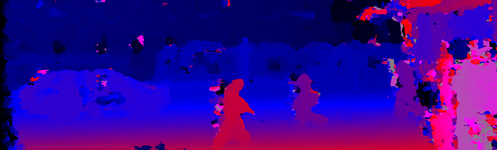}
\includegraphics[width=0.24\textwidth]{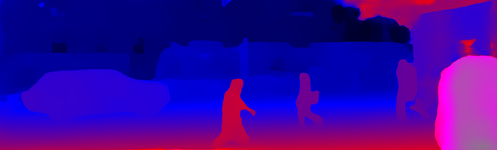}
\includegraphics[width=0.24\textwidth]{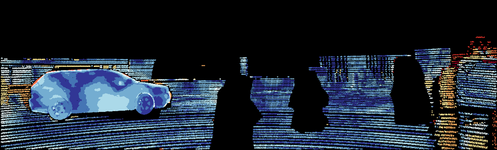}\\
\vspace{1pt}
\includegraphics[width=0.24\textwidth]{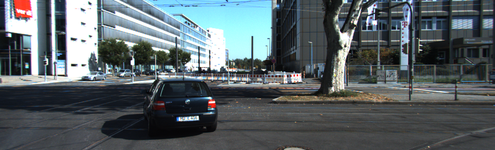}
\includegraphics[width=0.24\textwidth]{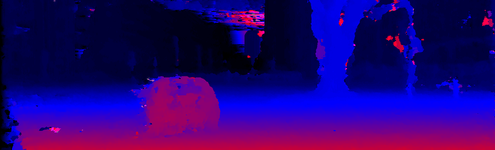}
\includegraphics[width=0.24\textwidth]{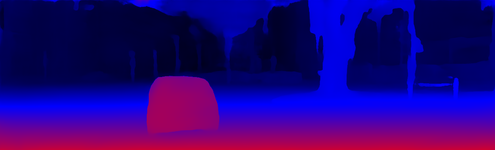}
\includegraphics[width=0.24\textwidth]{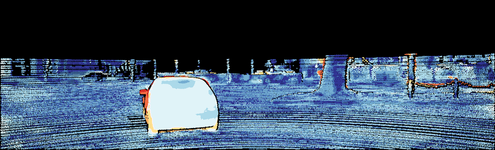}\\
\vspace{1pt}
\includegraphics[width=0.24\textwidth]{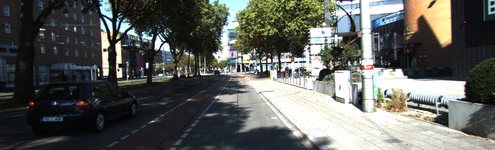}
\includegraphics[width=0.24\textwidth]{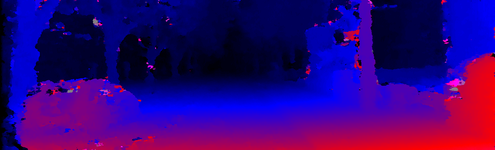}
\includegraphics[width=0.24\textwidth]{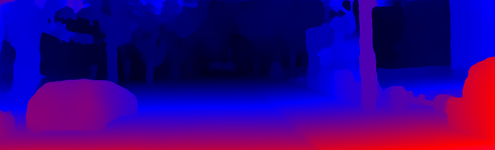}
\includegraphics[width=0.24\textwidth]{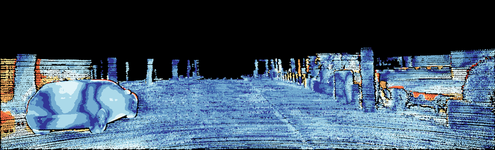}\\
\vspace{1pt}
\includegraphics[width=0.24\textwidth]{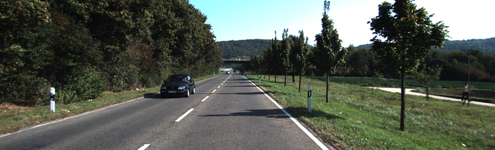}
\includegraphics[width=0.24\textwidth]{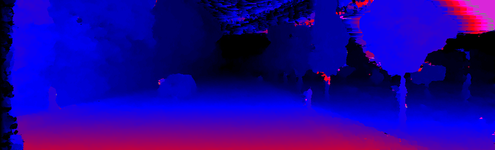}
\includegraphics[width=0.24\textwidth]{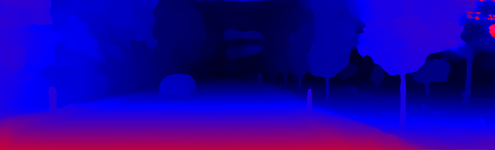}
\includegraphics[width=0.24\textwidth]{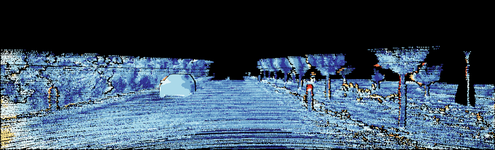}\\
\vspace{1pt}
\includegraphics[width=0.24\textwidth]{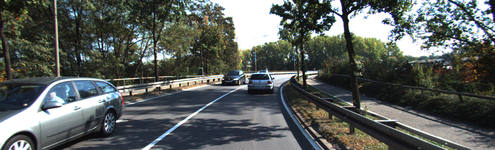}
\includegraphics[width=0.24\textwidth]{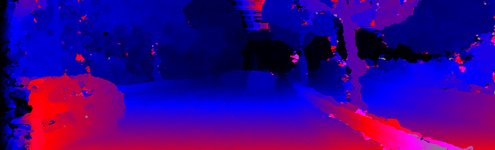}
\includegraphics[width=0.24\textwidth]{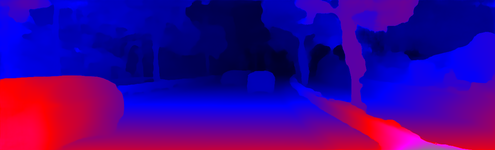}
\includegraphics[width=0.24\textwidth]{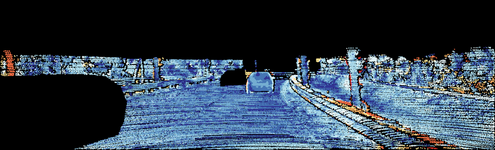}\\
\vspace{1pt}
\includegraphics[width=0.24\textwidth]{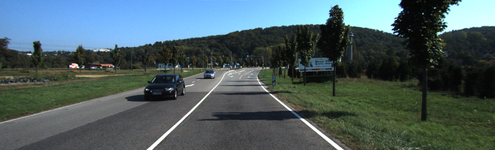}
\includegraphics[width=0.24\textwidth]{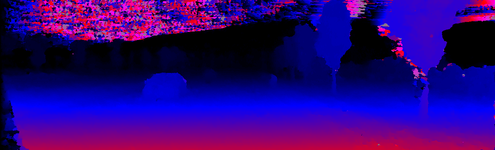}
\includegraphics[width=0.24\textwidth]{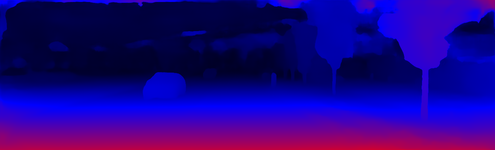}
\includegraphics[width=0.24\textwidth]{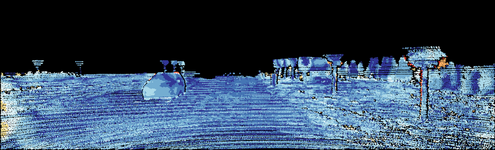}\\
\vspace{1pt}
\includegraphics[width=0.24\textwidth]{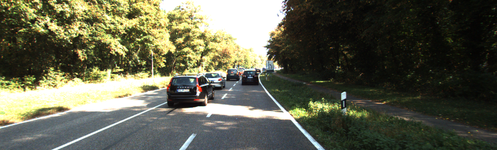}
\includegraphics[width=0.24\textwidth]{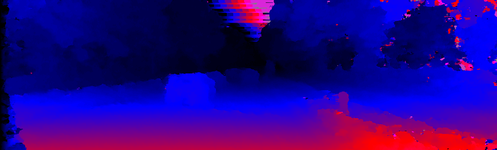}
\includegraphics[width=0.24\textwidth]{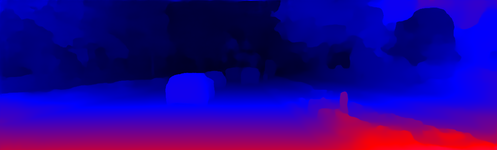}
\includegraphics[width=0.24\textwidth]{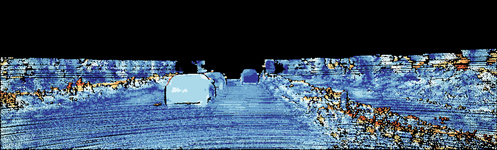}\\
\vspace{1pt}
\includegraphics[width=0.24\textwidth]{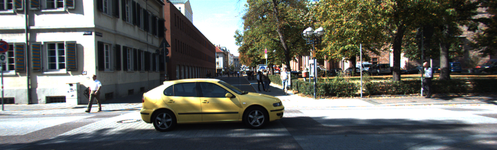}
\includegraphics[width=0.24\textwidth]{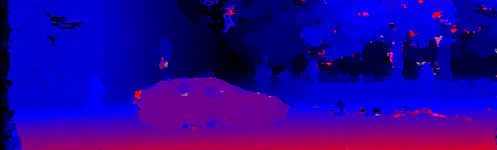}
\includegraphics[width=0.24\textwidth]{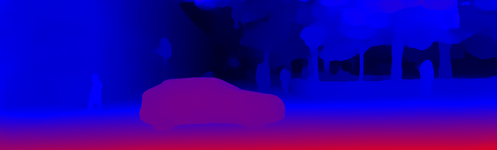}
\includegraphics[width=0.24\textwidth]{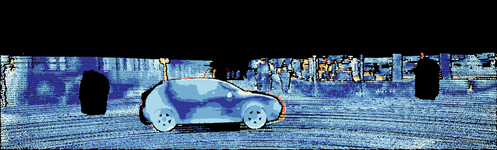}\\
\vspace{10pt}
\caption{
Qualitative results in the validation set of KITTI 2015. From left to right, we depict the left image $X$, the initial labels $Y$, the labels $Y'$ that our model estimates, and finally the errors of our estimates w.r.t. ground truth. 
}
\label{fig:qualitative}        
\end{figure*}
We provide qualitative results from KITTI 2015 validation set in Figure~\ref{fig:qualitative}.
In order to generate them we used the \emph{Detect + Replace + Refine x2} architecture that gave the best quantitative results. 
We observe that our model is able to recover a good estimate of the actual disparity map even when the initial label estimates are severely corrupted.

\section{Conclusions} \label{sec:Conclusions}
In our work we explored a family of architectures that performs the structured prediction problem of dense image labeling by learning 
a deep joint input-output model that 
(iteratively) improves some initial estimates of the output labels.   
In this context our main focus was on what is the optimal architecture for implementing this deep model. 
We argued that the prior approaches of directly predicting the new labels with a feed-forward deep neural networks are sub-optimal and we proposed to decompose the label improvement step in three sub-tasks: 
1) detection of the incorrect input labels, 2) their replacement with new labels, and 3) the overall refinement of the output labels in the form of residual corrections.
All three steps are embedded in a unified architecture, which we call \emph{Detect + Replace + Refine}, that is end-to-end trainable.
We evaluated our architecture in the disparity estimation (stereo matching) task and we report state-of-the-art results in the KITTI 2015 test set.

\section{Acknowledgements}
This work was supported by the \emph{ANR SEMAPOLIS} project and hardware donation by \emph{NVIDIA}.
We would like to thank Sergey Zagoruyko, Francisco Massa, and Shell Xu for their advices with respect to the Torch framework and fruitful discussions.
\FloatBarrier

{\small
\bibliographystyle{ieee}
\bibliography{mypaper_arxiv}
}

\end{document}